\documentclass{article}

\usepackage[preprint]{neurips_2023}

\usepackage[utf8]{inputenc} 
\usepackage[T1]{fontenc}    
\usepackage{hyperref}       
\hypersetup{breaklinks=true}

\usepackage{url}            
\usepackage{booktabs}       
\usepackage{amsfonts}       
\usepackage{nicefrac}       
\usepackage{microtype}      
\usepackage{xcolor}         
\usepackage{caption}
\usepackage{subcaption}
\usepackage{graphicx}
\usepackage{amsmath}
\usepackage{amssymb}
\usepackage{amsfonts}
\usepackage{amsthm}
\usepackage{mathrsfs}
\usepackage{dsfont}
\usepackage{algorithm}
\usepackage{algpseudocode}
\usepackage{authblk}

\DeclareMathOperator{\Unif}{Unif}
\newcommand{\RR}{\mathbb{R}}
\newcommand\norm[1]{\left\lVert#1\right\rVert}
\DeclareMathOperator{\sm}{softmax}

\newtheorem{thm}{Theorem}

\newtheorem{prop}[thm]{Proposition}
\newtheorem{lemma}[thm]{Lemma}

\newtheorem{rem}[thm]{Remark}

\title{Neural Computation Without Slots: Steps Towards Biologically Plausible Memory and Attention in Natural and Artificial Intelligence}

\author[a,b,*]{Shaunak Bhandarkar} 
\author[b,*]{James L. McClelland}   

\affil[a]{Princeton Neuroscience Institute, Princeton, NJ 08540}
\affil[b]{Department of Psychology, Stanford University, Stanford, CA 94305}

\begin{document}
                                                   
\maketitle

\footnotetext{\textsuperscript{*}All correspondence may be addressed to Shaunak Bhandarkar or James L. McClelland. Emails: shaunak@princeton.edu and jlmcc@stanford.edu.

An earlier version of the sparse, distributed, slot-free memory model in Section \ref{sec:kwinnermhn} was presented as an extended abstract at the CCN meeting in Oxford, UK, in August, 2023.

{\small{S. Bhandarkar, J.L. McClelland. Sequential Learning and Retrieval in a Sparse Distributed Memory: The K-winner Modern Hopfield Network. In Conference on Cognitive Computational Neuroscience, Oxford, UK. 24 -- 27 August 2023. \href{https://www.2023.ccneuro.org/view_paper041b.html?PaperNum=1361}{https://doi.org/10.32470/CCN.2023.1361-0}.}}

Data and code availability: All data generated for this project, as well as all code used to run simulations and perform data analyses, will be made publicly available within a GitHub repository upon publication in a peer-reviewed journal.}









\begin{abstract}
Many models used in artificial intelligence and cognitive science rely on multi-element patterns stored in "slots” -- dedicated storage locations -- in a digital computer. As biological brains likely lack slots, we consider how they might achieve similar functional outcomes without them by building on the neurally-inspired modern Hopfield network (MHN; Krotov \& Hopfield, 2021), which stores patterns in the connection weights of an individual neuron. We propose extensions of this approach to increase its biological plausibility as a model of memory and to capture an important advantage of slot-based computation in contemporary language models. For memory, neuroscience research suggests that the weights of overlapping sparse ensembles of neurons, rather than a dedicated individual neuron, are used to store a memory. We introduce the K-winner MHN, extending the approach to ensembles, and find that within a continual learning regime, the ensemble-based MHN exhibits greater retention of older memories, as measured by the graded sensitivity measure $d'$, than a standard (one-neuron) MHN. Next, we consider the powerful use of slot-based memory in contemporary language models. These models use slots to store long sequences of past inputs and their learned encodings, supporting later predictions and allowing error signals to be transported backward in time to adjust weights underlying the learned encodings of these past inputs. Inspired by these models' successes, we show how the MHN can be extended to capture both of these important functional outcomes. Collectively, our modeling approaches constitute steps towards understanding how biologically plausible mechanisms can support computations that have enabled AI systems to capture human-like abilities that no prior models have been able to achieve.
\end{abstract}





\section{Introduction}

It is common to model memory using an individual \emph{slot} for each item to be stored.  A slot is an ensemble of storage elements allocated uniquely to an item, distinct from the storage elements of any other item. Subsequent retrieval of information from memory often involves retrieving either the contents of the slot that best matches a query or a goodness-of-match-weighted combination of the contents of all slots \cite{Hintzman1986}. On a digital computer, slot-based computation can be realized by storing individual items in separate locations. A retrieval operation consists of comparing the contents of all locations to a query and evaluating each item's similarity to it in order to retrieve the best match or the weighted match across items.  

In recent years, the idea of using slots for storing and retrieving relevant memories has been embodied both in cognitive science -- as "episodic memory" \cite{botvinick2019reinforcement, lu2022neural, giallanza2024EGO, webb2024relational, Beukers2024working} -- and in artificial intelligence -- as "external memory" \cite{Graves2014NeuralTM, Graves2016, santoro2016metalearning, pritzel2017neural, Wayneetal2018, Ritter2018beenthere}. Furthermore, slot-based computations lie at the heart of the transformer \cite{vaswani2017attention}---the architectural backbone of today's large language models---both in performing query-based retrieval (called 'attention' in such models) and in enabling temporal credit assignment, as we will discuss below. 

\begin{figure}[t]
        \centering
        \includegraphics[width=\textwidth]{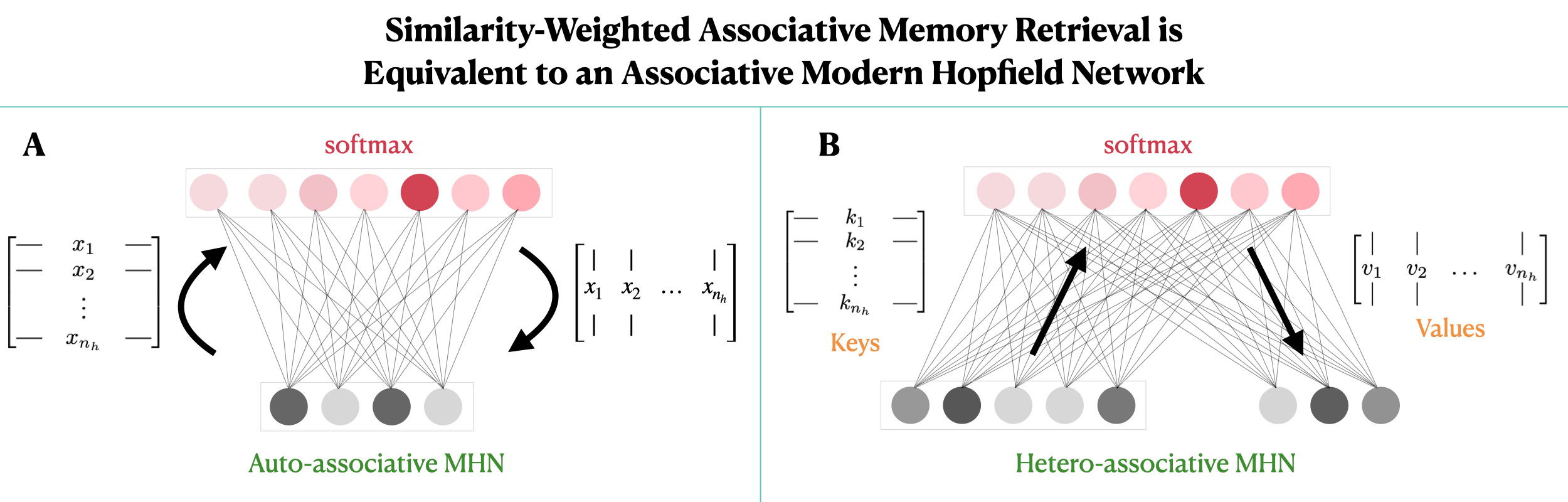}
        \caption{An illustration conceptualizing the equivalence between softmax-weighted associative memory retrieval and MHN-based retrieval, in two main settings. \textbf{A.} The original \emph{auto-associative} MHN, as conceived by Krotov and Hopfield \cite{KrotovHopfield2021}. For a given queried input, its dot products with predetermined vectors $x_1, \dots, x_{n_h}$ are computed, and the resulting softmaxed dot products are used to return a weighted combination of these same vectors. Such a computation may be realized within an autoencoder architecture with bidirectional connection weights storing the $x_i$'s. Performing multiple cycles of retrieval enables stable retrieval of the best-matching $x_i$. \textbf{B.} A \emph{hetero-associative} MHN that stores pairs of vectors---keys and values---and that uses keys to retrieve associated values. Crucially, instead of storing keys and values as neural activity \emph{states}, they may be encoded in the incoming and outgoing \emph{weights} of single neurons, respectively. The resulting network's computation coincides with that of the transformer self-attention mechanism. In contrast to the autoassociative MHN, this network is feedforward, i.e. it cannot be run for multiple cycles. In \textbf{A} and \textbf{B}, hypothetical input/output neuron and "memory" neuron activations are shown in gray and red, respectively.}
        \label{fig:tf_mhn_correspondence}
    \end{figure}

How might the computational equivalent of slot-based computation be implemented in the brain?  We take as our starting place the proposal that the storage elements used for past activity states are the connections among neurons, rather than activity patterns as such.  This view was articulated by Donald Hebb \cite{Hebb1949} at the dawn of the computational era.
An influential recent work characterizing memory storage and retrieval in this way is the modern Hopfield network (MHN) \cite{KrotovHopfield2021}. 
As in several early biologically inspired models \cite{grossberg1976b, von1973self, mcclelland1981retrieving}, 
this neural network model---which we call the \textit{original} or sometimes the \textit{auto-associative MHN}---stores individual memories in the connection weights that project into and out of a dedicated individual "memory" neuron (Fig. \ref{fig:tf_mhn_correspondence}A). For retrieval, each memory neuron computes a quantity representing the similarity of the query to the pattern stored in its incoming connection weights, and the system returns the best matching item or a similarity-weighted blend of all memory items, where the weightings for most items are typically very small.  The MHN and its predecessors \cite{Hopfield1982, krotov2016dense, demircigil2017associative} serve to capture a process called \emph{pattern completion}, whereby presentation of a noisy or incomplete memory and subsequent iteration of the network for multiple cycles (via the neuron's incoming and outgoing weights) gradually settles to a stable state that closely approximates a single stored pattern or a blend of such patterns, although for simplicity we simulate this process as a one-step retrieval operation.

Using the MHN as our starting place, we address three key issues.  We consider the first two together before turning to the third.

\textbf{1. Biological memory systems encode memories in connections of ensembles of neurons.} Although the auto-associative MHN constitutes a step towards understanding how memories may be stored and retrieved by relying on connection weights, it still uses distinct storage elements (the distinct incoming and outgoing connections of a dedicated neuron) to represent different memories. A more biologically realistic approach
\cite{Marr1971,mcnaughton1987hippocampal,kanerva1988sparse,rolls1990relative,oreilly1994hippocampal} is to model memories as being distributed across the connections of a small ensemble of neurons, each connected only to a subset of the neurons representing the item to be stored, and each participating in many different memories; indeed, evidence of sparse yet distributed coding of memories has been found in the hippocampus \cite{wixted2014sparse, wixted2018codingEM, stefanini2020distributed,katlowitz2025learning}, the brain region generally viewed as the storage site of episodic memories \cite{squire1992memory}. Inspired by the usefulness of slot-based memory systems in both cognitive science and AI, we consider whether aspects of their behavior can be captured as emergent properties of sparse distributed memory systems. If so, this would help build bridges between computational abstractions and their possible biological implementations; and if sparse, distributed memories have advantages, they might be taken up to enhance AI systems.

\textbf{2. Biological memory systems have finite capacity.} A common assumption in computational theory is to treat memory as unbounded, at least in principle \cite{Graves2016, turing1936turing}. However, brains have limited capacity. Thus, we seek to characterize how memories may be efficiently stored and retrieved within a fixed-capacity setting. In particular, unlike proposals in which a slot is assumed to be available for each new item to be learned, we require a learning policy for allocating the existing finite resources of a given memory system to learning each item as it is presented. 



We address these two issues in Section \ref{sec:kwinnermhn} by introducing the K-winner MHN, an extension of the original MHN. The key features of this model are a fixed, finite capacity; the use of graded updates to connection weights into and out of a small ensemble of memory neurons (of size $K$ greater than 1) to store each memory; and sparse network connectivity, so that each memory neuron's connections only encode a subset of the elements in the pattern corresponding to the memory to be stored.

Our K-winner MHN is more biologically aligned than the MHN since it allows memories to be stored in a graded and distributed manner across the weights of the network, rather than utilizing a single neuron (or slot) to represent a single memory. We compare the retention properties of the K-winner MHN to those of a fixed, finite capacity version of the original MHN in a continual learning setting, where items to be learned are presented sequentially so that as the capacity of the system is reached, newer memories will compete with older ones, conferring an advantage to recent memories.  As our results will show, the K-winner MHN has advantages compared to the original MHN in its ability to retrieve older memories, with only a slight cost to its performance on the most recent memories.

We now consider the third issue, one that is central to the functionality of today's transformer-based language models.

\textbf{3. Biological neural networks are unlikely to actively maintain arbitrary temporal sequences of prior states, yet must be able to update knowledge of past inputs based on future outcomes to support the acquisition of advanced cognitive abilities.} In transformer-based language models, prior state information in the form of sets of multi-valued key and value patterns, and other network state vectors associated with preceding items in context, are maintained for use in prediction and for propagating learning signals that allow the network to learn connection weights that map items to such representations for their effective utilization, when they are later tested after this learning has occurred. 
To illustrate this important capability more fully, we unpack the query-based attention mechanism used in 
decoder-only variants of the transformer as used in GPT models \cite{brown2020gpt3}. These systems contain stacks of transformer blocks, each containing an attention layer that computes and retains keys and values for long sequences of past time steps in slots, using query-based attention over all past slots to predict the next token at each position or time point in the sequence.  In this way, a query-based attention computation amounts to directly accessing slots that contain keys and values from previous time steps, comparing the current query against all of these keys, and using the resulting attention scores to produce a weighted combination of values. Thus, one consequence of query-based attention is that it requires explicit physical storage and maintenance of activity states corresponding to keys and values from as many as thousands or even millions of past timesteps.

\begin{figure}[t]
        \centering
        \includegraphics[width=\textwidth]{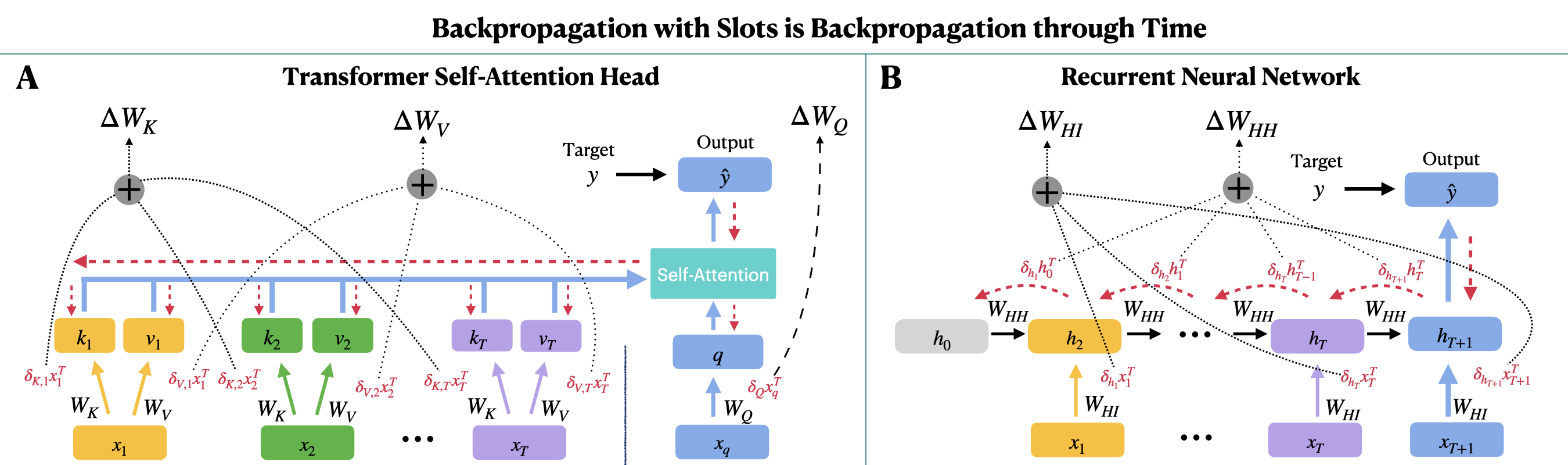}
        \caption{Backpropagation of gradients through an aggregation of slots effectively constitutes backpropagation through time---a ubiquitous issue among recurrent neural architectures more broadly. \textbf{A.} The backpropagation algorithm in a transformer self-attention head that processes the input sequence $x_1, \dots, x_T$ followed by the query $x_q$ for which the ground truth label is $y$. Gradients (shown via red arrows) backpropagate through the self-attention computation and iteratively produce gradient updates (i.e. for the key and value weights) that consist of an outer product between the gradient of a given key $k_t$ or value $v_t$ with the current context item $x_t$; see SI Appendix \ref{subsec:gradient_eqns}, for a detailed breakdown. These updates are then summed over the entire context window to produce the total gradient update for a given learnable weight matrix (e.g. $W_K$, $W_V$). \textbf{B.} The backpropagation through time (BPTT) algorithm in a standard recurrent neural network (RNN), where the hidden state is given recursively as $h_t = \sigma\left( W_{HH}h_{t-1} + W_{HI}x_t \right)$. For any learnable weight matrix, gradient updates for that matrix are iteratively computed for earlier time steps (again, as outer products between gradients of receiving units and activity patterns of sending units), and these updates are summed to produce the total weight update.}
        \label{fig:slotbased_backprop}
    \end{figure}

Furthermore, in order for the transformer to learn useful representations, including representations of the keys and values of items in its context, it is trained using the backpropagation learning algorithm. This requires propagating gradient signals backward through the slot aggregation operation to each item in each slot in the context window; the full gradient update for a given trainable weight matrix is the sum of the corresponding gradient updates for each item in the temporal sequence (Fig. \ref{fig:slotbased_backprop}A; see SI Appendix \ref{subsec:gradient_eqns} for details). In this manner, gradient updates for items occurring in \emph{previous} time steps allow weights that determine the representations of tokens and their corresponding keys and values to become useful for predictions at much \emph{later} points in time, so that later encounters with similar sequences will result in enhanced predictions.

Notably, retention of long sequences of prior state information is also required for the backpropagation through time algorithm \cite{werbos1990bptt} that is often used to train a wide range of recurrent neural architectures on long temporal sequences of information \cite{hochreiter1997long, van2016wavenet, wu2016google, wang2017learningreinforcementlearn, Wayneetal2018} (Fig. \ref{fig:slotbased_backprop}B).

Long sequences of past activation states are unlikely to be maintained as neural activity patterns in the brain.  Neural activity arising from sensory input and then propagating through subsequent stages of processing is transient, dying out quickly unless actively maintained, a process that competes with processing or maintaining prior or subsequent inputs, and has an effective maximum capacity of 4-5 unrelated items \cite{baddeley2020working}.  Indeed, if the biological implementation of retaining state information from past time steps relied on patterns of activity in neurons dedicated to successive time intervals, this would be extremely biologically costly. In a transformer, this would require the same number of neurons to maintain the key and value information from each past time step as were required to produce this representation when it was formed at the current time step. Storing such information in connection weights would be far more efficient, in that maintaining connections requires far less space and energy than maintaining neural activity states, and would exploit the fact that the brain provides 4-5 orders of magnitude more connection weights than neurons as a potential substrate for the storage of information.  From similar motivations, others have also explored weight-based maintenance of past state information and its use in making later predictions \cite{ba2016using}.

The question then arises: can a functional approximation of the transport of error gradient information that occurs in transformers, allowing later-arising prediction error to adjust the slow-changing weight matrices that transform past inputs into the appropriate keys and values, be implemented in a system that stores the relevant key and value information in connection weights?  Answering this question would be a major step toward understanding how biological brains might capture the powerful capability of transformers to update their encodings of past inputs based on later outcomes, an important contributor to their emergent capabilities.

We address this third issue in Section \ref{sec:slot_free_tf}, where we explore several candidate connection weight-based instantiations of a minimal decoder-only transformer, removing elements other than the self-attention mechanism. Given the success of large transformer-based architectures across a variety of task domains \cite{Zhao2023surveyllm, Bubeck2023SparksOA, brown2020gpt3, dosovitskiy2021visiontf, didolkar2024metacognitive}, we chose to work towards a slot-free instantiation of this minimal transformer to better understand how the human mind may be able to utilize similar yet more biologically aligned mechanisms to achieve success across the same task domains.

To store context information in the connection weights of a memory system in a way that is analogous to their representation in a transformer \cite{Ramsaueretal2021}, we utilize a variant of the original MHN in which keys and values are sequentially stored in the connection weights of a \emph{hetero-associative} MHN (Fig. \ref{fig:tf_mhn_correspondence}B). Unlike the autoassociative MHN, this network associate pairs of vectors, corresponding to the transformer's keys and values, which serve distinct roles. By design, when a query is presented, the stored keys are used to evaluate similarity to the query; the resulting similarity scores are used to retrieve a weighted combination of the corresponding values. We describe several variants of our connection-based transformer system that enable learning of weights that encode representations of past inputs to enhance their utilization for processing inputs at later times. Among these variants, we find one that approaches the performance of a standard decoder-only transformer on a task that depends on this capability.

In summary, our primary contributions are to design a sparse, distributed variant of the autoassociative MHN and a heteroassociative MHN-based analog of a minimal transformer architecture, both of which approach the functionality and efficiency of slot-based systems while retaining greater biological plausibility. 
In the \textit{Discussion}, we consider future extensions of our work that might integrate these two contributions.

\section{Sparse, Distributed, Slot-Free Memory Model} \label{sec:kwinnermhn}

In this section, we present a sparse distributed memory system whose design is centered around the first two issues articulated in the introduction:

\begin{enumerate}
    \item Rather than utilizing individual slots or neurons to represent individual memories, biological memory systems likely rely on representing items via the connection weights going into and out of a sparse, distributed ensemble of neurons. Moreover, any one neuron may participate in representing multiple memories.

    \item Memory systems have limited capacity and must allocate their resources to faithfully and continually store inputs as they are presented.
\end{enumerate}

Starting from the architecture of a fixed-capacity modern Hopfield network (MHN; see SI Appendix \ref{app:mhn_background} for more background on the MHN), we subsequently turn to our adjustments and extensions of this base architecture in addressing these two issues.


\begin{figure}[t]
        \centering
        \includegraphics[width=\textwidth]{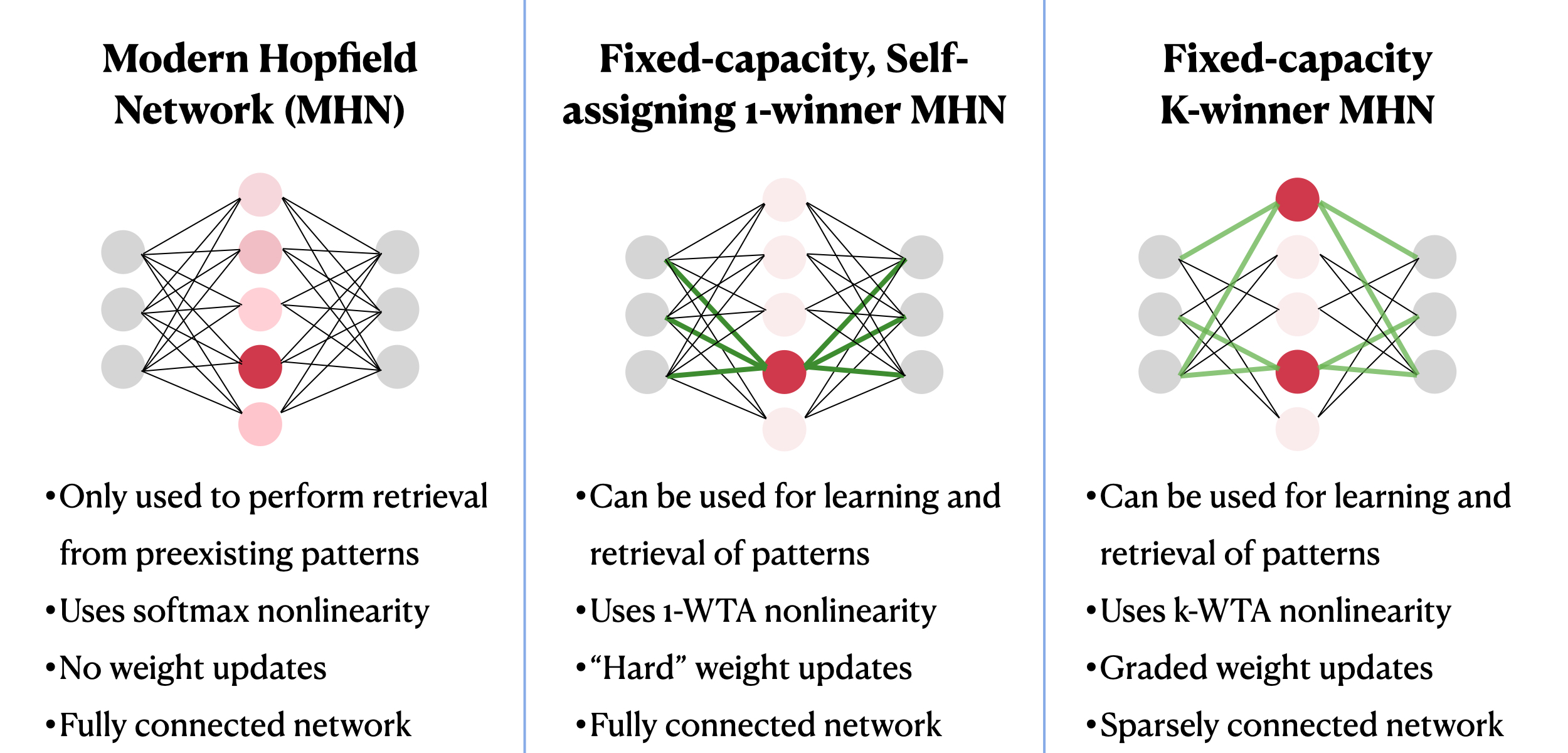}
        \caption{A comparison between the modern Hopfield network (MHN) and its variants. In contrast to the original MHN (left), K-winner MHNs (middle and right) are able to learn through weight updates as each input pattern is observed. Moreover, in the K-winner MHN, only weights projecting into and out of the $k$ hidden units with the largest values are updated (shown in green). In contrast to the 1-winner MHN (middle), general K-winner MHNs (right) allow for distributed hidden state representations, graded weight updates, and sparse network connectivity.}
        \label{fig:mhn_variants}
    \end{figure}

\subsection{Model Design}

To address the issue of modeling memory systems as a possibly unbounded set of slots or individual neurons to represent individual memories, we propose a fixed-capacity, sparse distributed memory system that continually learns as items are sequentially presented. Extending the modern Hopfield network (MHN; see Fig. \ref{fig:mhn_variants}, left), we present the K-winner MHN, an \emph{auto-associative} memory with visible layer size $n_v$ and hidden layer size $n_h$ (Fig. \ref{fig:mhn_variants}, right). We see the visible layer as a proxy for the input to the hippocampus (i.e., the entorhinal cortex) while the hidden layer is a simplified proxy for the hippocampus. The visible layer receives binary patterns with fixed sparsity $s_v$, corresponding to the fraction of the $n_v$ visible neurons set to $1$. The hidden layer forms a binary representation of the input with sparsity $s_h$, less than $s_v$. We assess retrieval of previously seen patterns from complete or partial input cues.  The original MHN is a special case of this approach, in which $s_h = 1/n_h$, so that a single hidden unit is chosen to represent each memory.

As in the MHN, the stored memories reside in two sets of weights: a matrix $M \in \RR^{n_h \times n_v}$ from the visible layer to the hidden layer, and a return matrix $M' \in \RR^{n_v \times n_h}$. Before training, each entry in each matrix is initialized uniformly to a number in $(0, 1)$. Additionally, we incorporate the concept of synaptic sparsity via a binary "fan-in" matrix $F \in \RR^{n_h \times n_v}$, in which each row of $F$
is randomly initialized to have a $f \cdot n_v$ of 1's, where $f$ is the fraction of visible units with connections to each hidden unit. We thus obtain the effective weight matrices $W = M \odot F$ and $W' = M' \odot F^T$ (where $\odot$ denotes element-wise multiplication), enforcing symmetry as in the MHN. Then, for input pattern $x$, the retrieved output is given by 
\begin{equation} \label{eq:kwinner_eqn}
    x^{out} = \sigma_{v}(W'\sigma_{h}(Wx)).
\end{equation}
For simplicity, the functions $\sigma_{l}$ ($l \in \{v,h\}$) are hard k-winner-take-all (k-WTA) functions, such that the $k_l := s_l \cdot n_l$ units of layer $l$ with the highest activations are set to 1 and the rest to 0. Using $z := \sigma_{h}(Wx)$ for the hidden representation, we generalize the MHN via the biologically inspired weight update rule \cite{von1973self, grossberg1976a}: 
\begin{equation} \label{eq:kwinner_update_rule}
    W_{ij} \leftarrow W_{ij} + \epsilon (x_{j} - W_{ij})z_{i}F_{ij}
\end{equation}
where $\epsilon \in (0, 1)$ is the learning rate. We apply the same update to the $ji$th entry in the return matrix $W'$ \cite{grossberg1976b}, imposing symmetry as in the MHN.
Note that the model becomes a binary, fixed-capacity version of the original MHN when when $s_h = 1/n_h$ (so that $k_h = 1$), $f=1$, and $\epsilon=1$, storing each new memory that enters the system in the incoming and outgoing connections of one of its $n_h$ neurons, completely replacing one old memory (Fig. \ref{fig:mhn_variants}, middle). We call this model the "fixed-capacity, self-assigning, 1-winner MHN," though for brevity we sometimes refer to this model as the 1-winner MHN, or simply just the MHN. 

When we compare the 1-winner MHN to its distributed counterpart ($k_h > 1, f < 1$), we equate the total number of weights (learnable parameters) between the models by setting $n_h$ for the K-winner case to $1/f$ times the $n_h$ of the localist case. Note that $k_h$ must be considerably larger than $1/f$ to ensure that all elements of an input are encoded in one or more connection weights, and that in this case knowledge of each feature tends to be stored in several connection weights, so that $\epsilon$ can be considerably less that 1.  Within these constrains, we chose specific values of $k_h$ and $\epsilon$ through exploratory simulations, reporting below on parameter settings that bring out interesting comparisons with the original (1-winner) MHN. 

\subsection{Experiments}

\begin{figure}[t]
    \centering
     \includegraphics[width=\textwidth]{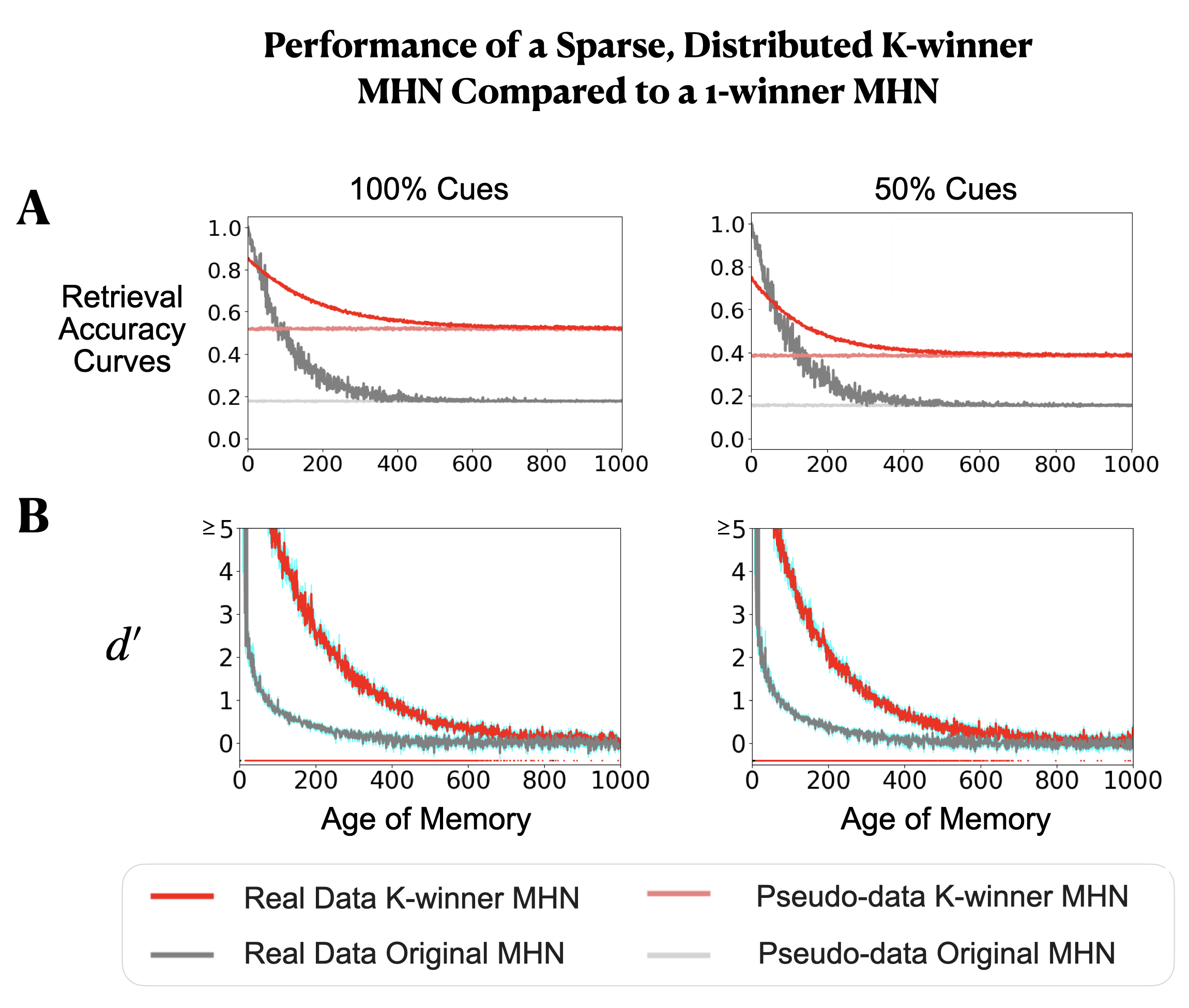}
  \caption{\textbf{A} Comparison of a candidate K-winner MHN's and original MHN's retrieval ability for memories of different ages, relative to their an untrained (pseudo-memory) baselines, for both 100\% cues (left) and 50\% cues (right). Results were averaged over $100$ independent runs of each model. \textbf{B} Retrieval sensitivity $d'$ for the given K-winner MHN and original MHN using 100\% cues (left) and 50\% cues (right). Cyan: $d'$ standard error. Horizontal segments show ages where K-winner MHN $d'$ is higher (red) and MHN $d'$ is higher (black), with uncorrected $p < 0.01$.}
  \label{fig:kwinnernet_retrieval_properties}
\end{figure}

\subsubsection{Sequential Learning with Unstructured Random Patterns}

\paragraph{K-winner MHN model details.}

We first present a K-winner MHN with $\epsilon = 0.3, f = 0.05, n_h = 2000, s_h = 0.025$ ($k_h = 50$) and compare it to a 1-winner MHN with $\epsilon = f = 1, n_h = 100, s_h = 0.01$ ($k_h = 1$). In our experiments, we test the K-winner MHN and original MHN's ability to retrieve previously learned patterns after a \emph{single} pass through a $4000$-pattern sequence of random binary patterns of dimension $n_v = 1000$ and sparsity $s_v = 0.1$. Each pattern has exactly 100 active neurons, uniformly sampled from the space of all 
all such length-1000 patterns. Preliminary analyses showed that no detectable trace of the first $3000$ patterns remained in either model. 
We then examined each model's performance in retrieving the most recent $1000$ patterns, with weights frozen, on the following measures.

\textit{Full pattern retrieval.} For the pattern $x_a$ of age $a$ (the $a$th most recently learned pattern), we assess the proportion $\rho(x_a)$ of the units of $x_a$ correctly retrieved at test time, when the full pattern $x_a$ is passed as input. For each age $a$, we also consider an untrained "pseudo-memory" $\tilde{x}_a$ (with $s_v = 0.1$) and compute the baseline retrieval measurement $\rho(\tilde{x}_a)$. The MHN perfectly retrieves the most recent pattern, but its performance rapidly degrades; a given memory is erased once its corresponding hidden neuron is assigned a new memory, leading to a simple exponential decay to the pseudo-memory baseline as we formally describe in SI Appendix \ref{app:theory_analysis}. In contrast, with the chosen values of $\epsilon = 0.3$ and $k_h = 50$, the K-winner MHN's performance is slightly worse initially, because the distributed weight update is not large enough to ensure that the winning units' outgoing weights can reproduce the stored item perfectly. 
However, its retrieval accuracy degrades more slowly (Fig. \ref{fig:kwinnernet_retrieval_properties}A, left) and the precise rate of decay is smaller than that of the MHN (SI Appendix \ref{subsubsec:appendix_RD_analysis}). To further assess quality of retrieval of real memories relative to pseudo-memories at each age, we computed the $d'$ measure from signal detection theory \cite{green1966signal} that has been used across prior studies on recognition memory \cite{standing1973learning, ratcliff1990list} to study sensitivity of retrieval of a particular memory relative to baseline 'noise'. (See Appendix~\ref{app:random_patterns} for discussion of the difference in baseline retrieval between the two models, details of our methods for estimating $d'$, and further simulation results based on an alternate metric of memory performance). Fig. \ref{fig:kwinnernet_retrieval_properties}B (left) shows a trial-averaged $d'$ measure indicating how well $\rho(x_a)$ represents retrieval above baseline. The K-winner MHN's advantage is reliable from age $16$.

\textit{Pattern completion.} We also assessed the proportion of units correctly retrieved from 50\% partial cues (half of the 1's in $x_a$ are randomly chosen and set to 0) from real memories and pseudo-memories (Fig. \ref{fig:kwinnernet_retrieval_properties}A, right).  Again, the K-winner MHN's performance is slightly worse than that of the original MHN at first, but degrades more slowly, as is confirmed by $d'$ in Fig. \ref{fig:kwinnernet_retrieval_properties}B (right). The K-winner MHN's advantage is reliable from age $10$.




\paragraph{K-winner MHN with Increased Network Connectivity.} 
Although the K-winner MHN reported in Fig. \ref{fig:kwinnernet_retrieval_properties} exhibits superior retention of older memories, its initial retrieval accuracy (i.e. for the most recent memory) falls below that of the MHN -- notably within the case of 50\% cues. We wondered whether a suitably initialized K-winner MHN could demonstrate greater initial retrieval accuracy while maintaining an advantage over the MHN in retaining older memories. Initial exploratory simulations suggested that one way to achieve this was by raising the fan-in parameter, $f$ while decreasing the number of hidden units to maintain the same total number of parameters corresponding to $f n_h$.   indeed, increased network connectivity would provide fuller coverage of the input units by the $k_h$ selected hidden units, potentially allowing for enhanced retrieval. Thus, we also analyzed the retention properties of a second K-winner MHN with $\epsilon = 0.3$, $k_h = 50$, and with $f$ increased to $0.1$, and $n_h = 1000$, comparing this to the parameter-matched Original MHN with $n_h = 100$ as before.

Our results in this case, summarized in SI Appendix Section \ref{app:kwin_mhn_larger_fan_in} and Fig. \ref{fig:kwinner_mhn_f=0.1_results}, indicate that this modified K-winner MHN does indeed exhibit higher (and, in fact, near-perfect) initial retrieval accuracy, both in the full pattern retrieval (100\% cues) and pattern completion (50\% cues) cases. Furthermore, measuring $d'$ reveals that this K-winner MHN does possess a memory retention advantage over the original MHN for a large class of "intermediate-age" memories (i.e. ages $\sim20 - 250$). However, increasing $f$ also appears to cause faster decay in raw retrieval accuracy, and the K-winner MHN no longer maintains a $d'$ recognition sensitivity advantage for very old memories (i.e. ages $\geq 300$). 

Altogether, we have found that the K-winner MHN with $f= 0.05$ compromises somewhat on initial retrieval accuracy while maintaining a significant retention advantage over the MHN for a large range of older memories; on the other hand, when $f$ is raised to $0.1$, the resulting K-winner MHN exhibits near-perfect initial retrieval accuracy while compromising somewhat on longer-timescale retention of (older) memories. Together, these results support the notion that across the entire family of possible K-winner MHN models (which includes the 1-winner MHN), there exists a trade-off between high initial retrieval accuracy and retention of older memories. 


\subsubsection{Memory Performance with Structured Patterns}

It is natural to wonder how our sparse distributed model fares against the 1-winner MHN when memories have nontrivial similarity structure. Items reflective of natural experience often possess similarity structure that can be captured by a hierarchical generative process \cite{saxe2019mathematical,griffiths2003hierarchical}. 
Accordingly, we use a generative model we call the tree-generating Chinese restaurant process (TGCRP) described in SI Appendix ~\ref{app:tgcrp_section} to explore this issue.  The TGCRP algorithm generates a tree of binary patterns of size $n_v$ and sparsity level $s_v$, in which each child node's pattern is the same as its parent node's pattern, except that $b$ of its $1$'s are switched to $0$'s and $b$ of its $0$'s are switched to $1$'s (at random), where $b$ specifies the number of bit flips going from a parent to a child; the leaf nodes of the tree are used as the patterns with which to train our memory models (see SI Appendix Fig. \ref{fig:sample_hierarchical_tree}). The parameter $b$ has a nonlinear effect on the similarity structure of the patterns when large numbers of them are produced by the TGCRP algorithm, so that 
$b = 5$ out of $100$ $1$-bits corresponds to patterns with $\sim50\%$ similarity and $b = 30$ out of $100$ $1$-bits corresponds to patterns with close to $10\%$ similarity -- the baseline similarity for random, non-structured patterns (see SI Appendix Fig. \ref{fig:treedata_covar_hist}). In presenting the results of these simulations we therefore focus on the mean similarities, rather the values of the $b$ parameter.

For each of 5 values of $b$, we generated large pattern trees using the TGCRP with $n_v = 1000$, $s_v = 0.1$, training the model on a total of $4000$ randomly ordered memory patterns sampled from the leaf nodes of the tree.  As before, we tested the $1000$ newest memories (of these 4000 patterns) for retrieval with weights frozen; we also sampled $1000$ untrained "pseudo-memory" patterns from the leaf nodes of the same tree. 
To assess the performance of our K-winner MHN and the original MHN with such structured patterns, we examined their respective retrieval sensitivities relative to the structured pseudo-data baseline, quantified by $d'$, as a function of memory age. We for each choice of $b$, we computed $d'$ measurements averaged over 10 independent $d'$ samples, each calculated using 20 independent runs, using full (100\%) cues (SI Appendix Fig. \ref{fig:treedata_dprimes}). Notably, the original MHN is largely insensitive to this parameter, while the K-winner MHN is strongly affected. The K-winner model performs worse than the original MHN when the mean pattern similarity is $0.49$, and approximately matches the original MHN when the mean similarity is $0.28$. The advantage of the K-winner model begins to emerge when mean pattern similarity is reduced below this level. When the mean pattern similarity is $0.17$, as in SI Appendix Fig. \ref{fig:treedata_dprimes}C), the advantage is quite clear, growing further as similarity is further reduced toward that in the unstructured case, as shown in SI Appendix Figure~\ref{fig:treedata_dprimes}F.
This analysis demonstrates that for structured patterns that possess nontrivial (but not excessive) similarity, the K-winner MHN exhibits a strong advantage for retaining older memories (relative to baseline structured pseudo-data).

Thus far we have considered recognition sensitivity with respect to a baseline of \textit{structured} pseudo-patterns drawn from the same data distribution as the learned memories. For higher levels of similarity, each such structured pseudo-pattern is likely to be similar to a pattern already stored in the memory, with the consequence that the retrieved pattern when the model is probed with a pseudo-pattern tends to be more similar to the probed pattern than it is to a random pattern stored in the memory.  This behavior is a form of similarity-based generalization from learned patterns to others drawn from the training data distribution.  Both the original and the K-Winner models exhibit this form of generalization, performing better on untrained pseudo-patterns than completely unstructured pseudo-patterns (see SI Appendix~\ref{app:treedata_unstructured_baseline}).  Delineating the relative advantages of the two model variants with respect to this form of generalization remains to be explored.



\subsection{The K-winner MHN Model Family}

Our goal in this section has been to begin to achieve convergence between conceptions of memory retrieval used in artificial intelligence and cognitive neuroscience.  To that end we have begun with the modern Hopfield network \cite{KrotovHopfield2021}, which builds an explicit link between slot-based computational architectures and architectures that store memories in connection weights, and extended it to a sparse, distributed, fixed capacity formulation we have called the K-Winner MHN, taking a step closer toward a biologically plausible implementation.  This version of the MHN has similarities to earlier biologically inspired distributed memory models \cite{oreilly1994hippocampal,Treves1992ComputationalCS} and to more recent models using sparse distributed representations in fuller implementations of a hippocampus-like neural network architecture \cite{norman2003modeling, schapiro2017complementary, wu2025simple}.

The novel finding from our work is the observation that a K-winner MHN can have superior retention of older memories at a relatively low cost in its initial retrieval accuracy.  With completely unstuctured patterns, initial accuracy can be high enough that the initial cost is not a serious consideration. This advantage reflects, in part, the fact that the learning that occurs when each new pattern is presented is only a partial update to the relevant connection weights, instead of a complete replacement of existing weights with new values.  Indeed, in SI Appendix~\ref{app:mhnlearningrate} we show that some of the retention advantage of the K-winner model can be captured in the 1-winner MHN by setting the learning rate $\epsilon$ to a value less than 1 so that each 'memory neuron' now stores traces of many memories, as in the K-winner model. 

The advantage of the K-winner MHN over the original holds not only for random patterns but also for patterns with hierarchically structured similarity relationships, as long as the mean pattern overlap is not too high.  At higher levels of similarity, the K-winner model's ability to retrieve the exact details of a previously stored pattern falls below that of the original MHN.  It is interesting, however, to note that this may be as much of a blessing as a curse.  As similarity grows both models exhibit sensitivity to the shared structure, allowing them to apply what they have learned about previously-presented patterns to novel patterns from the same underlying data distribution, allowing them to use similarity to existing memories to infer plausible fatures for novel memories.  Further work should explore this issue in more detail.

In situations where it is important to differentiate patterns with high input similarity, our models could be extended by incorporating a preprocessing step that reduces input similarity, in accordance with ideas first described by Marr \cite{Marr1971} and subsequently by \cite{mcnaughton1987hippocampal, kanerva1988sparse,oreilly1994hippocampal,rolls1990relative}. (See SI Appendix Section \ref{app:patsep} for further discussion).  

These initial explorations of the family of extensions of the Modern Hopfield network open up a vast model space for future explorations that we hope will increase of understanding of memory in both biological and artificial neural networks.  

\section{Connection Weight-Based Implementation of the Transformer} \label{sec:slot_free_tf}


In the previous section, we adapted the auto-associative modern Hopfield network (MHN) to a continual learning setting, allowing memories to be stored in connection weights and retrieved without the need for explicitly storing items in slots. We now extend our study to modify a more complex neural architecture: the transformer \cite{vaswani2017attention}. Here, we use the MHN as a starting place for solving the temporal credit assignment problem in a slot-free manner. 


Guided by biological considerations, a core feature of our approach is to use connection \emph{weights} to store information about previous item representations, rather than explicitly maintaining distinct neural activity \emph{states} (i.e. slots) for each of these items.
Consequently, one goal of our work is to utilize a fully connectionist memory system that encodes item representations within its weights (via "fast" local updates) as each item is presented. We do this by reframing the transformer architecture using a hetero-associative 1-winner MHN model endowed with additional "item reinstatement" weights to help with making predictions and facilitating credit assignment without using past state information (as will be explained subsequently).


\subsection{An In-Context Learning Task}

We consider the setting of \emph{in-context learning}, in which items presented during the contextual window must subsequently be used to assist in providing the correct output associated with a query item. Specifically, we construct a canonical in-context task which we call the \emph{Case Sequence Task} (Fig. \ref{fig:case_sequence_tf_description}A). In this task, inputs and outputs are one-hot tokens, as in transformer-based language models. Each input token is an uppercase or lowercase instance of some abstract letter type. For any particular input sequence, each token in the context window of length $C$ is of a different abstract letter type, uniformly sampled from a set of $L \geq C$ letter types. The instance of the letter type (uppercase versus lowercase) is chosen independently with equal probability for each letter type. This context sequence is followed by a query item 
encoding the assigned name of the letter type of one of the letters in the context window. The goal of the task is to output the case type of this particular letter. 

The Case Sequence Task is particularly well-suited for studying the transformer and its biologically inspired variants because it imposes strong constraints on the learning of keys, queries, and values. In particular, such a task encourages a model to keep track of letter identities as keys and queries and cases as values so that when the query probe is introduced, it is able to recall the case corresponding to the queried letter; in machine learning, such a task may be interpreted as having a "skip-trigram" structure \cite{elhage2021mathematical}. Crucially, such a task enables us to examine the capacity of a model to \emph{align} its key and query representations, i.e. in learning to generate intelligent queries that can be compared against previously stored keys to correctly identify the letter whose case must be outputted, and also to learn value representations in order to retrieve the correct case of the member of the queried letter class.

We emphasize that the tokens presented in the context window of the Case Sequence Task are arbitrarily assigned to the letter class and the case within that class that the token represents. Consequently, any model that solves this task must learn not only to accurately represent the class membership of both the context letters and letter identity probes, but also to learn how to map tokens to their arbitrarily assigned cases. 
The only information that can inform this mapping lies in the statistical correspondence between tokens in the context, probe, and target elements of each training example.

Unless otherwise stated, each of the results presented for the baseline transformer as well as the proposed MHN-based variants describes model performance on the Case Sequence Task with $L = C = 4$. Precise implementation details can be found in SI Appendix \ref{subsec:full_model_implementation_details}.


\subsection{A Baseline Transformer Architecture and Computational Goals} \label{subsec:tf_and_goals}

\begin{figure}[t]
        \centering
        \includegraphics[width=\textwidth]{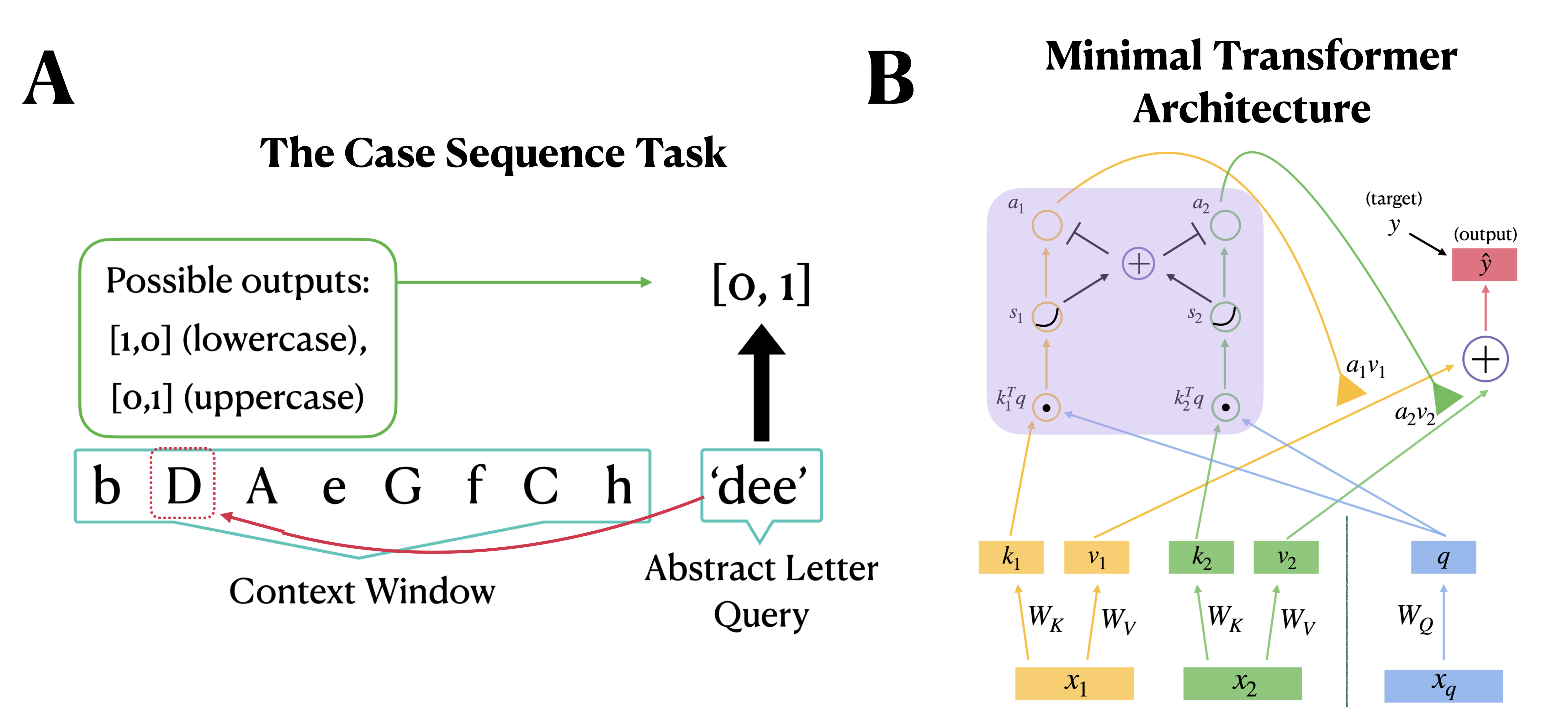}
        \caption{\textbf{A.} A visual description of the Case Sequence Task for a sample input sequence. \textbf{B.} A diagram of the "minimal" transformer architecture, in which the context window consists of two inputs (for simplicity). The dot products between the embedded query $q$ and the contextual keys $k_t$ are computed, and the softmax of these values is computed by applying an exponential nonlinearity to each $k_t^T q$ term and subsequently normalizing (illustrated in the purple shading). The resulting attention scores are used to modulate the linear combination of the values $v_t$ that is produced as the output. The supervisory training signal for gradient descent only arrives at this final output.}
        \label{fig:case_sequence_tf_description}
    \end{figure}
    

We first demonstrate that a "minimal" transformer model can solve the Case Sequence Task. This will constitute a baseline result and provide a concrete neural architecture from which to subsequently design our fully connection weight-based variants. This model's parameters consist of a query matrix $W_Q \in \RR^{N \times D}$, a key matrix $W_K \in \RR^{N \times D}$, and a value matrix $W_V \in \RR^{2 \times D}$. Here, $D$ is the dimension of the input data and $N$ is the dimension of the embedded key and query vectors. Moreover, we have set the dimension of the value vectors at $2$. Given a sequence of context inputs $x_1, x_2, \dots, x_C$ followed by a query $x_q$, the model computes $k_t = W_K x_t \in \RR^{N}$ and $v_t = W_V x_t \in \RR^2$ for each $1 \leq t \leq C$. It also computes the query embedding $q = W_Q x_q$. The model then outputs 

\begin{equation} \label{tf_equation}
    \hat{y} = \sum_{t=1}^{C} \text{softmax}\left( \frac{1}{\sqrt{N}} q^T k_t  \right) v_t =  \sum_{t=1}^{C} \text{softmax}\left( \frac{1}{\sqrt{N}} x_q^T W_Q^T W_K x_t  \right) W_V x_t.
\end{equation} 
This specific architecture entails using an attention mask such that the items $x_1, \dots, x_C$ do not interact with one another and the query item $x_q$ has access to the keys $k_t$ and values $v_t$ of all previous context items. During each iteration of training, $W_Q$, $W_K$, and $W_V$ are updated by minibatch gradient descent (with replacement) on the Sum-Squared Error (SSE) loss function (SI Appendix \ref{subsubsec:training_details}). A diagram of our minimal transformer architecture is shown in Fig. \ref{fig:case_sequence_tf_description}B.

In each of our biologically viable alternatives of the baseline transformer, we still make use of learnable query ($W_Q$), key ($W_K$), and value ($W_V$) matrices, which still serve to compute meaningful embeddings of input tokens. In the subsequent sections, we study how well our MHN-based transformer models (and the baseline transformer) perform on the Case Sequence Task. Crucially, because we aim to study how a model system like the brain might learn semantically-relevant task representations, we additionally test for the following phenomena:

\begin{enumerate}
    \item \textbf{Values encode case type.} Because each model is designed to output a vector indicating the case type of a particular letter in the context window, and any value $v_t = W_V x_t$ for a given one-hot input $x_t$ precisely identifies some column of $W_V$, we expect (after training) the column of $W_V$ containing the vector $W_V x_t$ to match either $[1, 0]^T$ (lowercase) or $[0, 1]^T$ (uppercase), in accordance with the case of $x_t$.

    \item \textbf{Alignment between query and keys.} Because the query representation $q = W_Q x_q$ is compared against the keys $k_t = W_K x_t$ (for all $1 \leq t \leq C$), the key corresponding to the item in the context window that the query item $x_q$ references should align with $q$. That is, when we compute the query-key similarity matrix $W_Q^T W_K$, each entry $\left( W_Q^T W_K \right)_{ij}$ for which row $i$ and column $j$ correspond to the same letter type would reasonably be expected to be higher than other non-letter-aligned entries.

    \item \textbf{Keys encode case-agnostic letter identity.} Because the query representation $q$ corresponding to any given letter type must align with the keys corresponding to both the lowercase \emph{and} uppercase versions of that letter, we probe the sub-matrix of $W_K^T W_K \in \RR^{D \times D}$ containing the dot product similarities between the keys for lowercase and uppercase letters. We expect a positive alignment between keys corresponding to the same letter type and non-alignment between keys for different letter types.
\end{enumerate}


Results for a baseline transformer trained on the Case Sequence Task are shown in Fig. \ref{fig:baseline_tf_results} (SI Appendix). Importantly, we find that the model attains perfect task accuracy and exhibits the query-key-value structure listed above. Consequently, the above three criteria constitute a set of desiderata that we wish to be reflected in our proposed MHN-based transformer models as well.

\subsection{MHN-based Instantiations}

Like the baseline transformer, our proposed MHN-based variants utilize the "slow-learning" weights $W_Q$, $W_K$, and $W_V$. Crucially, these models also contain the following "fast-learning" components:

\begin{enumerate}
    \item \textbf{An MHN Module with Fast Weights.} We use a hetero-associative MHN (with hidden size $n_h$) to reproduce the transformer self-attention mechanism. This network contains "fast weights" \cite{ba2016using} that sequentially encode keys and values (corresponding to each input item $x_t$ from the context window) one by one. These buffer weights later aid in producing the desired output corresponding to the query probe $x_q$. We call these key- and value-storing weights $W_{HK}^{(t)} \in \RR^{n_h \times N}$ and $W_{VH}^{(t)} \in \RR^{2 \times n_h}$, respectively. 

    \item \textbf{Fast Weights for Item Reinstatement.} We additionally utilize a second set of "fast weights" $W_{IH}^{(t)} \in \RR^{D \times n_h}$, whose columns are used to encode the context items $\{x_t\}_{1 \leq t \leq C}$ as they are presented. Subsequently, during the query timestep, the MHN hidden layer activation $a_q \in \RR^{n_h}$ is projected through the loaded-up weight matrix $W_{IH}$ to produce a \emph{reinstated} vector $\tilde{x}$. This vector represents a reconstructed combination of the input items $\{x_t\}_{1\leq t \leq C}$, weighted by their matches to the query probe (as reflected by $a_q$). \footnote{We denote the input, hidden, and output layers of this MHN by 'K' (for 'key'), 'H' (for 'hidden'), and 'V' (for 'value'). Additionally, we let the 'I' denote the input token layer of the overall model. As an example, $W_{HK}$ refers to a layer of weights going from the 'key' layer to the 'hidden' layer within the MHN.}  
\end{enumerate}


The weights $W_{HK}^{(t)}$, $W_{VH}^{(t)}$, and $W_{IH}^{(t)}$ are iteratively updated during each step $t$ of the context window by a "fast" Hebbian learning rule that has the effect of replacing a specific row of $W_{HK}^{t}$ and the corresponding columns of $W_{VH}^{(t)}$, and $W_{IH}^{(t)}$ by the current key, value, and item vector, respectively.
We used Hebbian learning because it constitutes a biologically plausible learning mechanism that may be involved in storage and associative recall of memories \cite{Hebb1949}. Furthermore, each time a new input sequence is processed, a new set of fast weights $W_{HK}^{(0)}$, $W_{VH}^{(0)}$, and $W_{IH}^{(0)}$ are initialized and used for subsequent storage. The use of binding items, keys, and values for future retrieval has been well-characterized in the cognitive science and machine learning literature \cite{giallanza2024EGO, Graves2014NeuralTM, Graves2016, fortunato2019rl_memory}.

In contrast to these weights that learn over the time steps \emph{within} any given input sequence, $W_Q$, $W_K$, and $W_V$ are updated \emph{across} entire input sequences via gradient descent. Gradients propagate backward through the fast weights, but do not actually update these weights. Taken together, this usage of memory-associated weights operating on fast timescales, in conjunction with weights operating on slower timescales that gradually consolidate query, key, and value representations, is in accordance with the Complementary Learning Systems Theory as posited by McClelland et al (1995) \cite{mcclelland1995cls} and relates to previous work on modeling "fast" and "slow" weights in recurrent networks \cite{ba2016using}.




\begin{figure}[t]
        \centering
        \includegraphics[width=\textwidth]{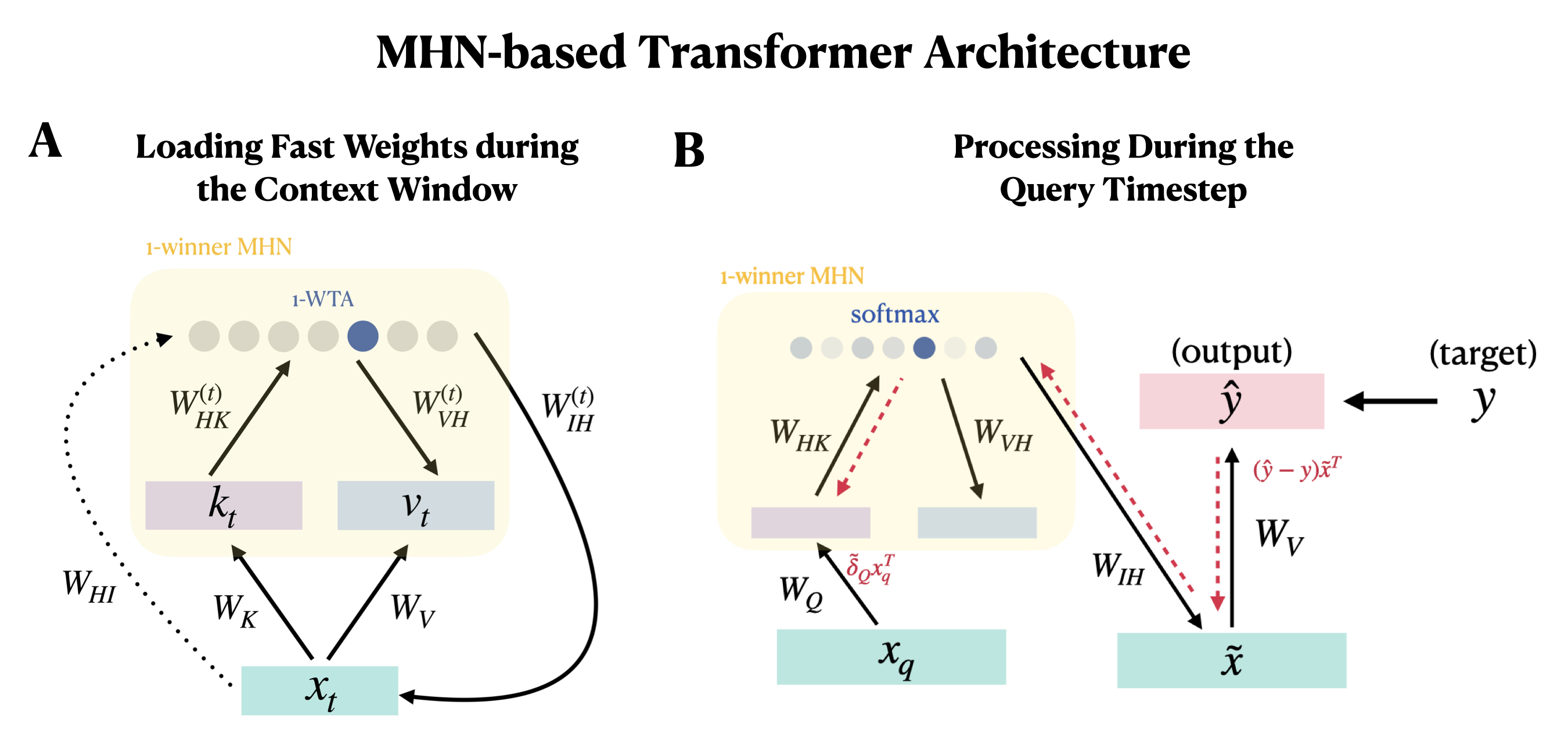}
        \caption{\textbf{A.} A diagram of the the MHN-based transformer during any time step $t$ of the context window. The item, key, and value representations are encoded into three separate sets of "fast" weights. For some of our models, we incorporate an "input projection" matrix $W_{HI} \in \RR^{n_h \times D}$ that helps ensure a new hidden neuron is recruited to represent each new context item. Throughout the context window, $W_Q$, $W_K$, and $W_V$ remain fixed. \textbf{B.} A diagram of the MHN-based transformer during the query time step. The buffer weights $W_{HK}, W_{VH}$, and $W_{IH}$ are now held fixed, and a reinstated input $\tilde{x}$ is produced. The output $\hat{y} = W_V \tilde{x}$ is used to backpropagate a signal (shown via red arrows) that is used to update $W_Q$ and $W_V$; gradient outer products used to update $W_V$ and $W_Q$ are also shown in red (see SI Appendix \ref{subsec:gradient_eqns} for precise equations). In both \textbf{A} and $\textbf{B}$, the yellow shaded rectangle depicts the 1-winner MHN module embedded within the larger network.}
        \label{fig:mhn_tf_fixed_WK}
    \end{figure}


\subsubsection{Model Implementation}

We present three candidate versions of a fully connection weight-based MHN-based transformer architecture, whose core mechanisms are summarized in Fig. \ref{fig:mhn_tf_fixed_WK}. The difference between our models lies in how $W_K$ is trained; complete implementation details can be found in SI Appendix \ref{subsec:full_model_implementation_details}. For any given input sequence $(x_1, \dots, x_C, x_q)$, $x_1$ through $x_C$ first are presented sequentially. Each $x_t$, along with its corresponding keys and values, are encoded within the model's fast weights, where a hard 1-winner-take-all function is used to determine the MHN hidden neuron whose weights will encode these representations. Then, during the query time step, the probe $x_q$ is presented, and activity propagates to the hidden layer of the MHN. Here, the loaded-up item reinstatement weights $W_{IH}$ are used to reinstate a blended input $\tilde{x}$, which projects through $W_V$ to produce the output $\hat{y}$. Applying one step of backpropagation, not through slots but rather layers of weights within just this time step, updates $W_Q$ and $W_V$. To train $W_K$, we take the following approaches (see SI Appendix Fig. \ref{fig:training_W_K}):

\begin{enumerate}
    \item \textbf{Leaving $\mathbf{W_K}$ fixed.} As a first pass, we chose to fix $W_K$ at initialization for the sake of simplicity; we reasoned that $W_Q^T W_K$ would still develop the same query-key similarity structure, because $W_Q$ can learn. We refer to this model as the Fixed $W_K$-MHN-Transformer.

    \item \textbf{Training $\mathbf{W_K}$ via the MHN.} A second approach is to project the reinstated item vector $\tilde{x}$ through $W_K$ and subsequently the hetero-associative MHN to produce an output. Applying one step of backpropagation (through weights, not slots) then provides an update for $W_K$ (see SI Appendix, Fig. \ref{fig:training_W_K}A). We call this model the MHN-Transformer.

    \item \textbf{Training $\mathbf{W_K}$ by query-key alignment.} During the query timestep, a query representation $q = W_Q x_q$ is passed as input to the MHN. Then, after $\tilde{x}$ is produced, it too can be projected via $W_K$ to the input of the MHN. Accordingly, our third approach utilizes a \emph{supervised delta rule} \cite{widrow1960adaptive, McClelland1985DistributedMA} to update the weights $W_K$ using $\tilde{x}$ as the input and $q$ as a supervisory signal (see SI Appendix, Fig. \ref{fig:training_W_K}B). We call this model the QK-MHN-Transformer. 
\end{enumerate}

For the third approach, we remark that such a learning rule is suitable for "cue-based recall" task settings in which keys and queries must be selectively aligned with one another. For tasks that do not possess this structure, other learning rules must be considered.

\subsubsection{Adding Input Projections to Improve Learning}

One limitation of MHN-based storage is that the $1$-winner-take-all mechanism may, on occasion, replace an existing key and value with those of a new item rather than recruiting a new hidden neuron to represent this new key and value through its incoming and outgoing weights. 
To prevent such interference, in addition to testing each of our proposed transformer variants, we also augment each of these variants with preexisting connection weights that go directly from the input token layer to the hidden layer of the 1-winner MHN; this helps in allocating distinct hidden neurons to represent distinct context items. We refer to these fixed connection weights as \emph{input projections}, and they are only used during the context window (see SI Appendix \ref{subsubsec:W_HI_details} for implementation details). 

We also remark that the role of these auxiliary weights may parallel certain biological mechanisms underlying memory storage. For instance, in the hippocampus, it has been hypothesized that mossy fibers form sparse, strong, and unidirectional projections from the dentate gyrus (DG) to CA3, which assist in producing well-separated neural representations in the CA3 during learning (though they are not necessarily involved in recall) \cite{Treves1992ComputationalCS, cerasti2010mossyfibers}. (Additionally, there is evidence that complementary pathways---such as the perforant pathway from the entorhinal cortex (EC) to the CA3---may subsequently be used to perform memory retrieval \cite{lassalle2000reversible, lee2004hippocampus}.)

\subsection{Results}

\begin{figure}[t]
        \centering
        \includegraphics[width=\textwidth]{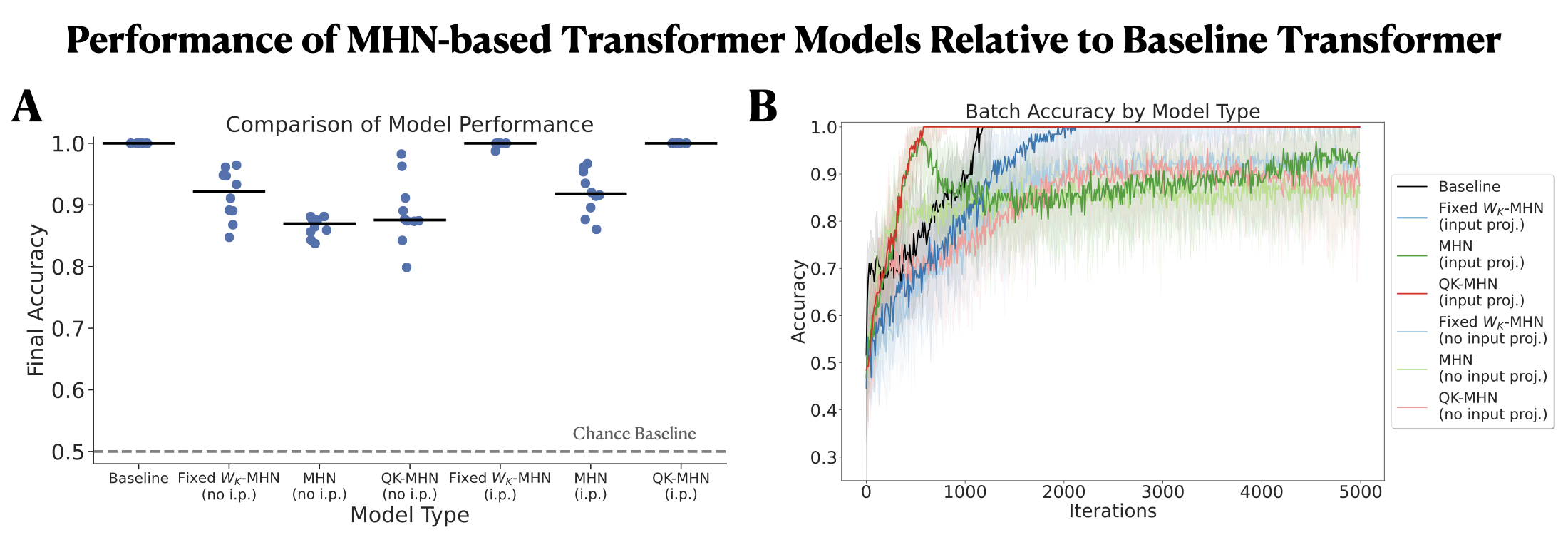}
        \caption{Comparisons of accuracy on the Case Sequence Task across model types. \textbf{A.} Average accuracies for $10$ independent runs of each model over the last 1,000 training steps. Median accuracies for each model are shown as black bars. (Here, "i.p." denotes "input projections.")  \textbf{B.} Median batch training accuracy curves shown across $10$ independent runs of each model, with shading to indicate the minimum and maximum accuracies attained for any given model at any particular iteration.}
        \label{fig:model_performances}
    \end{figure}


Having presented the design for each of our proposed MHN-based transformer variants, we present these models' performance on the Case Sequence Task. Fig. \ref{fig:model_performances}A illustrates the batch accuracy near the end of training across $10$ random initializations of each model\footnote{The names for each model in Fig. \ref{fig:model_performances} indicate which kind of transformer model it is. For instance, 'Baseline' refers to the baseline transformer and 'MHN' refers to the MHN-based transformer (with $W_K$ learnable).}---including the baseline transformer. The baseline model successfully solved the Case Sequence Task with $100\%$ accuracy across all runs. Turning to the MHN-based transformer models, we find that the addition of input projections boosts accuracy across each model type, meaning that it successfully allows new memories to be encoded in the MHN module without overwriting existing ones. 

Among the models without input projection, which attain accuracies of around $90\%$ on the task, the Fixed $W_K$-MHN-Transformer appears to more consistently attain higher ending accuracy. This indicates that the Fixed $W_K$-MHN-Transformer is already expressive enough to attain high task accuracy; however, fixing $W_K$ comes at the cost of relevant semantic structure learned by the model. In particular, fixing $W_K$ to be a random matrix prevents the uppercase and lowercase key for any given letter type from developing meaningful correlations (as $W_K^T W_K$ is static).

Fig. \ref{fig:model_performances}A also shows that the Fixed $W_K$-MHN-Transformer---when endowed with input projections---largely attains perfect accuracy at the end of training. Moreover, the QK-MHN-Transformer also attains perfect task accuracy, indicating that this model's approach for training $W_K$ (in addition to $W_Q$ and $W_V$) has not compromised on accuracy. In fact, the QK-MHN-Transformer even realizes the desiderata listed in Section \ref{subsec:tf_and_goals}: the $W_V$ matrix has learned to represent case, keys and queries corresponding to the same (resp. different) letter type are aligned (resp. anti-aligned), and keys corresponding to the same (resp. different) letter type possess aligned (resp. anti-aligned) representations (SI Appendix, Fig. \ref{fig:best_mhn_tf_model}). This largely matches with the structure learned by the baseline transformer (SI Appendix, Fig. \ref{fig:baseline_tf_results}).

We also present median accuracies across each training iteration for each model type, which allows us to gauge the timescales over which each model is able to learn (Fig. \ref{fig:model_performances}B). Notably, the baseline transformer (shown in black) learns quickly compared to the Fixed $W_K$-MHN-Transformer and input projections (shown in dark blue), illustrating that having a learnable $W_K$ matrix can speed up learning.
However, the given QK-MHN-Transformer with input projections (shown in dark red) learns more rapidly than the given baseline transformer, sharply rising to perfect accuracy within 1,000 training iterations. This suggests that training $W_K$ by simply aligning the reinstated key vector $\tilde{k} = W_K \tilde{x}$ with the query representation $q = W_Q x_q$ is an efficient "training bias" for rapidly solving the task as well as learning to encode queries, keys, and values in a semantically relevant manner. 

Interestingly, we note that the standard MHN-Transformer (i.e. with $W_K$ learnable) and with input projections (shown in dark green) results in a median accuracy that steeply rises to nearly 100\% at the start of training---and appears to match the initial accuracy trajectory for the QK-MHN-Transformer with input projections---but subsequently decreases and settles around $90\%$. This trend appears to indicate a tension between training $W_K$ and the pair of weights $(W_Q, W_V)$ for this particular model, and remains to be further explored in future work.

Finally, to further draw out the comparison between the baseline transformer and the QK-MHN-Transformer, we analyze the general trends observed in $W_V$, $W_K^T W_K$, and $W_Q^T W_K$ across the iterations of training in either network. Crucially, we find that in both models, $W_V$ successfully learns representations of case before the expected structure begins to emerge for keys and queries (SI Appendix, Figs. \ref{fig:tf_qkv_over_learning} and \ref{fig:qk_mhn_tf_qkv_over_learning}); a formal mathematical derivation of this fact is presented in SI Appendix \ref{subsubsec:W_V_learns_first}. Moreover, for the baseline transformer, the expected structure in $W_K^T W_K$ and $W_Q^T W_K$ emerge in parallel once values have been learned (SI Appendix, Fig. \ref{fig:tf_qkv_over_learning}). This indicates that once the values are learned, structure pertaining to keys and queries emerges through the same process. In contrast, for the QK model, alignment between queries and corresponding keys in $W_Q^T W_K$ emerges before uppercase-lowercase alignment of keys occurs in $W_K^T W_K$ (SI Appendix, Fig. \ref{fig:qk_mhn_tf_qkv_over_learning}). This trend suggests that the learning dynamics for the QK model constitutes a staged process, with values first learning to encode case, then queries learning to align with keys, followed by uppercase and lowercase keys learning to align with one another. 

\section{Discussion}

In this article, we have considered two central paradigms in artificial intelligence and cognitive science that utilize \emph{slot-based computation}, and we have presented alternative models that aim to achieve the same broad computations while avoiding the use of slots. The first paradigm relies on copies of previous experience, stored as distinct vectors in slots, as a basis for memory. Here, we extend the auto-associative modern Hopfield network, which replaces each slot with the input and output weights of a neuron-like processing unit, to the family of K-winner modern Hopfield networks. These models show how memories can be encoded in connection weights distributed across a population of neurons; our findings show that these connection weights can have advantages over the one-winner MHN, which corresponds to the standard slot-based implementation. The second paradigm relies on slots to hold copies of vectors corresponding to items stored in the context window of a transformer-based language model to support context-sensitive generation of outputs and backward transport of prediction error information, allowing models to learn representations of past items to better support future predictions. Here, we use the hetero-associative modern Hopfield network to store the relevant context information, and show how this architecture can be used to implement an approximation to the backward transport of prediction error information as implemented in transformers.  For simplicity, we did not use a K-winner variant of the auto-associative modern Hopfield network in our work in this second paradigm. Instead, we rely on our findings from using the K-winner approach in the first part of this work to propose that the encodings of context information---stored as slots in language models---might be stored in connection weights distributed across neuronal populations within the brain.  In this way, we have pointed the way toward understanding how the biological hardware of the brain could implement key aspects of the computations that underlie the power of language models to use temporally delayed error information to learn how to use past context to support future predictions.  This research is just a first step along both of these lines, with many limitations and extensions. There is also an emerging body of work by others that complements our findings and addresses the promise of this work in other ways \cite{wu2025simple,benna2021place,rolls2025slow,sun2023organizing}, making this an exciting domain for further exploration.  

Thus far, we have treated the paradigms of slot-based memory and slot-based attention as separate.  However, in both the brain and artificial neural networks, there may not be a clear distinction between external memory and current contextual state.  When a human or a machine is engaged in language-based cognition, there is no specific time window that limits the extent of prior information that can be considered relevant.  In the human memory literature, informed both by patterns of behavior in normal humans and deficits in amnesic patients, the prevailing view since the 1950's has been that only a very limited number of distinct items can be maintained in active form in a hypothetical construct labeled 'working memory'; beyond this limit, a system often labeled 'long-term memory' is usually invoked.  Retention of previously unfamiliar and arbitrary information including new factual information and information related to specific episodes and experiences is thought to depend on the hippocampal system in the medial temporal lobes of the brain.  Language models in artificial intelligence can have much larger immediate context windows, but there is always some finite limit to this, requiring the system, to be able to access some form of external memory.  A hippocampus-like structure could be the structure that the biological brain uses, both for storing current context beyond a very limited time window and for many aspects that are often referred to as human long-term memory.

\section{Acknowledgements} 

SB was supported by a summer internship award (2022) from the Symbolic Systems Program at Stanford University.  We thank members of the PDP lab at Stanford for discussions. 


\medskip

{
\small



\bibliographystyle{unsrt}
\bibliography{references}

}

\newpage 

\section*{Supplementary Information (SI)}

\appendix
\renewcommand{\thefigure}{S\arabic{figure}} 
\setcounter{figure}{0} 
\renewcommand{\thetable}{S\arabic{table}}
\setcounter{table}{0} 
\renewcommand{\theequation}{S\arabic{equation}}
\setcounter{equation}{0}
\section{The K-winner MHN} \label{appendix}

\subsection{Background on the Modern Hopfield Network} \label{app:mhn_background}


The modern Hopfield network (MHN) is a connectionist autoassociative memory model that can be used to retrieve patterns from a fixed collection of patterns that have each been stored in the weights of an individual memory neuron \hyperlink{Ramsaueretal2021}{[S1]}. When a query vector $\xi \in \RR^N$ is presented, retrieval is facilitated by evaluating the similarity of $\xi$ against each of the $M$ patterns in $X = \begin{bmatrix}
    \vert & \vert &  & \vert \\
    x_1   & x_2 & \dots & x_{M}   \\
    \vert & \vert &  & \vert 
\end{bmatrix} \in \RR^{N \times M}$ (stored in memory), and using them to retrieve a similarity-weighted vector as output. Dynamics in both the MHN and its previous iterations are governed by an energy function that result in various update rules and corresponding storage capacities. In particular, Ramsauer et al (2021) \hyperlink{Ramsaueretal2021}{[S1]} show that, up to constant terms, the MHN can be described by the energy function
\begin{equation}
    E = -\text{lse}\left( \beta, X^T \xi \right) + \frac12 \xi^T \xi
\end{equation}
and the corresponding update rule
\begin{equation} \label{mhn_update_rule}
    \xi^{\text{out}} = X \text{softmax}\left( \beta X^T \xi \right)
\end{equation}
which decreases the energy across each application and enables exponential storage capacity in the number of features $N$. We may visualize the update given by Eq. \ref{mhn_update_rule} as that of an autoencoder with a visible (i.e. input) layer of size $N$ and one hidden layer of size $M$. The stored memories $x_1, \dots, x_M \in \RR^N$ are stored in the rows of the visible-to-hidden-layer weights $W := X^T$ as well as in the hidden-to-visible-layer weights $W' := X$, with a softmax nonlinearity being applied at the hidden layer. It should be noted that the MHN is only used to retrieve from a fixed set of patterns; the weights in $W$ and $W'$ encode these fixed patterns and are never updated. 

\subsection{Baseline Retrieval Accuracy and Metrics for Memory Performance with Random Patterns} \label{app:random_patterns}

\subsubsection{Baseline Accuracy}

Here we briefly discuss the difference in baseline retrieval accuracy between the K-winner MHN and the original 1-winner MHN when trained on random untrained patterns (shown in Fig.~\ref{fig:kwinnernet_retrieval_properties}A).

We first consider the original MHN's retrieval accuracy (where $n_v = 1000,  n_h = 100$). Here, all of the real patterns and "pseudo-memory" patterns were sampled uniformly from the collection of length 1000 binary patterns with 100 1-bits. When an untrained pattern is presented to an MHN with $n_h = 100$ other memories stored in it, it will replace the best-matching memory presently in the system, meaning that this pattern has is expected to have higher-than-average correlation with the new incoming pattern.
This is why the retrieval accuracy baseline for the 1-winner MHN is greater than $0.1$---the average correlation between random patterns. More precisely, the accuracy baseline should roughly equal the maximum of $n_h = 100$ samples from the distribution of possible pattern overlaps; see Section \ref{app:theory_analysis} for more details.

We do not have a full analysis of the exact baseline retrieval accuracy of the K-winner MHN. However, we can offer an intuitive characterization of some factors contributing to its higher baseline accuracy (as compared to the original MHN). Recall that in the K-winner MHN studied in the main text (where $n_v = 1000$, $n_h = 2000$, $k_h = 50$, $f = 0.05$, and $\epsilon = 0.3$), each of the hidden units 'sees' only a subset of the bits from any given input pattern. In this case, the distribution of best matches will be based on a smaller sample size ($fn_h = 100$ rather than $n_h = 2000$ elements); correspondingly, the standard deviation of a sampled proportion is proportional to $\frac{1}{\sqrt{n}}$ and the maximum of such a distribution will be larger than the maximum of a distribution of samples each with a larger $n$. Additionally, it is possible that having $k_h > 1$ adds a measure of robustness to retrieval (under uncertainty), enabling the retrieved output to be averaged over $k_h$ rows of the weight matrix $W$ rather than a single row. Other factors likely influence the actual baseline retrieval accuracy as well.

\subsubsection{Sensitivity metric $d'$ and relevant simulation details} \label{subsubsec:dprime_details}

The sensitivity metric $d'$ characterizes the distance between two probability distributions in units of a composite measure of their variability \cite{green1966signal}. To generate the top and middle panels of Fig. \ref{fig:kwinnernet_retrieval_properties}C, we averaged over 10 independent sample estimates of $d'$, which were each themselves an average across $20$ independent runs of the given model type. 
Within each sample, for the memory $x_{a,i}$ and pseudo-memory $\tilde{x}_{a, i}$ of age $a$ in run $i$, we use $\rho(x_{a,i})$ and $\rho(\tilde{x}_{a, i})$ to compute $\delta_{a,i} = \rho(x_{a,i}) - \rho(\tilde{x}_{a, i})$. We then compute $d' = {\mu_{\delta_a}}/{\sigma_{\delta_a}}$ where $\mu_{\delta_a} = \frac{1}{20} \sum_{i=1}^{20} \delta_{a,i}$ and $\sigma_{\delta_a} = \sqrt{{\sum_{i=1}^{20}(\delta_{a,i} - \mu_{\delta_a})^2}/{20}}$.

\subsubsection{Raw Difference Metric} \label{subsubsec:appendix_RD_analysis}

\begin{figure}[!h]
    \centering
        \includegraphics[width=\textwidth]{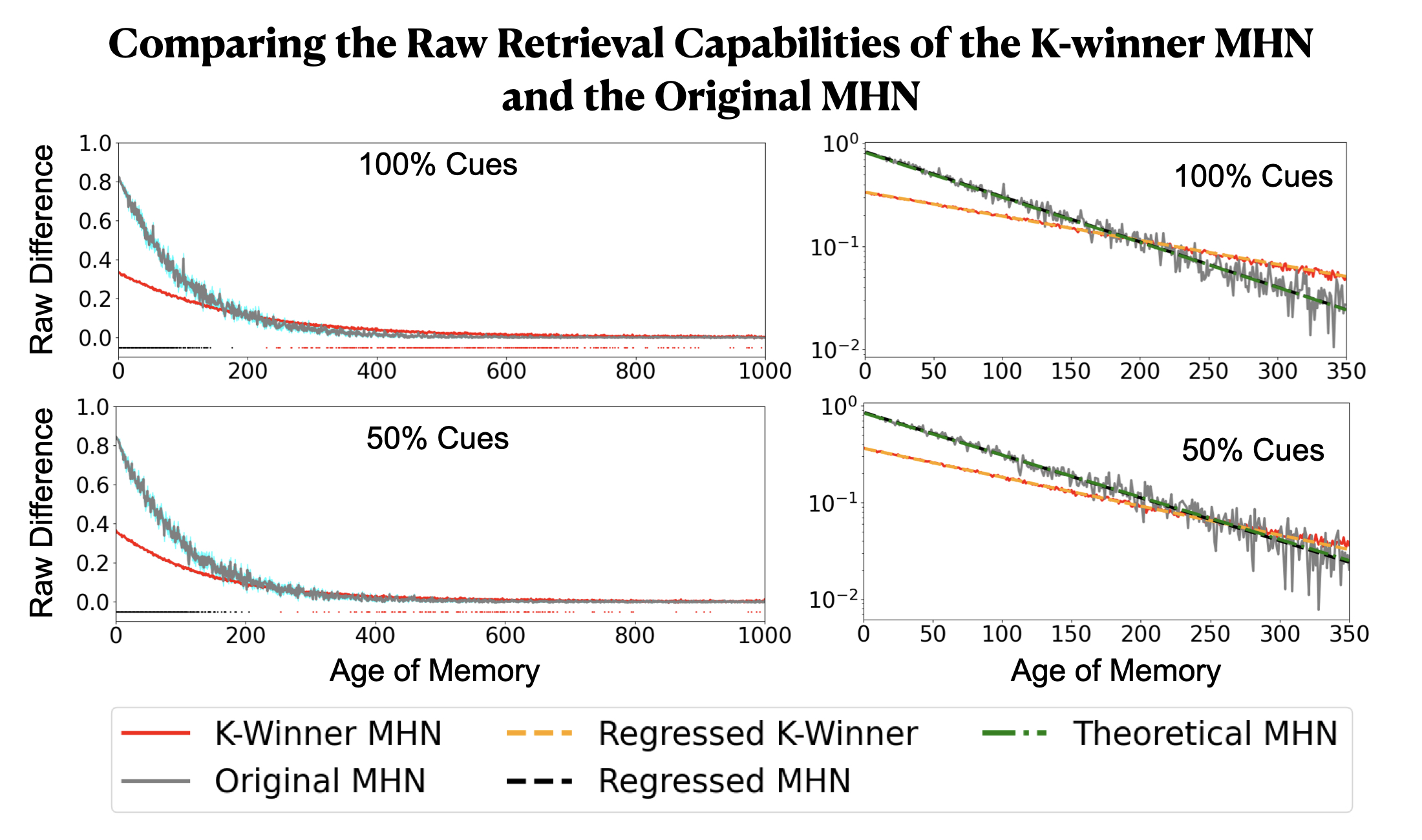}
        \caption{Raw difference (R.D.) across memory ages
        for a parameter-matched K-winner MHN and Original MHN, for 100\% and 50\% cues.
        For the left panels, standard error is shown in cyan, and horizontal segments show ages where the K-winner MHN has a higher R.D. (red) and original MHN has a higher R.D. (gray), with uncorrected $p < 0.01$. For the right panels, plotting R.D. on a log scale reveals precise exponential decay regression curves for the K-winner MHN (orange dashed) and the Original MHN (black dashed); all $r^2$ values $\geq 0.96$. Moreover, the theoretically predicted R.D. decay curve for the MHN is shown (green dashed).}
    \label{fig:iid_rawdiffs}
\end{figure}


In addition to the $d'$ measure, we chose to study a \emph{raw difference} (R.D.) \emph{metric}, averaged over the results of $10$ samples from our (large-scale) networks that are studied in the main text (see Fig. \ref{fig:iid_rawdiffs}). Each R.D. sample itself is computed over 20 independent runs of the given model type; defining $\delta_{a, i} := \rho(x_{a, i}) - \rho(\tilde{x}_{a, i})$ (as explained in Section \ref{subsubsec:dprime_details}), we compute the R.D. sample as $\mu_{\delta_a} = \frac{1}{20} \sum_{i=1}^{20} \delta_{a, i}$. Intuitively, this simplified metric (which is amenable to theoretical characterization) quantifies a network's raw retrieval capability \emph{past baseline}, owing to the learning process. It should be noted that, unlike the $d'$ measure, this metric does not clearly map on to measures of signal detection (or discriminability) and is rather considered here because of its simplicity. 

In studying the R.D. metric, we computed exponential regression curves for the R.D.'s of each network over memory age (Fig. \ref{fig:iid_rawdiffs}). This was motivated by the fact that, for the original MHN, the retention for a given memory decays by a factor of $\left( 1 - \frac{1}{n_h} \right)$ (proability of slot replacement) at each timestep; for a more detailed explanation, see Section \ref{app:theory_analysis} below. We perform exponential regression ($\text{R.D.}(a) \sim C e^{-\beta (a-1)}$) over the first $200$ memory ages. For the networks with 50\% cues, we obtain $(C_{\text{k-win}}, \beta_{\text{k-win}}) = (0.363, 0.007)$ and $(C_{\text{MHN}}, \beta_{\text{MHN}}) = (0.858, 0.010)$, respectively; for the networks with 100\% cues, we obtain $(C_{\text{k-win}}, \beta_{\text{k-win}}) = (0.336, 0.005)$ and $(C_{\text{MHN}}, \beta_{\text{MHN}}) = (0.833, 0.010)$, respectively (Fig. \ref{fig:iid_rawdiffs}, right panels). 

In Fig. \ref{fig:iid_rawdiffs}, we also compute the theoretical decay curve for the original MHN (derived in Section \ref{app:theory_analysis} below): $(C_{\text{MHN-theory}}, \beta_{\text{MHN-theory}}) = \left( 1 - s_v - \sqrt{\frac{2c}{n_v}\left( 1 - s_v \right) \log(n_h)} , -\log\left( 1 - \frac{1}{n_h} \right)\right)$. Here, $c \in (0, 1]$ is the pattern cue level, expressed as a proportion. Each MHN theory curve matches its empirical counterpart. Indeed, for the case of 50\% cues, we obtain $(C_{\text{MHN-theory}}, \beta_{\text{MHN-theory}}) = (0.836, 0.010)$; for the case of 100\% cues, we obtain $(C_{\text{MHN-theory}}, \beta_{\text{MHN-theory}}) = (0.809, 0.010)$. Comparing the empirical constants $C_{\text{k-win}}$ and $C_{\text{MHN}}$ shows that the original MHN yields higher R.D.'s for small ages; comparing decay rates ($\beta$) shows that the K-winner MHN's retrieval performance degrades more slowly, supporting its superior retention of older memories.

\subsubsection{Theoretical Analysis of MHN Retention Decay} \label{app:theory_analysis}

The raw difference (R.D.) metric used in the previous section captures the retention decay in both our networks (past their respective baselines); understanding how an exponential decay curve specifically arises, as well as how the network hyperparameters ($\epsilon$, $f$, $k_h$) contribute to the particular exponential decay curve obtained, warrants further theoretical analysis of the K-winner MHN. In this section, we present a detailed analysis of how the original MHN's (where $k_h = 1, \epsilon = 1, f = 1$) R.D. curve follows an exponential decay equation, given some light constraints on the cue level $c$ and size of the network; in future work, we seek to theoretically characterize the decay equation for any K-winner MHN, using ideas from order statistics \hyperlink{deHaan}{[S2]}.

\begin{thm} \label{thm:mhn_theory}
    Suppose $M$ is a slot-based (localist) MHN with input size $n_v$, input sparsity level $s_v$, and (sufficiently large) hidden size $n_h$, feedforward function $f$, and that $M$ has been trained on $N$ patterns where $N \to \infty$. Assume additionally that there exists a positive constant $c_0 << s_v \log\left( \frac{1}{s_v} \right)$ such that $n_h \leq e^{c_0 n_v}$. Let $x_a$ denote the pattern of age $a$ (with $a \geq 1$, and letting $a = \infty$ mean that $x_a$ is a pseudo-pattern), and moreoever let $x_{a, c}$ ($c \in (0, 1]$) be a partial cue for $x_a$, where a random proportion $1 - c$ of the $1$ bits in $x_a$ are made $0$. Finally, let $\tilde{x} := f(x_{a, c})$. Then, we have that $$\theta := \frac{\log\left(1 - \left(1 - \frac{1}{n_h^{1+\epsilon}}\right)^{\frac{1}{n_h - 1}}\right)}{k_v \log(s_v)} << 1$$ for any $\epsilon \in (0, 1)$; moreover, if it holds that $c > \theta$, it follows that $$\text{R.D.}(a) := \mathbb{E}\left[ \frac{1}{k_v} \tilde{x}^T x_a  - \frac{1}{k_v} f(x_{\infty, c})^T x_{\infty}\right]$$ $$= \left( 1 - s_v - \sqrt{\frac{2c}{n_v}\left( 1 - s_v \right) \log(n_h)} \right) \left( 1 - \frac{1}{n_h} \right)^{a-1} + O\left(\frac{\log\log(n_h)}{\log(n_h)}\right).$$ Here, $k_v = s_v n_v$ and $\text{R.D.}(a)$ denotes the expectation of the raw difference at age $a$, where the expectation is taken over all possible runs of the model $M$.
\end{thm}

To prove this theorem, we need to first prove two propositions.

\begin{prop} \label{prop:xa_not_in_memory}
    Suppose that, with the conditions of the above proposition, the slot that held the memory $x_a$ in the MHN $M$ has been replaced, or alternatively that $a = \infty$. Then, with this additional condition, we have $$\mathbb{E}\left[ \frac{1}{k_v} \tilde{x}^T x_a  \right] = s_v + \sqrt{\frac{2c}{n_v}\left( 1 - s_v \right) \log(n_h)} + O\left(\frac{\log\log(n_h)}{\log(n_h)}\right).$$
\end{prop}

\begin{rem}
    Observe that this proposition gives a formula for the baseline retrieval accuracy of the MHN $M$ (for the cue level $c$), or equivalently, the expected retrieval accuracy for any given pseudo-pattern of cue level $c$.
\end{rem}


\begin{proof}
    Suppose that $p_1, \dots, p_{n_h}$ are the patterns stored in $M$. Then, for the input query $x_{a,c}$, and any random stored pattern $p_i$, we have $$x_{a, c}^T p_i = \sum_{j : (x_{a,c})_j = 1} (p_i)_j$$ is a sum of $c k_v$ random variables. Assuming these variables are approximately i.i.d. Bernoulli random variables with parameter $\frac{k_v}{n_v} = s_v$, the Central Limit Theorem implies that $x_{a, c}^T p_i \sim \mathcal{N}\left( \frac{c k_v^2}{n_v}, \sqrt{c k_v s_v (1 - s_v)} \right)$. At this point, the stored pattern $p_i$ such that $x_{a, c}^T p_i$ is maximal (over $i \in \{1, \dots, n_h\}$) will be retrieved. That is, retrieving $p_i$ is equivalent to finding the largest value in a sample of $n_h$ Gaussian random variables. To further this insight, we make note the following theorem from Extreme Value Theory (see \hyperlink{deHaan}{[S2]}, Example 1.1.7):

    \begin{thm}
        Suppose that $s_1, s_2, \dots, s_N$ are i.i.d. samples from the standard Gaussian distribution $\mathcal{N}\left( 0, 1 \right)$. Let $S := \max_{1 \leq j \leq N} s_j$. Then, as $N \to \infty$, $\frac{S - a_N}{b_N}$ converges in distribution to the Gumbel distribution given by the pdf $p_G(z) = e^{-z - e^{-z}}$, where $a_N := \sqrt{2\log{N}} - \frac{ \log\left( 4\pi \log(N) \right)}{\sqrt{2\log(N)}}$ and $b_N := a_N^{-1}$.
    \end{thm}


     (See Hall (1979) \hyperlink{Hall1979}{[S3]} for precise convergence rates). With the terminology of the above theorem, we thus have reason to suspect that $$\frac{\mathbb{E}\left[ S \right] - a_N}{b_N} \to \mathbb{E}[G] = \gamma,$$ where $G$ is a standard Gumbel variable (as specified by the theorem). It is well-known that the expectation of this variable is the Euler-Mascheroni constant, $\gamma$. While convergence in distribution need not always imply convergence in expectation, in this case, convergence in expectation does hold; Dasgupta et al (2014) \hyperlink{DasGuptaEtal2014}{[S4]}, for example, calculate an asymptotic expansion of $\mathbb{E}[S]$ (in Theorem 5.1 of their paper), which in particular implies that $\mathbb{E}[S] = \sqrt{2 \log N} + O \left( \frac{\log \log N}{\sqrt{\log N}} \right)$.
    
    Applying this fact to our problem, for the random variables $x_{a, c}^T p_i$ ($1 \leq i \leq n_h$) given by $\mathcal{N}\left( \frac{c k_v^2}{n_v}, \sqrt{c k_v s_v (1 - s_v)} \right)$, we have $$\mathbb{E}\left[\arg \max_{1 \leq i \leq n_h} x_{a, c}^T p_i\right] = \frac{c k_v^2}{n_v} + \sqrt{c k_v s_v (1 - s_v)}\mathbb{E}[S]$$ $$ \approx \frac{c k_v^2}{n_v} + \sqrt{2 c k_v s_v (1 - s_v) \log(n_h)} + O\left(\frac{\sqrt{k_v}\log\log(n_h)}{\sqrt{\log(n_h)}}\right).$$ Now, suppose that the pattern $p_l$ (for some $1 \leq l \leq n_h$ maximizes $x_{a, c}^T p_i$. Then, the MHN will simply retrieve the pattern $\tilde{x} := p_l$. Notice that $$\mathbb{E}\left[\tilde{x}^T x_a\right] = \mathbb{E}\left[p_l^T x_{a, c}\right] + \mathbb{E}\left[p_l^T (x_a - x_{a, c})\right].$$ The first summand is exactly the quantity we have just approximated, and the latter summand is simply the expected amount of correlation between a random pattern with $k_v$ $1$-bits and a random pattern with $(1 - c)k_v$ $1$-bits. This is just $(1 - c)k_v \cdot \frac{k_v}{n_v} = \frac{(1-c)k_v^2}{n_v}$ (again, simplifying by assuming that each bit of $p_l$ is a Bernoulli variable). Therefore, (for $n_h$ sufficiently large), we have $$\mathbb{E}[\tilde{x}^T x_a] = \frac{c k_v^2}{n_v} + \sqrt{2 c k_v s_v (1 - s_v) \log(n_h)} + O\left(\frac{\sqrt{k_v}\log\log(n_h)}{\sqrt{\log(n_h)}}\right) + \frac{(1-c)k_v^2}{n_v}$$ $$ = \frac{k_v^2}{n_v} + \sqrt{2 c k_v s_v (1 - s_v) \log(n_h)} + O\left(\frac{\sqrt{k_v}\log\log(n_h)}{\sqrt{\log(n_h)}}\right),$$ and therefore $$\mathbb{E}\left[ \frac{1}{k_v} \tilde{x}^T x_a  \right] = s_v + \sqrt{\frac{2 c}{n_v} (1 - s_v) \log(n_h)} + O\left(\frac{\log\log(n_h)}{\sqrt{k_v}\sqrt{\log(n_h)}}\right).$$ Finally, from the conditions of Theorem \ref{thm:mhn_theory}, we know that $n_h \leq e^{c_0 n_v}$, meaning that $n_v \geq \frac{1}{c_0}\log(n_h)$. Then, $\sqrt{k_v} \geq \sqrt{s_v n_v} \geq \sqrt{\frac{s_v}{c_0} \log(n_h)}$. Therefore, we in fact have $$\mathbb{E}\left[ \frac{1}{k_v} \tilde{x}^T x_a  \right] = s_v + \sqrt{\frac{2 c}{n_v} (1 - s_v) \log(n_h)} + O\left(\frac{\log\log(n_h)}{\log(n_h)}\right).$$
\end{proof}

Before proving the main theorem, we prove one further important proposition:

\begin{prop} \label{prop:xa_in_memory}
    Fix $\delta \in (0, 1)$. Suppose that when $x_{a,c}$ is queried to the MHN $M$, $x_a$ is currently stored in $M$. Then, provided that $$c > \frac{\log(1 - (1 - \delta)^{\frac{1}{n_h - 1}})}{k_v \log(s_v)},$$ it follows that $$\mathbb{E}\left[ \frac{1}{k_v} \tilde{x}^T x_a \right] > 1 - \delta.$$
\end{prop}

\begin{rem} \label{rem:cue_constraint_rem}
    In practice, the lower bound constraint on $c$ can be further refined by more careful analysis. However, we should note that for our purposes, this bound will suffice. Indeed, for the Original MHN described in the main paper, we have $n_h = 100, k_v = 100, s_v = 0.1$; if we took $\delta = 10^{-3}$, then applying the inequality yields the constraint $c > 0.05$. Therefore, for the experiments detailed in our paper (which use $c = 0.5$ (50\% cues) and $c = 1$ (100\% cues)), we may effectively assume that $\mathbb{E}\left[ \frac{1}{k_v} \tilde{x}^T x_a \right] = 1$ given that $x_a$ is still stored by $M$.
\end{rem}

\begin{proof}
    First, observe that if $x_a$ is still stored by $M$, then the only way for $\tilde{x}$ to not be $x_a$ is if there exists some other pattern $p$ in $M$ such that $x_{a, c}$ is a sub-pattern of $p$ and the network retrieves this other pattern. That is, $$\mathbb{E}\left[ \frac{1}{k_v} \tilde{x}^T x_a \right] = \mathbb{E}\left[ \frac{1}{k_v} \tilde{x}^T x_a \;\middle|\; \exists \text{ multiple super-patterns} \right] P(\exists \text{ multiple super-patterns}) + 1 \cdot P(\not\exists \text{ other super-patterns}).$$ If we let $\eta := P(\exists \text{ multiple super-patterns})$, then $P(\nexists \text{ other super-patterns}) = 1 - \eta$ and we have that $\mathbb{E}\left[ \frac{1}{k_v} \tilde{x}^T x_a \right] \geq 1 - \eta$. Now, we focus on bounding $\eta$ above. Notice first that  $P(\nexists \text{ other super-patterns})$ is the probability that each pattern $p$ among the $n_h - 1$ patterns stored by $M$ that are not $x_a$ satisfies $p^T x_{a, c} < c k_v$; in this case, the hidden neuron corresponding to $x_a$ will receive the highest activation and hence $x_a$ will be retrieved. Moreover, $$P(p^T x_{a,c} < c k_v) = 1 - P(p^T x_{a,c} = c k_v) = 1 - \frac{k_v}{n_v}\frac{k_v - 1}{n_v - 1}\cdots\frac{k_v - (c k_v - 1)}{n_v - (c k_v - 1)} \geq 1 - (\frac{k_v}{n_v})^{c k_v}.$$ Thus, $$1 - \eta = P(\nexists \text{ other super-patterns}) \geq \left( 1 - s_v^{c k_v} \right)^{n_h - 1},$$ so $$\eta \leq 1 - \left( 1 - s_v^{c k_v} \right)^{n_h - 1}.$$ Now, given that $$c > \frac{\log(1 - (1 - \delta)^{\frac{1}{n_h - 1}})}{k_v \log(s_v)},$$ rearranging gives us $$ck_v \log(s_v) < \log\left(1 - (1 - \delta)^{\frac{1}{n_h - 1}}\right)$$ $$\implies s_v^{c k_v} < 1 - (1 - \delta)^{\frac{1}{n_h - 1}}$$ $$\implies (1 - \delta)^{\frac{1}{n_h - 1}} < 1 - s_v^{c k_v}$$ $$\implies \delta > 1 - (1 - s_v^{c k_v})^{n_h - 1}.$$ Thus, given that $c$ is bounded below as specified, we have that $\eta < \delta$ and thus $$\mathbb{E}\left[ \frac{1}{k_v} \tilde{x}^T x_a \right] \geq 1 - \eta \geq 1 - \delta.$$
\end{proof}

Now, we are ready to prove our main result.

\begin{proof}[Proof of Theorem \ref{thm:mhn_theory}]
    Notice that, at test time, when $x_{a, c}$ is queried to the MHN $M$, the original pattern $x_a$ is either still in the network or it has been replaced by another memory. Thus, we have $$\mathbb{E}\left[ \frac{1}{k_v} \tilde{x}^T x_a  \right] = \mathbb{E}\left[ \frac{1}{k_v} \tilde{x}^T x_a  \;\middle|\; x_a \text{ in memory}\right] P(x_a \text{ in memory})$$ $$ \hspace{35mm} + \mathbb{E}\left[ \frac{1}{k_v} \tilde{x}^T x_a  \;\middle|\; x_a \text{ not in memory}\right] P(x_a \text{ not in memory}).$$ By fixing suitable $\delta$ close to $0$ (e.g., $\delta = 10^{-3}$, as in the case of our large-scale networks---see Remark \ref{rem:cue_constraint_rem}), the first expectation term above is bounded below by $1 - \delta$ and above by $1$ (using Proposition \ref{prop:xa_in_memory}). To obtain a precise error term (in the general case), suppose that $\delta = \frac{1}{n_h^{1 + \epsilon}}$, for any $\epsilon \in (0, 1)$. In this case, we have that $\frac{1}{n_h^2} \leq \delta \leq \frac{1}{n_h}$, so $$1 - (1 - \delta)^{\frac{1}{n_h - 1}} = 1 - \left( 1 - \frac{1}{n_h - 1} \delta + \frac{\frac{1}{n_h - 1} \left(\frac{1}{n_h - 1} - 1\right)}{2!} \delta^2 - \dots \right)$$ $$\hspace{4mm} \geq \frac{1}{n_h - 1} \delta - \frac{\frac{1}{n_h - 1} }{2!} \left( \delta^2 + \delta^3 + \dots \right)$$ $$\hspace{8mm} \geq \frac{1}{(n_h - 1)n_h^2} - \frac{\frac{1}{n_h - 1} }{2!} \left( \frac{1}{n_h(n_h - 1)} \right)$$ $$\hspace{35mm} \geq \frac{1}{(n_h - 1)n_h^2} - \frac12 \frac{1}{(n_h - 1)^2 n_h} = \frac{n_h - 2}{2n_h^2 (n_h - 1)^2} > \frac{1}{4n_h^3},$$ where we have assumed that $\frac{n_h - 2}{n_h - 1} > \frac12$. Consequently, we have that $$\frac{\log(1 - (1 - \delta)^{\frac{1}{n_h - 1}})}{k_v \log(s_v)} \leq \frac{\log(4n_h^3)}{-n_v s_v \log(s_v)} = \frac{1}{-s_v \log(s_v)} \frac{\log4 + 3\log(n_h)}{n_v}$$ $$\leq \frac{1}{s_v \log\left(\frac{1}{s_v}\right)} \left(\frac{\log4}{n_v} + 3c_0 \right) = o(1),$$ where we have used the assumption that $n_v$ is sufficiently large and that $n_h \leq e^{c_0 n_v}$ for some positive constant $c_0 << s_v \log\left( \frac{1}{s_v} \right)$. This indicates that for $\delta = \frac{1}{n_h^{1+\epsilon}}$, for any $\epsilon \in (0, 1)$, the term $$\theta := \frac{\log(1 - (1 - \delta)^{\frac{1}{n_h - 1}})}{k_v \log(s_v)}$$ which constrains $c$ from below is made to be sufficiently close to $0$; moreover, for this choice of $\delta$, we have $$\mathbb{E}\left[ \frac{1}{k_v} \tilde{x}^T x_a  \;\middle|\; x_a \text{ in memory}\right] = 1 + o\left( \frac{1}{n_h^{1+\epsilon}} \right)$$ for any $\epsilon \in (0,1)$.
    
    Furthermore, by Proposition \ref{prop:xa_not_in_memory}, both of the terms $\mathbb{E}\left[ \frac{1}{k_v} \tilde{x}^T x_a  \;\middle|\; x_a \text{ not in memory}\right]$ and $\mathbb{E}\left[ \frac{1}{k_v} f(x_{\infty, c})^T x_{\infty}\right]$ are well-approximated by $$s_v + \sqrt{\frac{2c}{n_v}\left( 1 - s_v \right) \log(n_h)} + O\left(\frac{\log\log(n_h)}{\log(n_h)}\right)$$ provided that $n_h$ is sufficiently large. Finally, because $M$ is a slot-based memory system, the probability that $x_a$ is still in the memory is the probability that the hidden neuron that represents $x_a$ has not since been replaced (for $a-1$ timesteps of learning). Thus, $$P(x_a \text{ in memory}) = \left(1 - \frac{1}{n_h}\right)^{a-1}$$ $$P(x_a \text{ not in memory}) = 1 - \left(1 - \frac{1}{n_h}\right)^{a-1}.$$ Putting all of our above facts together, we have that $$\text{R.D.}(a) = \mathbb{E}\left[ \frac{1}{k_v} \tilde{x}^T x_a  - \frac{1}{k_v} f(x_{\infty, c})^T x_{\infty}\right]
    = \mathbb{E}\left[ \frac{1}{k_v} \tilde{x}^T x_a \right] - \mathbb{E}\left[ \frac{1}{k_v} f(x_{\infty, c})^T x_{\infty}\right]$$ $$ = \left( 1 + o\left( \frac{1}{n_h^{1+\epsilon}}\right) \right) \left(1 - \frac{1}{n_h}\right)^{a-1} + \left(s_v + \sqrt{\frac{2c}{n_v}\left( 1 - s_v \right) \log(n_h)}  + O\left(\frac{\log\log(n_h)}{\log(n_h)}\right) \right) \left( 1 - \left(1 - \frac{1}{n_h}\right)^{a-1} \right)$$ $$ - \left(s_v + \sqrt{\frac{2c}{n_v}\left( 1 - s_v \right) \log(n_h)} + O\left(\frac{\log\log(n_h)}{\log(n_h)}\right)\right)$$ $$ = \left( 1 - s_v - \sqrt{\frac{2c}{n_v}\left( 1 - s_v \right) \log(n_h)} \right) \left(1 - \frac{1}{n_h}\right)^{a-1} + O\left(\frac{\log\log(n_h)}{\log(n_h)}\right).$$
\end{proof}

Theorem \ref{thm:mhn_theory} indeed shows that the MHN raw difference decay curves presented in Fig. \ref{fig:iid_rawdiffs} are expected to closely follow exponential decay equations of the form $\text{R.D.}(a) = C e^{\beta (a-1)}$, where $$C = 1 - s_v - \sqrt{\frac{2c}{n_v}\left( 1 - s_v \right) \log(n_h)}$$ $$\text{and } \beta = -\log\left( 1 - \frac{1}{n_h} \right).$$

\subsubsection{Analyzing the Effect of Raising Network Connectivity in the K-winner MHN}    \label{app:kwin_mhn_larger_fan_in}

\begin{figure}[t]
        \centering
        \includegraphics[width=\textwidth]{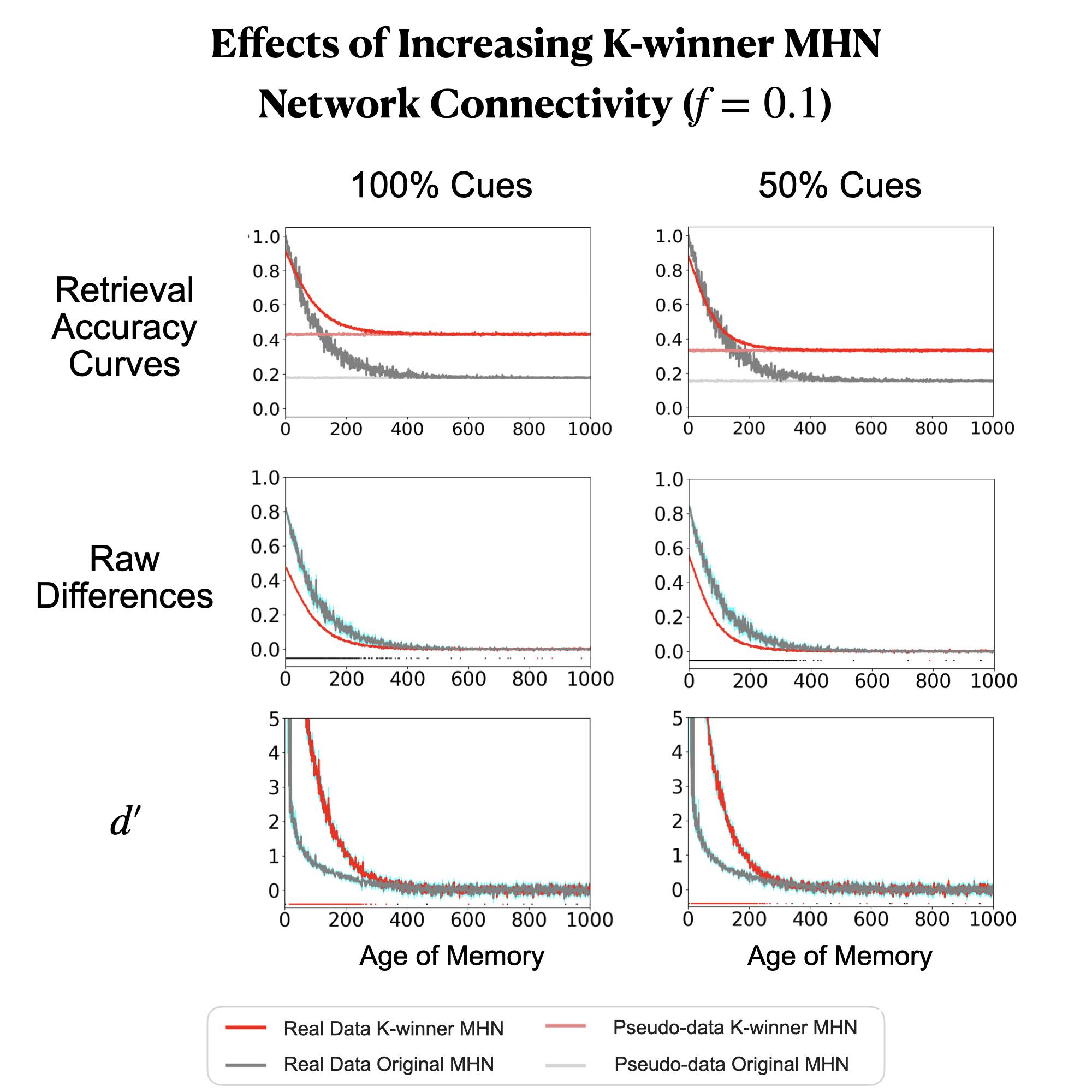}
        \caption{Comparison of retrieval performance between a K-winner MHN with slightly increased fan-in ($f$) and a parameter-matched original MHN. Raw retrieval accuracy curves (top), for both real data and untrained psuedo-patterns, are averaged over $100$ independent runs of either model. The raw difference and $d'$ measures were computed as averages over $10$ independent sample estimates, with each sample estimate itself being computed as an average over $20$ independent model runs. All three metrics are evaluated for in the case of full pattern retrieval (left panels) and pattern completion (right panels). Cyan: $d'$ standard error. Horizontal segments show ages where K-winner MHN $d'$ is higher (red) and MHN $d'$ is higher (black), with uncorrected $p < 0.01$.}
        \label{fig:kwinner_mhn_f=0.1_results}
    \end{figure}

To more fully present a comparison between possible instantiations of the K-winner MHN and the (baseline) original MHN, here we explore the effects of raising the fan-in ($f$) of the K-winner MHN, reasoning that this would raise the initial retrieval accuracy of the resulting model. The resulting K-winner MHN studied had the following hyperparameters: $\epsilon = 0.3$, $k_h = 50$, $f = 0.1$, $n_h = 1000$, $n_i = 1000$. Fig. \ref{fig:kwinner_mhn_f=0.1_results} shows the resulting retrieval accuracy curves, raw differences, and $d'$ measures for this new K-winner MHN compared against those of the same (parameter-matched) original MHN studied in the main text.

Fig. \ref{fig:kwinner_mhn_f=0.1_results} indeed shows that the raw retrieval accuracy of the K-winner MHN becomes almost perfect for the most recent memory, while also lowering the retrieval baseline for untrained psuedo-patterns (as compared to that of the K-winner MHN with $f = 0.05$ studied in the main text). This also contributes to a larger initial raw difference (R.D.) measure. However, the raw difference curves for this K-winner MHN consistently remain below those of the original MHN, indicating that the raw decay rate of information in this K-winner MHN is now larger than that of the K-winner MHN with $f = 0.05$. 

Furthermore, the $d'$ sensitivity curves (for both 100\% and 50\% cues) reveal that this new K-winner MHN still maintains a prominent advantage over the original MHN (in terms of recognition sensitivity) for a large swath of "intermediate-age" memories. In particular, while this K-winner MHN exhibits much larger $d'$ for a large class of memories that are not the most immediately recent, it no longer has a prominent advantage for especially old memories (e.g. of age $\sim$500). 

Just like our comparison of the K-winner MHN and the original MHN (in the main text) highlighted a trade-off between initial retrieval performance and retention of older memories, our analysis of the K-winner MHN with $f = 0.1$ again suggests the presence of such a trade-off. That is, while this new K-winner MHN demonstrates near-perfect initial retrieval accuracy, it exhibits a shorter-timescale advantage over the original MHN for older memories. Altogether, our combined set of results suggest that it is possible to design various K-winner MHNs that optimize one of these two objectives (maximizing small-age memory retrieval vs. better retaining older memories), but that it may be tricky (or otherwise infeasible) to simultaneously optimize both.

\subsubsection{Performance of MHN with Graded Weight Updates} \label{app:mhnlearningrate}


The theoretical analysis from Section \ref{app:theory_analysis} fully characterizes retrieval accuracy (and raw difference) for a special case of the K-winner network where $\epsilon = 1, f = 1, k_h = 1$. Our endeavor to analyze retention decay in the case of general $(\epsilon, f , k_h)$ can be seen as an endeavor to understand the parameter space of all possible K-winner MHN's. As a first step in this direction, one might ask how the original one-winner MHN performs when the weight updates do not correspond to slot-based replacements but rather graded adjustments (i.e., $\epsilon < 1$). 
Would a suitable sparse, distributed K-winner MHN (with $k_h > 1$) still retain its advantages over this modified instantiation of the MHN?

\begin{figure}[t]
    \begin{subfigure}[t]{0.01\textwidth}
    \textbf{A}
    \end{subfigure}
  \begin{subfigure}[t]{0.49\linewidth}
    \centering
    \includegraphics[width=\textwidth]{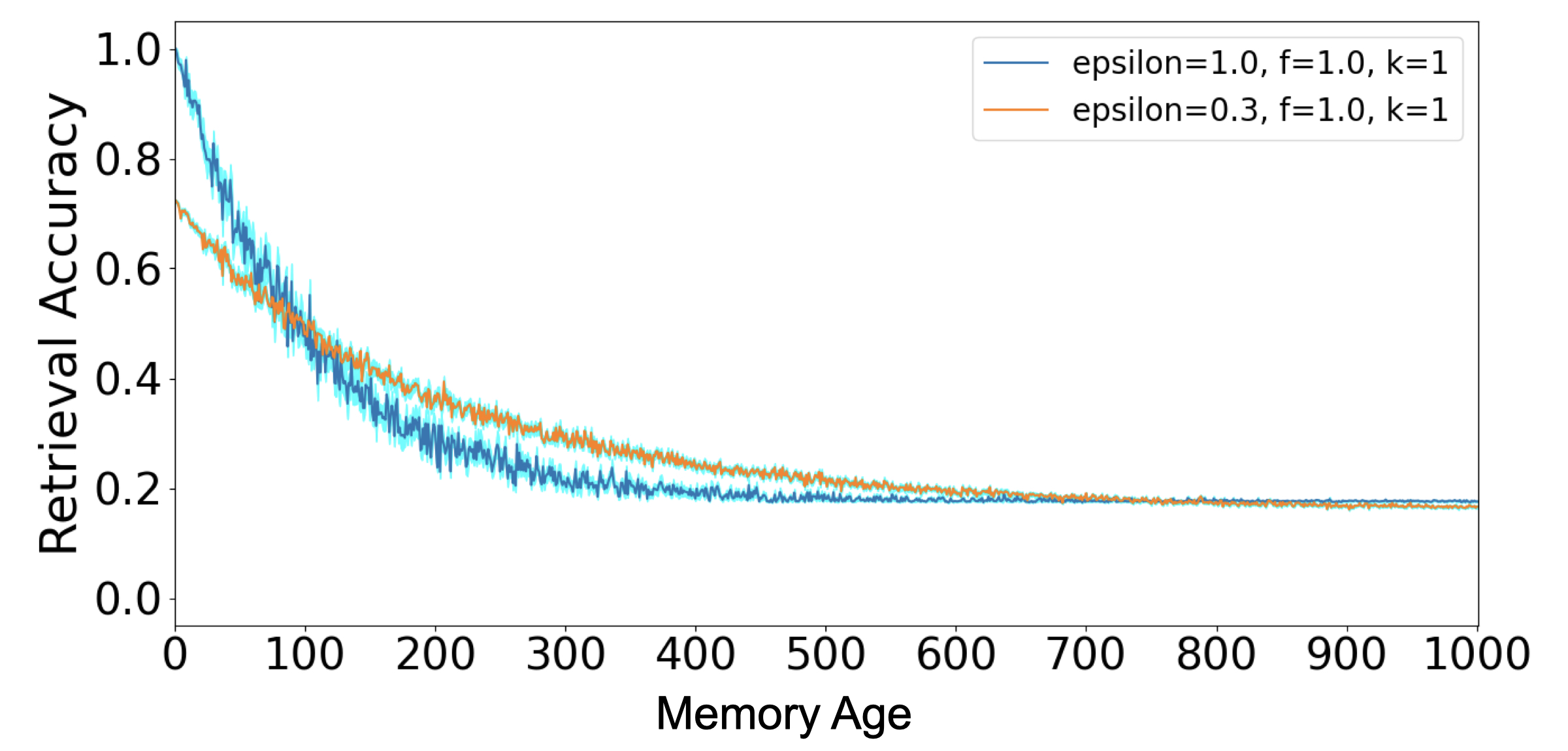}
    \label{fig:scaledup_lowerlr_accs}
  \end{subfigure}%
  \begin{subfigure}[t]{0.01\textwidth}
    \textbf{B}
    \end{subfigure}
  \begin{subfigure}[t]{0.5\linewidth}
    \centering
    \includegraphics[width=\textwidth]{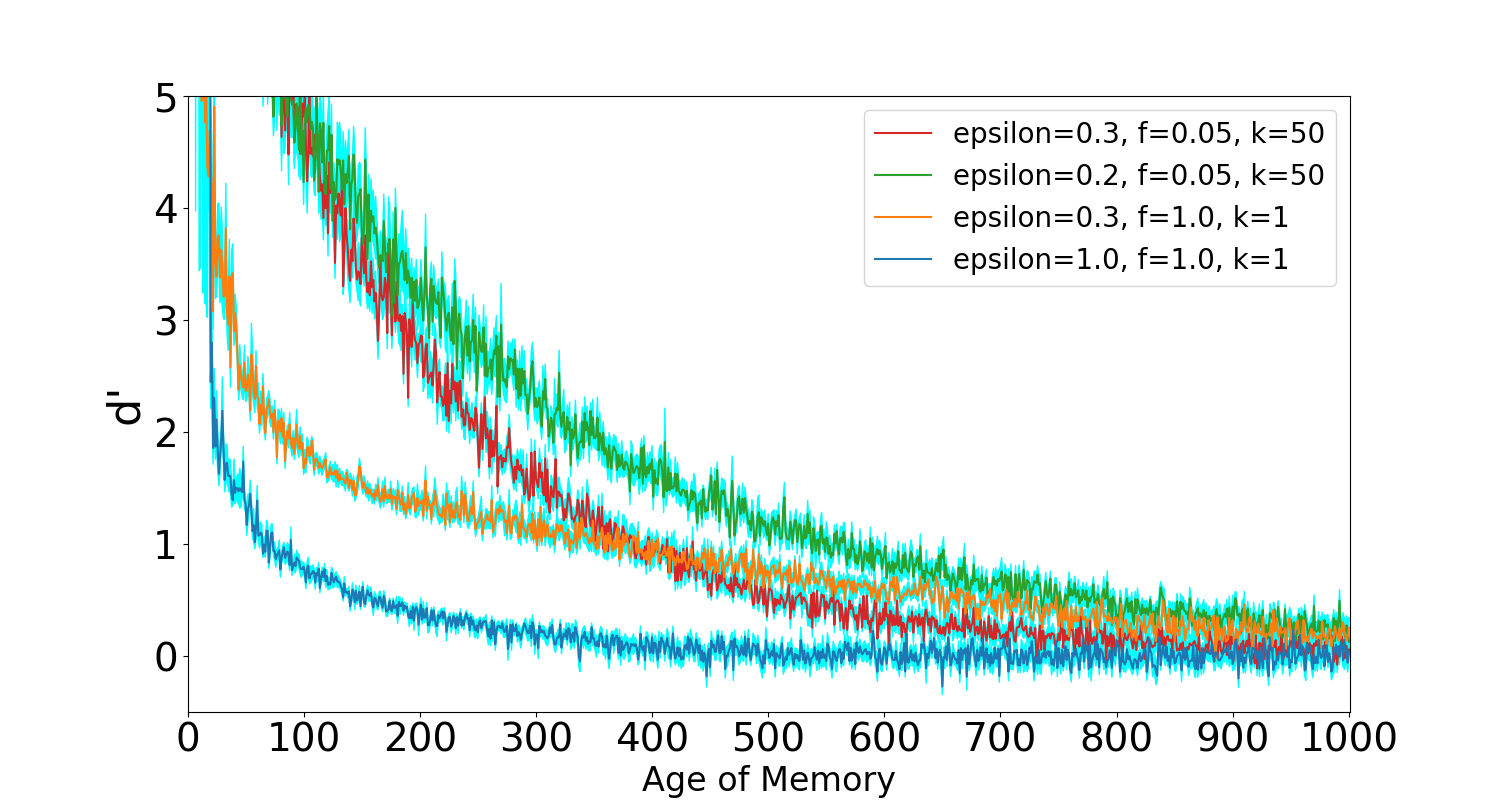}
    \label{fig:scaledup_lowerlr_metrics}
  \end{subfigure}
  \caption{\textbf{A} Raw retrieval accuracies for large-scale MHN's with $\epsilon = 1$ and $\epsilon = 0.3$, respectively, after training on random patterns, using 100\% cues during testing. \textbf{B} $d'$ measure for various K-winner MHN's (with the same total number of weights). Reducing $\epsilon$ from 0.3 (red) to 0.2 (green) enables the distributed K-winner MHN to outperform the graded MHN with $\epsilon = 0.3$ (shown in orange).}
  \label{fig:lowerlr_tests}
\end{figure}

To empirically probe this question, we first ran a test of raw retrieval accuracy for our large-scale MHN ($\epsilon = 1, f = 1, k_h = 1, n_v = 1000, n_h = 100$) compared against a graded version of this MHN in which $\epsilon$ is reduced to 0.3. Each of their respective retrieval accuracies, shown in Fig. \ref{fig:lowerlr_tests}A, is averaged over $10$ samples, where each sample itself was computed as an average over $20$ independent runs of the respective model (as described in Section \ref{subsubsec:dprime_details}). We also evaluated the $d'$ sensitivities for these networks, in addition to those for two (large-scale) distributed K-winner MHN's ($f = 0.05, k_h = 50, n_v = 1000, n_h = 2000$) in which the learning rates were set to $\epsilon = 0.3$ and $\epsilon = 0.2$, respectively (Fig. \ref{fig:lowerlr_tests}B).

Somewhat like the K-winner MHN, the graded MHN ($\epsilon = 0.3$) appears to possess lower initial retrieval accuracy but more graceful retention decay compared to the slot-based MHN ($\epsilon = 1$); interestingly, the graded MHN also demonstrates robust $d'$ sensitivity for older memories. In particular, it appears to maintain a higher $d'$ than the corresponding large-scale K-winner MHN (with $\epsilon = 0.3$) from roughly age 400 onwards. However, when the distributed K-winner MHN's learning rate is reduced from $\epsilon = 0.3$ to $\epsilon = 0.2$, its resulting $d'$ values appear to be higher than those of the graded MHN (Fig. \ref{fig:lowerlr_tests}B). This suggests that, for any given graded MHN with learning rate $\epsilon < 1$, we may be able to find a distributed K-winner MHN with learning rate $\epsilon' < 1$ that maintains a learning advantage over the graded MHN.

This preliminary exploration of the $(\epsilon, f, k_h)$-parameter space additionally suggests that decreasing $\epsilon$ reduces initial retrieval accuracy while also slowing down retention decay for older memories. More broadly, it highlights the need to theoretically characterize the $(\epsilon, f, k_h)$ parameter space, so as to understand whether there exist classes of "optimal" K-winner MHN's that have high accuracy baselines \emph{and} high learning ability past baseline, even for older memories.

\subsection{Details of simulations of memory for patterns with hierarchical similarity structure} \label{app:tgcrp_section}

\subsubsection{Tree-Generating Chinese Restaurant Process Algorithm} 

\begin{figure}[!h]
        \centering
        \includegraphics[width=\textwidth]{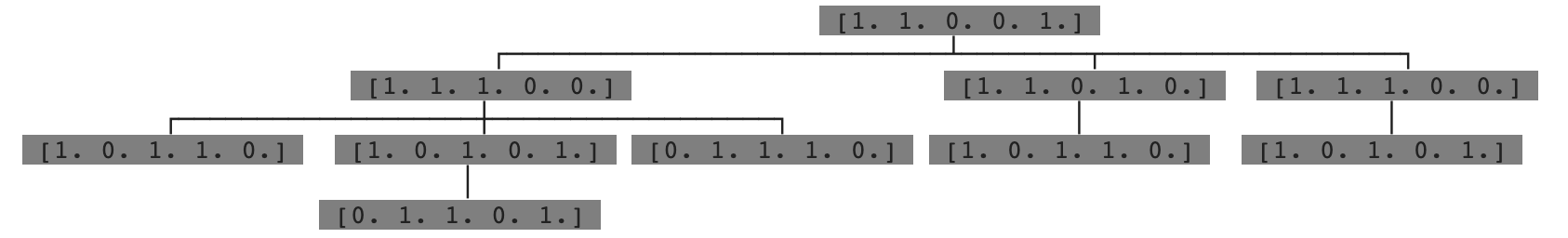}
        \caption{Sample tree generated by TGCRP consisting of $10$ patterns with $n_v = 5$, $s_v = 0.6$, $b = 1$.}
        \label{fig:sample_hierarchical_tree}
    \end{figure}

In this section, we illustrate how the Tree-Generating Chinese Restaurant Process Algorithm (TGCRP) works. The 'Chinese Restaurant Process' refers to an iterative (discrete) stochastic process for clustering a group of objects, typically into clusters with uneven frequencies; such a process could, for example, be useful in modeling objects whose frequencies follow a power law. Because objects in natural experience tend to have hierarchical similarity structure (e.g., 'living things' might branch out into 'animals' and 'plants', which could further branch out into 'dogs', 'cats', 'trees', 'bushes', etc.), we devise a modified variant of the Chinese Restaurant Process that is capable of probabilistically generating a tree of patterns; lower patterns in the tree may be thought of as specific instantiations of their higher-up ancestors in the tree. Moreover, we only take the leaf nodes of such a tree and use these as the patterns for training and testing. Fig. \ref{fig:sample_hierarchical_tree} illustrates one possible tree generated by this algorithm in a case of small pattern size. Detailed in Algorithm \ref{alg:TGCRP} (below) is the TGCRP algorithm. 

In considering the practical implementation of Algorithm \ref{alg:TGCRP}, a few important points should be noted:

\begin{enumerate}
    \item The slightly repetitive calculation of $P_i$ and $N_i$ over the course of generating the probabilistic tree appears to be onerous, but one can avoid such excessive computation by making them dynamically updating values that get updated after each iteration.

    \item This algorithm is used to generate a tree that contains 'num\_data' number of patterns. When running the TGCRP algorithm in practice, it should be noted that the number of leaf nodes of the tree is $\sim \frac12$ of 'num\_data'. 

    \item In each independent run of our retrieval $d'$ (and raw difference) experiments with TGCRP-generated data, we generated one big tree with 14,000 total nodes, meaning that there were roughly 7,000 leaf nodes. We shuffled these leaf nodes, and sampled 3,000 of them for bringing our models' respective weight distributions to equilibrium (steady state), 1,000 of them to be used as patterns for learning and retrieval, and 1,000 of them to be used as pseudo-memories. (Note that the $d'$ and raw difference measures themselves were calculated by averaging over many independent runs, each of which involved creating a new TGCRP tree.) 
\end{enumerate}

\begin{algorithm} 
\caption{Hierarchical Data-Generating Algorithm (TGCRP). 'num\_data' is the number of nodes in the entire tree. $b$ is the number of bit flips done when going from a parent node to a child node. Note that the function bit\_flipped$(p, b)$ returns a slight 'corruption' of pattern $p$ in which $b$ random $1$'s are set to $0$'s and $b$ random $0$'s are set to $1$'s. Additionally, let pattern$(Q)$ denote the pattern stored within the tree node $Q$.}\label{alg:TGCRP}
\begin{algorithmic}
\Require num\_data $\geq 1$, $1 \leq b \leq s_v n_v$
\State Make a root node $R_0$ for the tree, and initialize it with a random binary pattern of length $n_v$ and sparsity level $s_v$.
\For{$i$ in $1, \dots,$ num\_data - 1}
    \State Initialize $X$ to be the tree's root node.
    \State Create node $R_i$ but do not yet add it to the tree. 
        \While{node $R_i$ is not yet added to the tree}:
            \State Let $C_1, \dots, C_m$ be the children nodes of $X$ ($m \geq 0$). 
            \State $N_l \gets$ $1 + $ \#\{descendants of $C_l$\}for each $l = 1, \dots, m$.
            \For{$l$ in $1, \dots, m+1$}
                \If{$l \leq m$}
                    \State $P_l \gets \frac{N_l}{1 + \sum_{j=1}^{m} N_j}$
                \ElsIf{$l = m + 1$}
                    \State $P_l \gets \frac{1}{1 + \sum_{j=1}^{m} N_j}$
                \EndIf
            \EndFor

            \State Sample $d \in \{1, \dots, m+1\}$ using the categorical distribution $\text{Cat}([P_1, \dots P_{m+1}])$. 
            \If{$d = m+1$}
                \State Make node $R_i$ a new child node of $X$.
                \State Make node $R_i$ store the pattern bit\_flipped$(\text{pattern}(X), b)$.
            \Else
                \State $X \gets C_d$
            \EndIf
        \EndWhile
\EndFor
            
\end{algorithmic}
\end{algorithm}

\subsubsection{Similarity Structure of TGCRP-Generated Data}

\begin{figure}[!h]
        \centering
        \includegraphics[width=\textwidth]{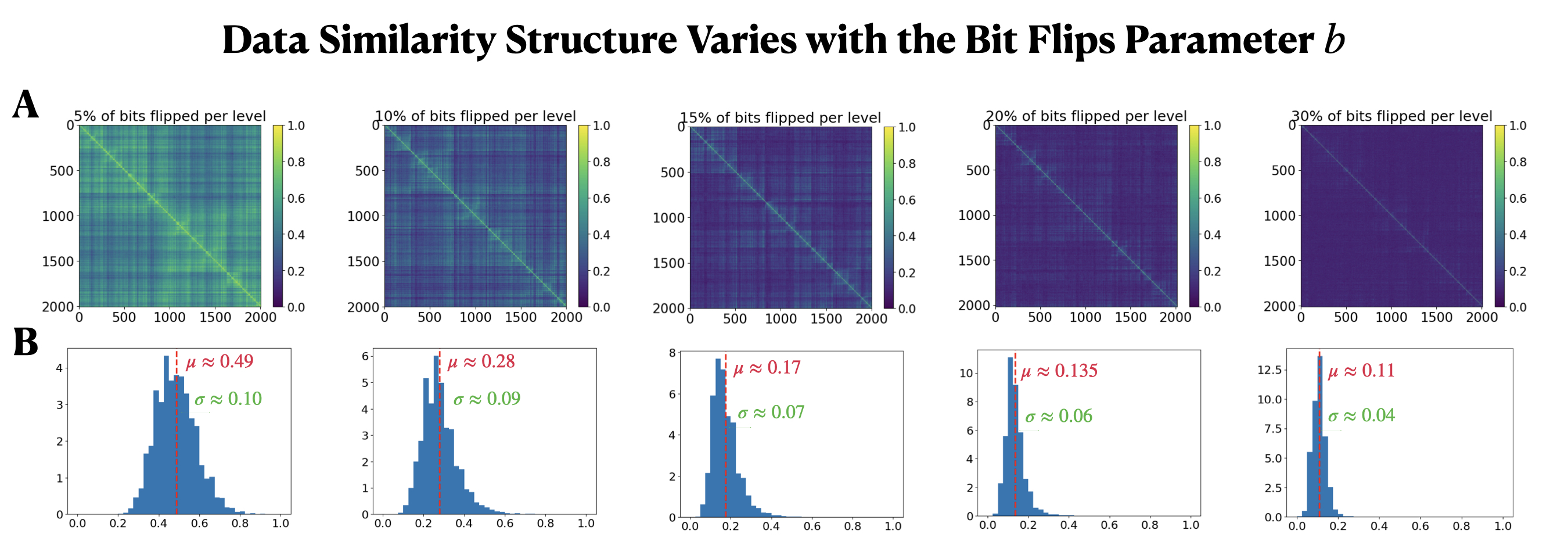}
        \caption{\textbf{A} Each matrix shown contains the pairwise dot product similarities for sample TGCRP-generated data. Matrices for TGCRP-generated data with $b = 5, 10, 15, 20, 30$ (from left to right) are shown. \textbf{B} Histograms depicting the upper-triangular entries in each of the matrices in \textbf{A}; as $b$ is increased (from left to right), the distribution of these entries tends to decrease to baseline. (For each similarity matrix-histogram pair, we generated a single tree with 4,000 nodes, using a pattern size of $n_v = 1000$ and $s_v = 0.1$.)}
        \label{fig:treedata_covar_hist}
    \end{figure}

In practice, we observe that applying the TGCRP algorithm while varying $b$ produces training data with varying similarity structure. This is reflected in the data similarity matrices for TGCRP-generated data (Fig. \ref{fig:treedata_covar_hist}), where we have used a pattern size of $n_v = 1000$ and sparsity level of $s_v = 0.1$. Observe that decreasing $b$ for a tree causes interspersed rectangular blocks of high similarity to appear, whereas increasing $b$ sufficiently removes off-diagonal similarity structure (effectively bringing us back to the case of random i.i.d. patterns).

\subsubsection{Testing retrieval sensitivity for structured patterns}

\begin{figure}[t]
  \centering
  \includegraphics[width=\textwidth]{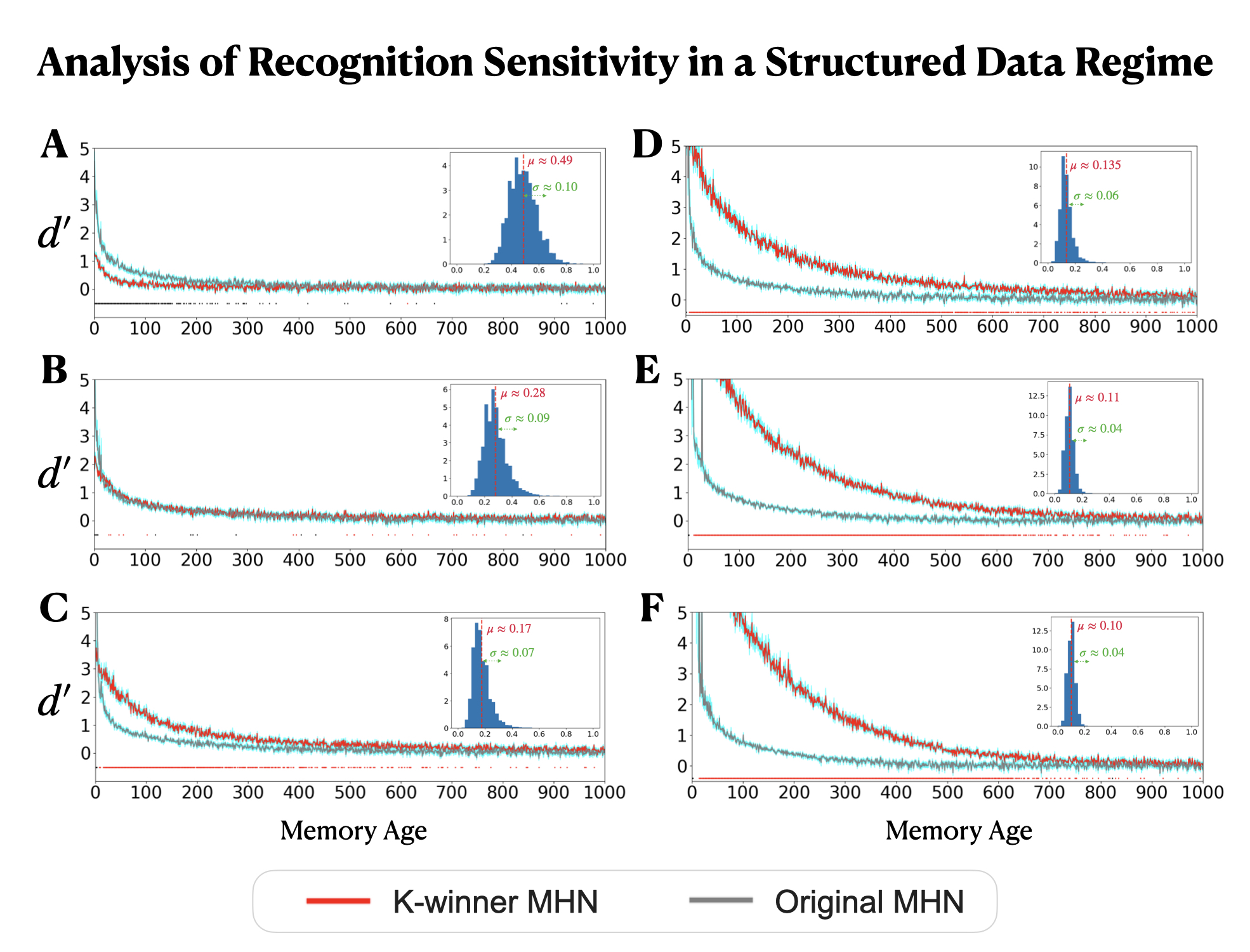}
  \label{fig:3B}
  \caption{
  Reconstruction sensitivity $d'$ as a function of memory age for our large-scale K-winner MHN and original MHN with structured patterns generated by the tree-generating Chinese restaurant process (TGCRP)
  using $100\%$ cues, under the values $5$, $10$, $15$, $20$, and $30$, of the TGCRP similarity parameter \textit{b}, corresponding to the number of bits flipped when generating a new child of a parent pattern (shown in \textbf{A}, \textbf{B}, \textbf{C}, \textbf{D}, and \textbf{E}, respectively). As a point of comparison, panel \textbf{F} illustrates the $d'$ curves in the case of unstructured data (same as Fig. \ref{fig:kwinnernet_retrieval_properties}B), corresponding to the limiting data distribution as $b$ becomes large. Within each $d'$ plot, the corresponding distribution of cosine similarities for $\sim$2,000 data samples generated by the respective process (TGCRP or unstructured) is shown. Cyan: $d'$ standard error. Horizontal segments show ages where K-winner MHN $d'$ is higher (red) and MHN $d'$ is higher (black), with uncorrected $p < 0.01$.}
  \label{fig:treedata_dprimes}
\end{figure}

In this section, we illustrate how the retrieval sensitivity of the K-winner MHN and the original MHN change over memory age, across varying levels of similarity structure in the patterns that are presented. Fig. \ref{fig:treedata_dprimes} illustrates these trends when performing retrieval using $100\%$ cues (i.e. full cues), in which trial-averaged $d'$ estimates are computed for each model type. Each $d'$ measurement is calculated relative to a baseline of untrained pseudo-patterns sampled from the same data-generating process. Furthermore, each $d'$ measurement is averaged over $10$ independent $d'$ samples, with each sample calculated using $20$ independent runs. Such $d'$ curves across memory age are shown for $5$ values of data similarity (Fig. \ref{fig:treedata_dprimes}A-E), as well as for the case of unstructured patterns (Fig. \ref{fig:treedata_dprimes}F). At a high level of data similarity, the K-winner MHN has no advantage in recognizing previously observed patterns relative to the original MHN. However, for nontrivial (but not high) levels of data similarity, the K-winner advantage for older memories readily emerges; this is evidenced by the case of patterns with a mean correlation of $\sim0.17$ (Fig. \ref{fig:treedata_dprimes}C).

\subsubsection{Model performance relative to an unstructured baseline} \label{app:treedata_unstructured_baseline}

\begin{figure}[!h]
        \centering
        \includegraphics[width=\textwidth]{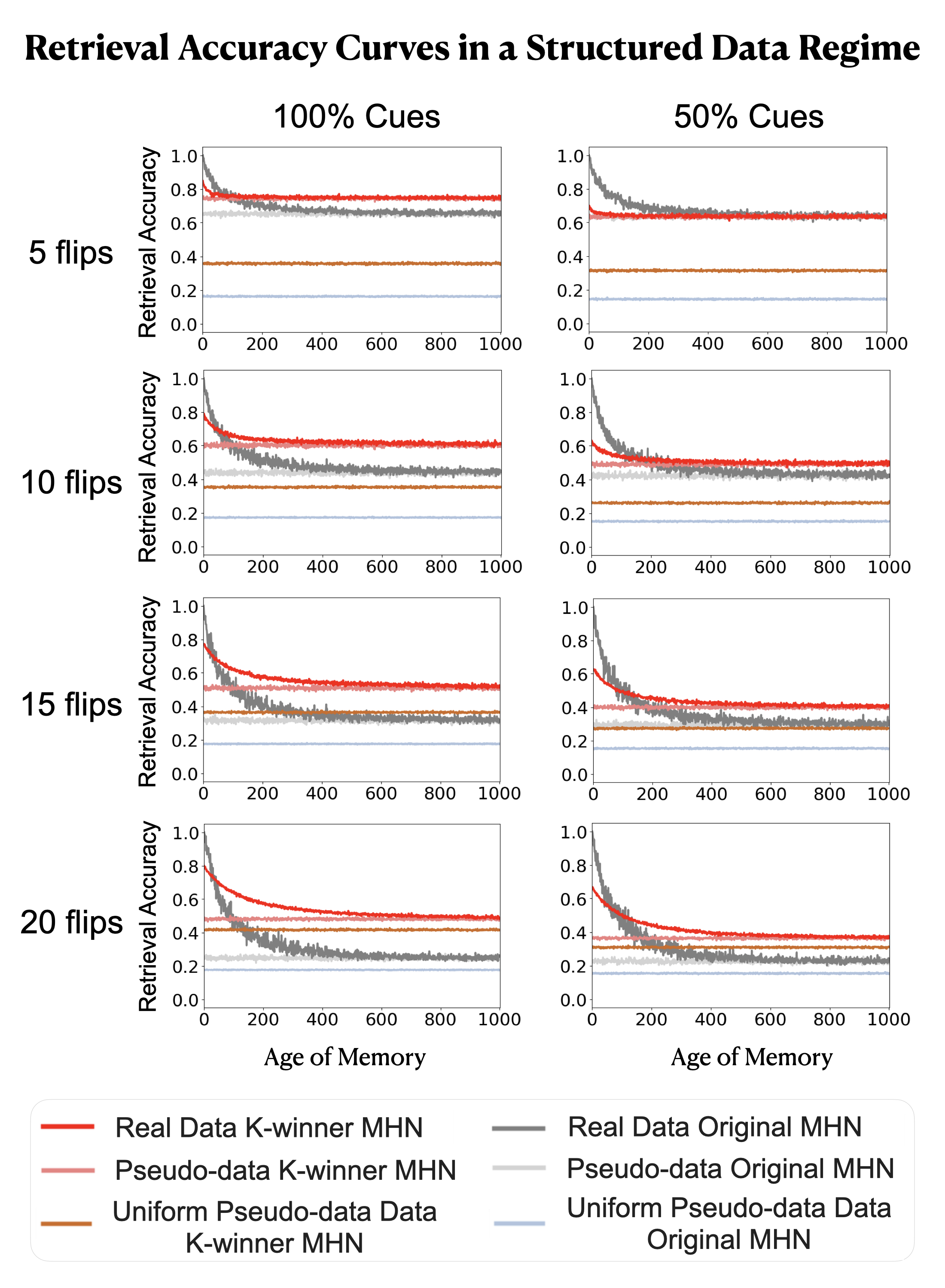}
        \caption{Retrieval accuracy curves (for various amounts of bit flips) showing reconstruction accuracy of a K-winner MHN and a parameter-matched original MHN, relative to their respective structured pseudo-data baselines \emph{and} their respective uniform random pseudo-pattern baselines. Curves shown are averaged over $100$ independent runs of each given model, and are shown in the cases of 100\% input cues (left) and 50\% input cues (right) during retrieval.}
        \label{fig:treedata_acc_curves_with_unif_baseline}
    \end{figure}

In this section, we show that both the K-winner MHN and the 1-winner MHN are able to retrieve memories well compared to a baseline of untrained pseudo-patterns sampled \emph{uniformly at random} (rather than also coming from the same TGCRP-generated distribution as the learned memories). For the K-winner MHN and original MHN analyzed in the main text, Fig. \ref{fig:treedata_acc_curves_with_unif_baseline} shows that both of these types of models are able to retrieve data from the TGCRP distribution to a greater level of accuracy than for random uniform pseudo-data, especially in the high data similarity regime (e.g. 5 or 10 bit flips). This indicates that both types of models have learned (collectively) the distribution of structured TGCRP patterns relative to an out-of-distribution baseline of random patterns.

\subsection{Preprocessing to reduce input pattern similarity using stacked K-winner MHNs}   \label{app:patsep}

A key principle from early work on distributed neural network models of memory by Marr \cite{Marr1971} was the idea that pre-processing of inputs to a memory system could reduce their overlap by using a conjunctive coding scheme, allowing more patterns to be stored in the memory with less interference.  To illustrate this idea with a simple example \cite{mcnaughton1987hippocampal}, consider two pattern vectors or length 6 with 3 active units each, but with 2/3 active elements in common. These patterns can be mapped to new vector of length 15, where each unit represents each of the possible conjunctions of two active input units.  In this representation, the patterns will only share 1 of 3 active elements in common.  While the number of $n$-wise conjuncts grows exponentially with the number of input elements, Marr proposed a coding scheme in which the units (called 'codons') to which the inputs are mapped each responds when $n$ or more of its $f$ connections are active. Because each codon now responds to many different conjunctions of order $n$, this ameliorates the combinatorial explosion problem. Related ideas were subsequently adopted in several proposed architectures \cite{oreilly1994hippocampal, Treves1992ComputationalCS}.  
In \cite{oreilly1994hippocampal}, it was observed that a K-winner-take-all layer with sparse connectivity performed similiary to a layer of Marr's codon units.  If one such K-winner layer with fixed symmetic binary-valued bi-directional connection weights with fan-in $f$ provides the input to a second, K-winner module with fast-changing bi-directionally symmetric connection weights as in our proposed K-winner MHN, this would allow input patterns with a high degree of similarity to be further differentiated prior to storage.  The fixed return connections between the pre-processing layer and the input layer would then allow the original highly overlapping patterns to be recovered, if needed for downstream processing.  A further refinement of this scheme would allow slow learning in the lower layer of this multi-layer system, using bi-directionally symmetric learning rules like the contrasive Hebbian learning rule used in the recirculation learning algorithm \cite{hinton1987recirculation}and in restricted Boltzmann machines \cite{hinton2012boltzmannmachine}.

\newpage

\section{The MHN-based transformer} \label{appendix_mhn_tf}

\subsection{Data Generation Details}

Synthetically generated data was used to train all models. Within any given input sequence, each input item (i.e. each context item and the query probe) is encoded as a one-hot vector of size $3L$. The first, middle, and last $L$ indices of this vector precisely correspond to lowercase, uppercase, and abstract probe versions of the letters, respectively, in their natural order. (For example, if $L = 2$, then the $6$ indices of an input vector $x \in \RR^{3L} = \RR^6$ correspond to 'a', 'b', 'A', 'B', 'ay', 'bee', in that order. Here, we have used the phonetic pronunciations of the letters 'a' and 'b' to represent their abstract, or case-agnostic, letter identities.) 

In particular, all networks were trained on a version of the Case Sequence Task in which $L = 4$ letters were used and the length of the context window was $C = 4$. Moreover, repeated occurrences of a letter (in either case) within the context window were not allowed. This meant that the context window of each input sequence contained one element from each of the sets $\{a, A\}$, $\{b, B\}$, $\{c, C\}$, and $\{d, D\}$, i.e., up to letter case, the four items in the context window were always a permutation of the first four letters of the alphabet. For each input sequence, the query probe was one of the four abstract letter queries in \{A/a (abstract query), B/b (abstract query), C/c (abstract query), D/d (abstract query)\}.

Moreover, each letter (and abstract query) was encoded with a specific one-hot vector of size $3L = 12$, where having a one in the positions [1, 2, 3, 4, 5, 6, 7, 8, 9, 10, 11, 12] corresponded to the letters [a, b, c, d, A, B, C, D, A/a (abstract query), B/b (abstract query), C/c (abstract query), D/d (abstract query)], respectively. Additionally, the target output for 'lowercase' and 'uppercase' were $[1, 0]^T$ and $[0, 1]^T$, respectively. When training models, a batch size of $B = 64$ was always used. All input sequences in each batch were uniformly generated at random (from the set of all valid case sequences).

\subsection{Model Implementation Details} \label{subsec:full_model_implementation_details}

In the subsequent section, we carefully illustrate the complete model implementation details for each of our three proposed MHN-based transformer architectures.

\subsubsection{Fixing the Matrix $\mathbf{W_K}$}

\begin{figure}[t]
        \centering
        \includegraphics[width=\textwidth]{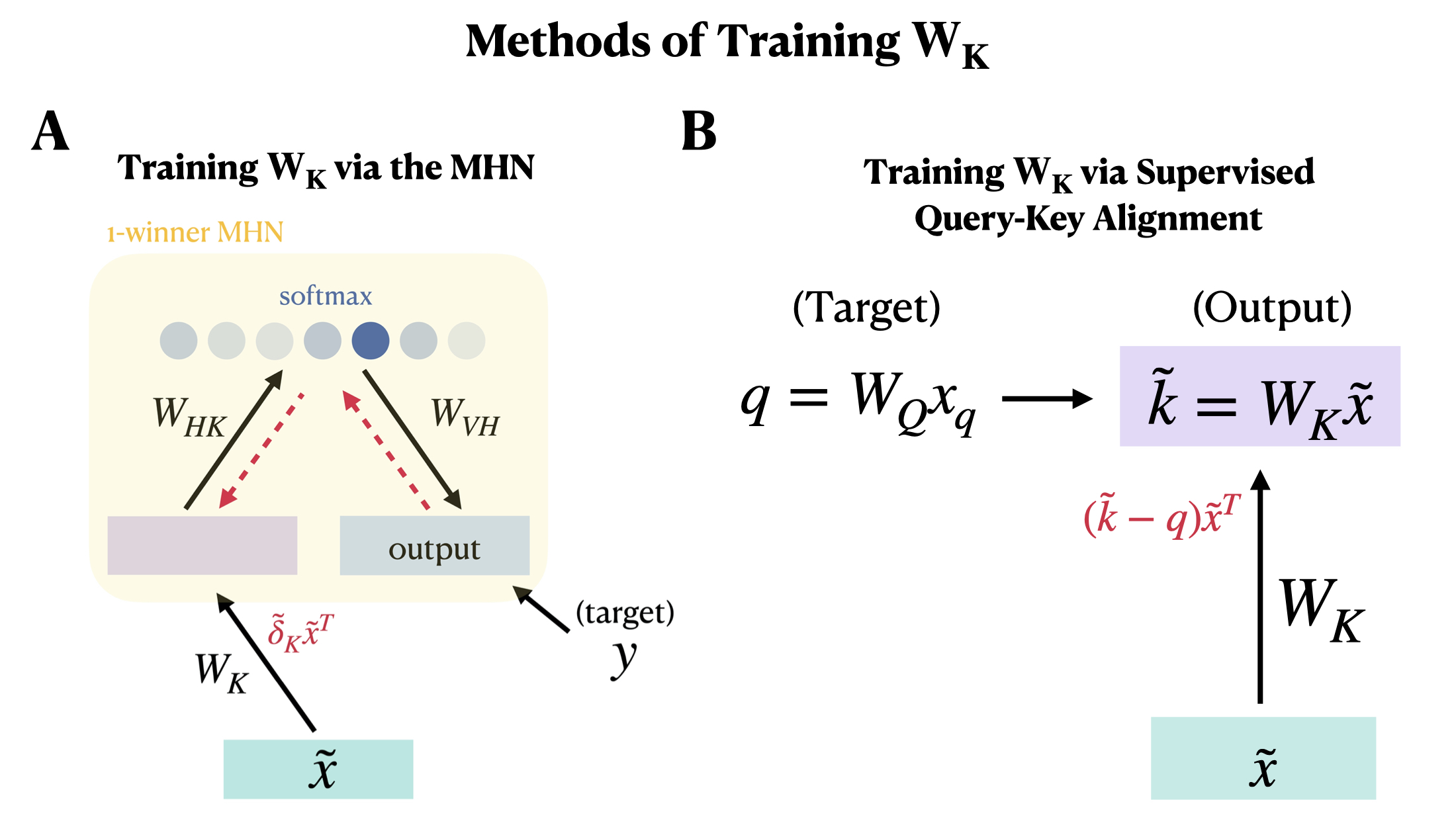}
        \caption{\textbf{A.} Training the matrix $W_K$ by using the loaded-up $1$-winner MHN. Once a reinstated vector $\tilde{x}$ is obtained, it is projected through $W_K$ to obtain a reinstated key. This key is then passed through the 1-winner MHN to produce an output, which is compared against the true target $y$ to produce a gradient signal used to train $W_K$. (Note that green arrows denote backpropagation of gradients.) \textbf{B.} Training $W_K$ by supervised query-key alignment. The reinstated vector $\tilde{x}$ is projected through $W_K$ to obtain the output $\tilde{k}$, which is compared against the target $q = W_Q x_q$ to produce a gradient signal used to update $W_K$.}
        \label{fig:training_W_K}
    \end{figure}

In our first instantiation of the MHN-based transformer, we focus on training $W_Q$ and $W_V$ while leaving $W_K$ fixed to a random initialization (explained further below). We begin by initializing the input-to-hidden and hidden-to-output weight matrices of a 1-winner MHN: $W_{HK}^{(0)} \in \RR^{n_h \times N}$ and $W_{VH}^{(0)} \in \RR^{2 \times n_h}$. We also initialize the "fast" weights $W_{IH} \in \RR^{D \times n_h}$, as well as the "slow" weights $W_Q$ and $W_V$ (see specific initialization and training details in Section \ref{subsubsec:training_details}).

For each time step $t = 1, 2, \dots, C$ in the context window, the one-hot input $x_t$ is used to obtain the key $k_t = W_K x_t$ and the associated value $v_t = W_V x_t$. Subsequently, the 1-winner-take-all attention vector $a^{(t)} = \sigma\left(W_{HK}^{(t-1)} k_t \right)$ is computed (where $\sigma$ denotes a hard 1-winner-take-all function). Then, each entry in the three matrices of "fast weights" is updated in a Hebbian fashion (Fig. \ref{fig:mhn_tf_fixed_WK}A): 
\begin{align}
    \left(W_{HK}^{(t)}\right)_{ij} &= \left(W_{HK}^{(t-1)}\right)_{ij} + \left( (k_t)_j - \left(W_{HK}^{(t-1)}\right)_{ij} \right) a_i^{(t)} \\
    \left(W_{VH}^{(t)}\right)_{li} &= \left(W_{VH}^{(t-1)}\right)_{li} + \left( a_i^{(t)} - \left(W_{VH}^{(t-1)}\right)_{li} \right) (v_t)_l \\
    \left(W_{IH}^{(t)}\right)_{di} &= \left(W_{IH}^{(t-1)}\right)_{di} + \left( (x_t)_d - \left(W_{IH}^{(t-1)}\right)_{di} \right) a_i^{(t)}. 
\end{align}

These updates precisely correspond to those used in a 1-winner MHN (where the learning rate $\epsilon = 1$). Finally, we set $W_{HK} = W_{HK}^{(C)}$, $W_{VH} = W_{VH}^{(C)}$, and $W_{IH} = W_{IH}^{(C)}$, to be the final weights obtained after iterating through the context window. Then, at the query timestep (Fig. \ref{fig:mhn_tf_fixed_WK}B), the query probe $x_t$ is linearly transformed to produce an embedded query vector $q = W_Q x_q$. Subsequently, we obtain the attention score vector $a_q = \text{softmax}\left( W_{HK} q \right)$ (noting that $W_{HK}$ now contains embedded key vectors for items from the context window). At this point, the buffer weight matrix $W_{IH}$ is used to reinstate a linear combination of input vectors (from the context window), weighted by the attention vector $a_q$: $\tilde{x} = W_{IH} a_q$. 

The vector $\tilde{x}$ may be viewed as a noisy version of the analogous vector $\sum_{i=1}^C \sm\left( x_q^T W_Q^T W_K x_i \right) x_i$ in a baseline transformer. Indeed, since the rows of $W_{HK} \in \RR^{n_h \times N}$ (where $n_h \geq C$) consist of keys obtained during the context window and also potentially un-used weights (i.e. rows that have never been used to store a key), $W_{HK}q \in \RR^{n_h}$ is a vector containing dot products between the query representation $q$ with keys from the context window but also entries that simply correspond to random noise. As a result, $a_q$ will be a slightly noisy version of the softmaxed attention vector obtained in a baseline transformer. Furthermore, computing $\tilde{x} = W_{IH}a_q$ will result in a softmax-weighted sum of previously seen items $x_t$ \emph{and} $n_h - C$ unfilled columns of $W_{IH}$ that effectively constitute noise. (In particular, in the case where $n_h = C$ and memory storage occurs with perfect fidelity, these computations coincide perfectly with those of the baseline transformer.)

Finally, the output $\hat{y} = W_V \tilde{x}$ is produced and compared against the ground truth target $y$ to produce a training signal that is used to update $W_Q$ and $W_V$ via backpropagation, using the Sum-Squared Error (SSE) loss function (for precise training details, see Section \ref{subsubsec:training_details}). The feedforward process undergone during the query timestep is temporally self-contained, so gradients can be backpropagated to $W_Q$ and $W_V$ without explicitly utilizing information from item representations at previous time steps (i.e. during the context window). 

As a first pass, we chose to fix $W_K$ for the sake of simplicity and to observe whether such a simplified model could solve the task; we call this model the Fixed $W_K$-MHN-Transformer. We reasoned that such a model could indeed solve the task, given this model's similarity to the baseline transformer with fixed $W_K$. Indeed, the output of the former model is 
\begin{equation}
    \hat{y} = W_V \tilde{x} = W_V W_{IH} \text{softmax}\left( W_{HK} W_Q x_q \right)
\end{equation}
where $W_V W_{IH}$ closely mirrors the matrix $V$ of values (defined in Section \ref{subsec:tf_and_goals} for the baseline transformer), since $W_{IH}$ has stored items $x_t$ from the context window within its columns; moreover, $W_{HK}W_Q$ closely mirrors $W_K^T W_Q$, given that $W_{HK}$ has encoded the keys $k_t = W_K x_t$ (for $t = 1, \dots, C$) within its rows. Noting furthermore that weight matrices feature in two parts of the baseline transformer's forward computation (Eq. \ref{tf_equation})---namely the '$W_V$' term and the '$W_Q^T W_K$' term---it is theoretically possible for $W_Q^T W_K$ to develop the same structure regardless of whether $W_K$ is fixed, because $W_Q$ can learn. Similarly, we expected $W_V$ to successfully learn, even when $W_K$ was held fixed. In addition to evaluating task accuracy, we wished to investigate whether representations of letter case would successfully emerge within $W_V$ (as in Fig. \ref{fig:baseline_tf_results}A) and whether the sub-blocks of $W_Q^T W_K$ in the MHN-based transformer with fixed $W_K$ would develop the same diagonal structure exhibited by the baseline transformer (Fig. \ref{fig:baseline_tf_results}D).





\subsubsection{Allowing $\mathbf{W_K}$ to Learn via the MHN} \label{subsubsec:W_K_learns}

While $W_K$ was set to be a fixed random matrix in the previous section, we now incorporate a mechanism for training $W_K$ as well (with $W_Q$ and $W_V$ being trained as described in Fig. \ref{fig:mhn_tf_fixed_WK}); we call this model the MHN-Transformer. Specifically, we ask how well the reinstated input vector $\tilde{x}$ is capable of predicting the target output $y$, through a feedforward process that involves $W_K$. As described in Fig. \ref{fig:training_W_K}A, we perform the feedforward computation 
\begin{equation}
    \hat{y}_K = W_{VH} \text{softmax}\left( W_{HK} W_K \tilde{x} \right)
\end{equation}
where $W_{HK}$ and $W_{VH}$ are the learned "fast weights" of the 1-winner MHN and $\hat{y}_K$ denotes the output produced by means of $W_K$. Consequently, we use the Sum-Squared-Error (SSE) loss between $\hat{y}_K$ and the ground truth label $y$ (and sum the squared errors over a batch) in order to compute the necessary gradient for $W_K$ (for precise training details, see Section \ref{subsubsec:training_details}). Note that gradients are only backpropagated from the output $\hat{y}$ to the reinstated input $\tilde{x}$ in order to update $W_K$.


\subsubsection{Allowing $\mathbf{W_K}$ to Learn via Query-Key Alignment}


In the previous section (Section \ref{subsubsec:W_K_learns}), our proposed model involved $W_K$ being adjusted via separately from the process used to adjust $W_Q$ and $W_V$---namely, by passing a reinstated key vector $\tilde{k} = W_K \tilde{x}$ through the $1$-winner MHN (Fig. \ref{fig:training_W_K}A). In this section, we present an alternate mechanism by which the matrix $W_K$ may be trained: supervised alignment with the query vector $q = W_Q x_q$. During the query timestep, we use the query vector $x_q$ to produce a reinstated vector $\tilde{x} = W_{IH} \text{softmax}\left( W_{HK} W_Q x_q \right)$ (as illustrated in Fig. \ref{fig:mhn_tf_fixed_WK}B). In this version of the MHN-based transformer, the matrix $W_K$ projects the reconstructed input $\tilde{x}$ to the same layer of neurons that constitutes the input to the $1$-winner MHN (which holds the representation $q$). Accordingly, we utilize a \emph{supervised delta rule} to locally update the weights in $W_K$, using $q$ as a supervisory signal: 
\begin{equation}    \label{eq:QK_learning_rule}
    \Delta W_K \propto (W_K \tilde{x} - q) \tilde{x}^T
\end{equation}

A visual description of this supervised alignment procedure is shown in Fig. \ref{fig:training_W_K}B; we call the associated model the QK-MHN-Transformer. This update precisely corresponds to the gradient arising from the squared-error loss between $W_K \tilde{x}$ and $q$. It also constitutes a biologically relevant learning rule, similar to Hebbian learning in that the update to each weight is local (i.e., a function of the sending neuron, receiving neuron, and the ground truth target activation). It should be noted that such a learning rule is suitable for "cue-based recall" task settings in which keys and queries must be selectively aligned with one another, so that the output for a given query is computed based on the output(s) of the most similar key(s) in memory. For tasks that do not possess this structure, other learning rules may need to be considered.

\subsubsection{Adding Input Projections to the MHN} \label{subsubsec:W_HI_details}

It is possible to augment each of our proposed MHN-based models with input projections by initializing a matrix $W_{HI} \in \RR^{n_h \times D}$ that contains all possible one-hot item vectors (collectively spanning all instances of uppercase and lowercase letters) within its rows (see Section \ref{subsubsec:training_details} for more details). Then, for an input $x_t$ presented at timestep $t$ of the context window, $x_t$ is projected through $W_K$ to obtain a key representation $k_t = W_K x_t$ (as before). However, when adding input projections, the vector of activations obtained at the hidden layer of the hetero-associative 1-winner MHN not only depend on $k_t$ but also on additional direct input from $x_t$ (Fig. \ref{fig:mhn_tf_fixed_WK}A): 
\begin{equation}
    a^{(t)} = \sigma \left( W_{HK}^{(t-1)} k_t + W_{HI} x_t \right).
\end{equation}
Here, $\sigma$ again denotes the 1-winner-take-all nonlinearity. The upshot of this construction is that, by design of $W_{HI}$, the vector $W_{HI} x_t$ is a one-hot vector, where the '1' appears in the index corresponding to the row of $W_{HI}$ that reads out the vector $x_t$. Because the context items $x_1, \dots, x_C$ are pairwise distinct, having input projections makes it more likely for the "winning" hidden neuron of the 1-winner MHN (as given by $a^{(t)}$) to be different from those hidden neurons that "won" at previous timesteps, thus preventing overwriting of existing memories.


\subsubsection{Training Details} \label{subsubsec:training_details}

All models (baseline transformer, MHN-Transformer with fixed $W_K$, MHN-Transformer, and QK-MHN-Transformer)---with and without input projections---were trained using vanilla batch gradient descent for $5000$ iterations with $B = 64$ (as mentioned earlier). That is, at each training iteration we perform the update $W_Z \leftarrow W_Z - \eta \frac{\partial \mathcal{L}}{\partial W_Z}$ for each $Z \in \{Q, K, V\}$, given a fixed learning rate $\eta$ and loss function $\mathcal{L}$ (defined below; also see Section \ref{subsec:gradient_eqns}). At each iteration, gradients for $W_Q$, $W_K$, and $W_V$ were computed manually (see Section \ref{subsec:gradient_eqns}). Moreover, for any given model, a fixed learning rate was used throughout training. To streamline notation, we let $D$ denote the input size (so for $L = 4$, $D = 3L = 12$), $N$ denote the size of key and query vectors, and let $n_h$ denote the hidden size of the 1-winner MHN (for all non-baseline transformer models). We set the dimension of model outputs---and correspondingly, the size of value vectors---to be $O = 2$. 

At the start of training any given model, the matrices $W_Q$, $W_K$, an $W_V$ were randomly initialized. For the baseline transformer, we initialized each entry in $W_Q$ and $W_K$ to be sampled iid from $\mathcal{N}\left( 0, \frac{1}{4\sqrt{N}} \right)$ and each entry in $W_V$ to be sampled iid from $\text{Unif}\left( [0, 0.1) \right)$. For each of our proposed transformer variants, each entry in $W_Q$ and $W_K$ was sampled iid from $\mathcal{N}\left( 0, \frac{1}{\sqrt{N}} \right)$ and each entry in $W_V$ was sampled iid from $\text{Unif}\left( [0, 0.1) \right)$.

Moreover, for each of our MHN-based variants, we initialized the fast weights as follows. Each entry in $W_{VH}^{(0)}$ and $W_{IH}^{(0)}$ was sampled iid from $\text{Unif}\left( [0, 1) \right)$. For models that did not use input projections, each entry in $W_{HK}^{(0)}$ was sampled iid from $\text{Unif}\left( [0, 0.5) \right)$. For models that utilized input projections, we initialized each entry of $W_{HK}^{(0)}$ to $0$ and initialized $W_{HI} \in \RR^{n_h \times D}$ so that $W_{HI}[0:2L, 0:2L] = I_{2L \times 2L}$ (note that $D = 3L$) and the rest of its entries were $0$. To facilitate batch computation (with batch size $B$), a separate set of fast weights $W_{HK}^{(0)}$, $W_{VH}^{(0)}$, and $W_{IH}^{(0)}$ were initialized and used for each training sequence in each batch.

At each iteration of training a particular model, a batch of $B = 64$ inputs was first randomly generated. (Note that we performed training over \emph{newly} generated batches, as opposed to sampling batches from a \emph{fixed} dataset.) Each input sequence in the batch was propagated through the given network to obtain a set of $B = 64$ model outputs. The resulting accuracies (resp. losses) for these outputs were averaged to produce a measurement of batch accuracy (resp. batch loss) at that iteration. After these loss and accuracy measurements were computed, batch-averaged gradients for $W_Q$, $W_K$, and $W_V$ were (manually) computed with respect to this batch of input sequences and were subsequently used to perform one step of gradient descent. 

As a slight technical aside regarding our MHN-based models, the minibatch sum-squared error loss used for training $W_Q$ and $W_V$ is computed over the outputs $\hat{y}_{\text{train}}^{(b)} = W_V W_{IH} \text{softmax}\left( W_{HK} W_Q x_q^{(b)} \right) \in \RR^2$ with respect to the targets $y^{(b)} \in \RR^2$ (for $b = 1, 2, \dots, B$) in the given batch. On the other hand, the loss and accuracy measures (computed at each iteration of training) that we report in our results are computed using the MHN outputs $\hat{y}_{\text{val}}^{(b)} = W_{VH} \text{softmax}\left( W_{HK} W_Q x_q^{(b)} \right) \in \RR^2$ with respect to the targets $y^{(b)} \in \RR^2$ (for $b = 1, 2, \dots, B$) in the given batch. Concretely, the batch loss between outputs $\hat{y}^{(b)}$ and targets $y^{(b)}$ is given by 
\begin{equation}
    \mathcal{L}\left( \left( \hat{y}^{(b)}, y^{(b)} \right)_{b=1}^{B} \right) = \sum_{b=1}^{B} \norm{\hat{y}^{(b)} - y^{(b)}}_2^2
\end{equation}
where the $\hat{y}^{(b)}$ are either the outputs used for training or for reporting loss. Furthermore, to calculate model accuracy across a given batch, for each $b \in \{1, 2, \dots, B\}$ we define $$\alpha_b := \begin{cases} 
      1 & \arg\max_{o \in \{1, 2\}} \left( \hat{y}_{\text{val}}^{(b)} \right)_o = \arg\max_{o \in \{1, 2\}} \left( y^{(b)} \right)_o \\
      0 & \text{else} 
   \end{cases}$$ and correspondingly define the  accuracy of the model across the given batch to be $\frac1B \sum_{b=1}^{B} \alpha_b$.

Below, we specify the exact training hyperparameters used for each model during training:

\paragraph{Baseline Transformer.} Training was performed with a fixed learning rate of 1e-3 and using $N = 10$. At each iteration, gradients for $W_Q$, $W_K$, and $W_V$ were computed from the sum-squared error (SSE) loss between the model outputs and true target labels across a given batch. 

\paragraph{MHN-Transformer with Fixed $W_K$ (with and without input projections).} Training was performed with a fixed learning rate of 1e-3, $N = 50$, and $n_h = 10$. At each iteration, gradients for $W_Q$ and $W_V$ were computed from the summed squared error loss between the output $\hat{y} = W_V W_{IH} \text{softmax}\left( W_{HK} W_Q x_q \right)$ and the true target $y$ for each input sequence in the given batch.

\paragraph{MHN-Transformer (with and without input projections).} Training was performed with a fixed learning rate of 5e-3 (for all three weight matrices), $N = 50$, and $n_h = 16$. At each iteration, gradients for $W_Q$ and $W_V$ were computed from the summed squared error loss between the output $\hat{y} = W_V W_{IH} \text{softmax}\left( W_{HK} W_Q x_q \right)$ and the true target $y$ for each input sequence in the given batch. Moreover, at each iteration, the gradient for $W_K$ was computed from the summed squared error loss between the output $\hat{y}_K = W_{VH} \text{softmax}\left( W_{HK} W_K \tilde{x} \right)$ and the true target $y$ for each input sequence in the given batch. (Here, $\tilde{x}$ denotes the reinstated item vector defined in the main text.)

\paragraph{QK-MHN-Transformer (with and without input projections).} Training was performed with a fixed learning rate of 5e-3 for $W_Q$ and $W_V$, a fixed learning rate of 1e-4 for $W_K$, $N = 50$, and $n_h = 16$. At each iteration, gradients for $W_Q$ and $W_V$ were computed from the summed squared error loss between the output $\hat{y} = W_V W_{IH} \text{softmax}\left( W_{HK} W_Q x_q \right)$ and the true target $y$ for each input sequence in the given batch. Moreover, at each iteration, the gradient for $W_K$ was computed from the summed squared error loss between the output $\hat{y}_K = W_K \tilde{x}$ and the true target $q = W_Q x_q$ for each input sequence in the given batch. (Here, $\tilde{x}$ denotes the reinstated item vector defined in the main text.)

Hyperparameters reported above were selected as a result of preliminary experiments that involved testing each model across different hyperparameters.

\subsection{Supplemental Results}

\subsubsection{Probing Query, Key, and Value Structure for Single-Trial Experiments}

\begin{figure}[t]
        \centering
        \includegraphics[width=0.9\textwidth]{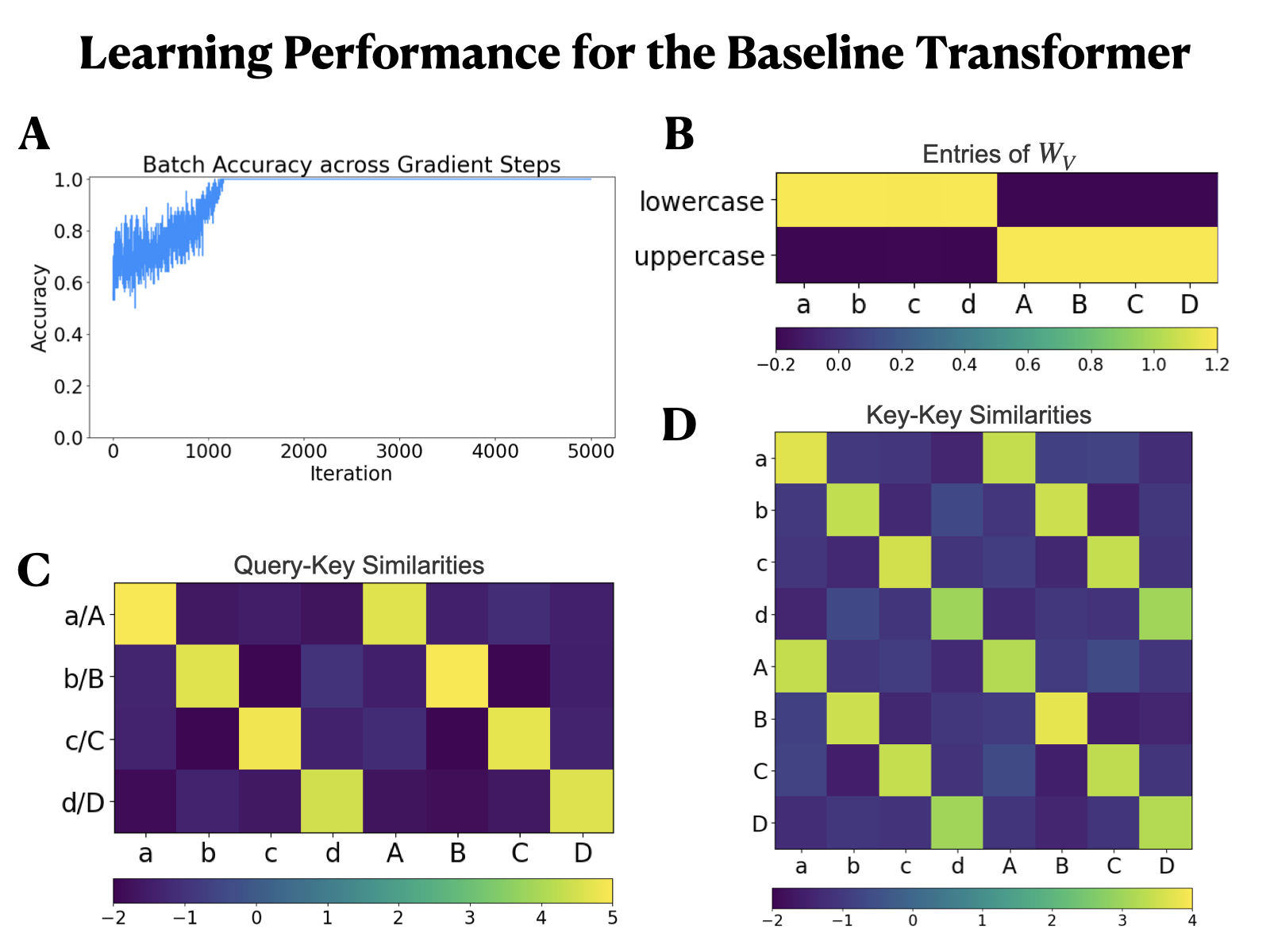}
        \caption{Results from a sample run of a baseline transformer model trained on the Case Sequence Task. \textbf{A.} Batch training accuracy over 5,000 gradient steps. \textbf{B.} A heatmap across the columns of $W_V$ corresponding to (non-query) letters. \textbf{C.} A heatmap of query-key alignment in $W_Q^T W_K$. \textbf{D.} A heatmap of key-key alignment in $W_K^T W_K$.} 
        \label{fig:baseline_tf_results}
    \end{figure}

\begin{figure}[t]
        \centering
        \includegraphics[width=0.9\textwidth]{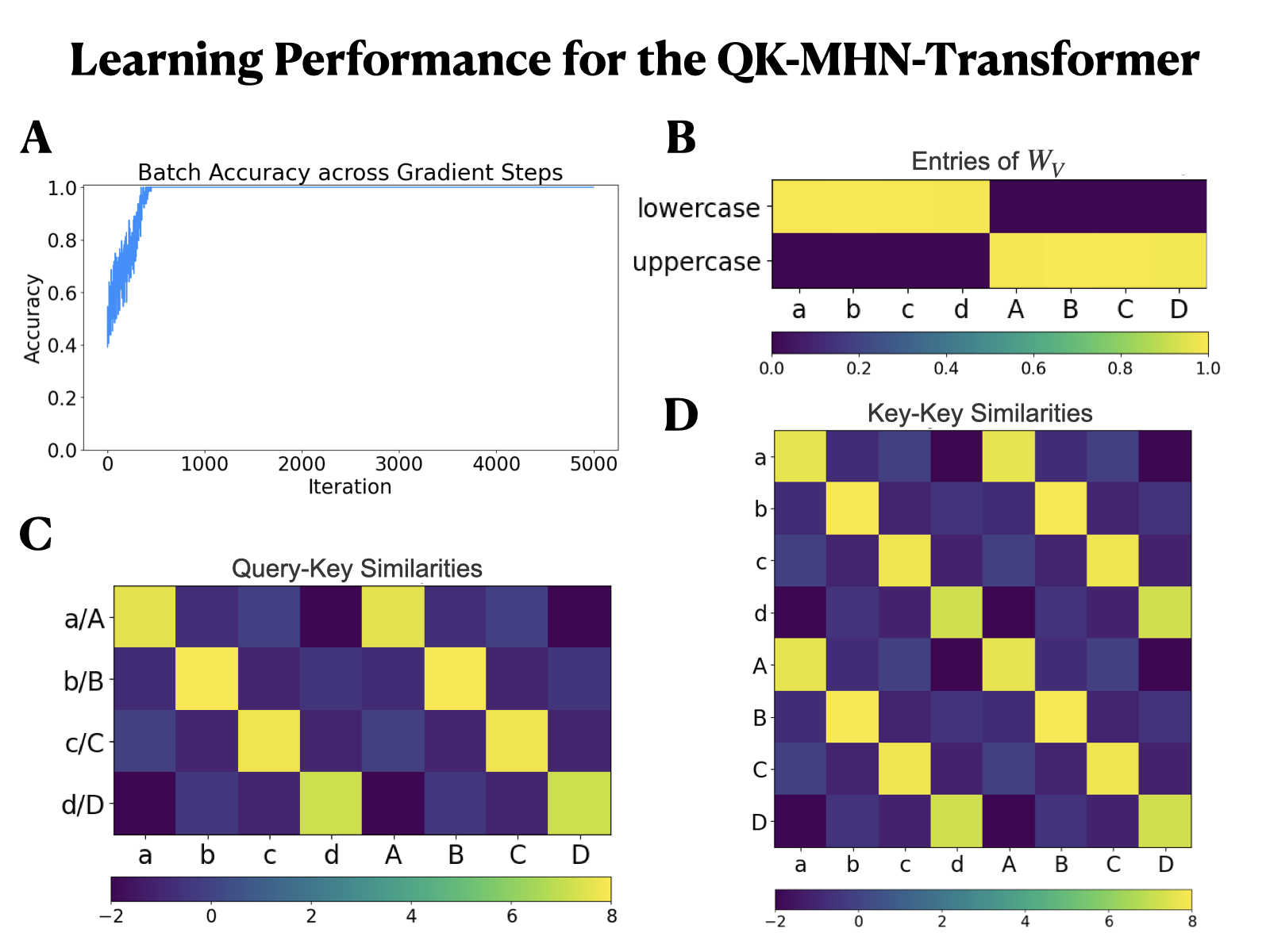}
        \caption{Single-trial results for the highest-performing MHN-based transformer model on the Case Sequence Task: the QK-MHN-Transformer (with input projections). \textbf{A.} Batch training accuracy over 5,000 gradient steps. \textbf{B.} A heatmap across the columns of $W_V$ corresponding to (non-query) letters. \textbf{C.} A heatmap of query-key alignment in $W_Q^T W_K$. \textbf{D.} A heatmap of key-key alignment in $W_K^T W_K$.}
        \label{fig:best_mhn_tf_model}
    \end{figure}

Here, we include the results of single-trial runs for the baseline transformer (Fig. \ref{fig:baseline_tf_results}) as well as for the QK-MHN-Transformer (Fig. \ref{fig:best_mhn_tf_model}), with $L = C = 4$. As highlighted in the main text, both models attain perfect accuracy during training. We also test for the semantically-relevant phenomena illustrated in Section \ref{subsec:tf_and_goals}. 
First, for either model, $W_V$ learns to represent the case types corresponding to the one-hot vectors for each specific letter (Fig. \ref{fig:baseline_tf_results}B and Fig. \ref{fig:best_mhn_tf_model}B). Second, analysis of the matrix $W_Q^T W_K$ indicates that the query vector for a given letter type aligns with both the lowercase and uppercase keys for that letter type and is anti-correlated with keys for other letter types (Fig. \ref{fig:baseline_tf_results}C and Fig. \ref{fig:best_mhn_tf_model}C). Finally, we find that the keys for lowercase and uppercase versions of any particular letter are aligned, whereas keys corresponding to different letter types are largely anti-correlated (Fig. \ref{fig:baseline_tf_results}D and Fig. \ref{fig:best_mhn_tf_model}D). Overall, this set of figures confirms that both the baseline transformer and the QK-MHN-Transformer exhibit the requisite query-key-value structure.  

\subsubsection{Trial-Aggregated Training Results}

In this section, we complement the results presented in Fig. \ref{fig:model_performances} of the main text with additional findings. As discussed in Fig. \ref{fig:model_performances}, the experiment used to generate these results consisted of performing $10$ independent training runs for each model type; for each training run for a given model, the learnable weights of the model were randomly initialized and trained as described in the previous section (Section \ref{subsubsec:training_details}). In addition to keeping track of batch accuracy across each run of each model, we also kept track of the mean-squared error loss as well as certain key statistics pertaining to the uppercase-lowercase submatrix of $W_K^T W_K$ (i.e. the kind of matrix presented in Figs. \ref{fig:baseline_tf_results}C and \ref{fig:best_mhn_tf_model}C), as well as the submatrices of $W_Q^T W_K$ describing the correlations between queries and lowercase keys and the correlations between queries and uppercase keys, respectively (i.e. the kind of matrices presented in the lower left of Figs. \ref{fig:baseline_tf_results}D and \ref{fig:best_mhn_tf_model}D). 

\begin{figure}[t]
        \centering
        \includegraphics[width=\textwidth]{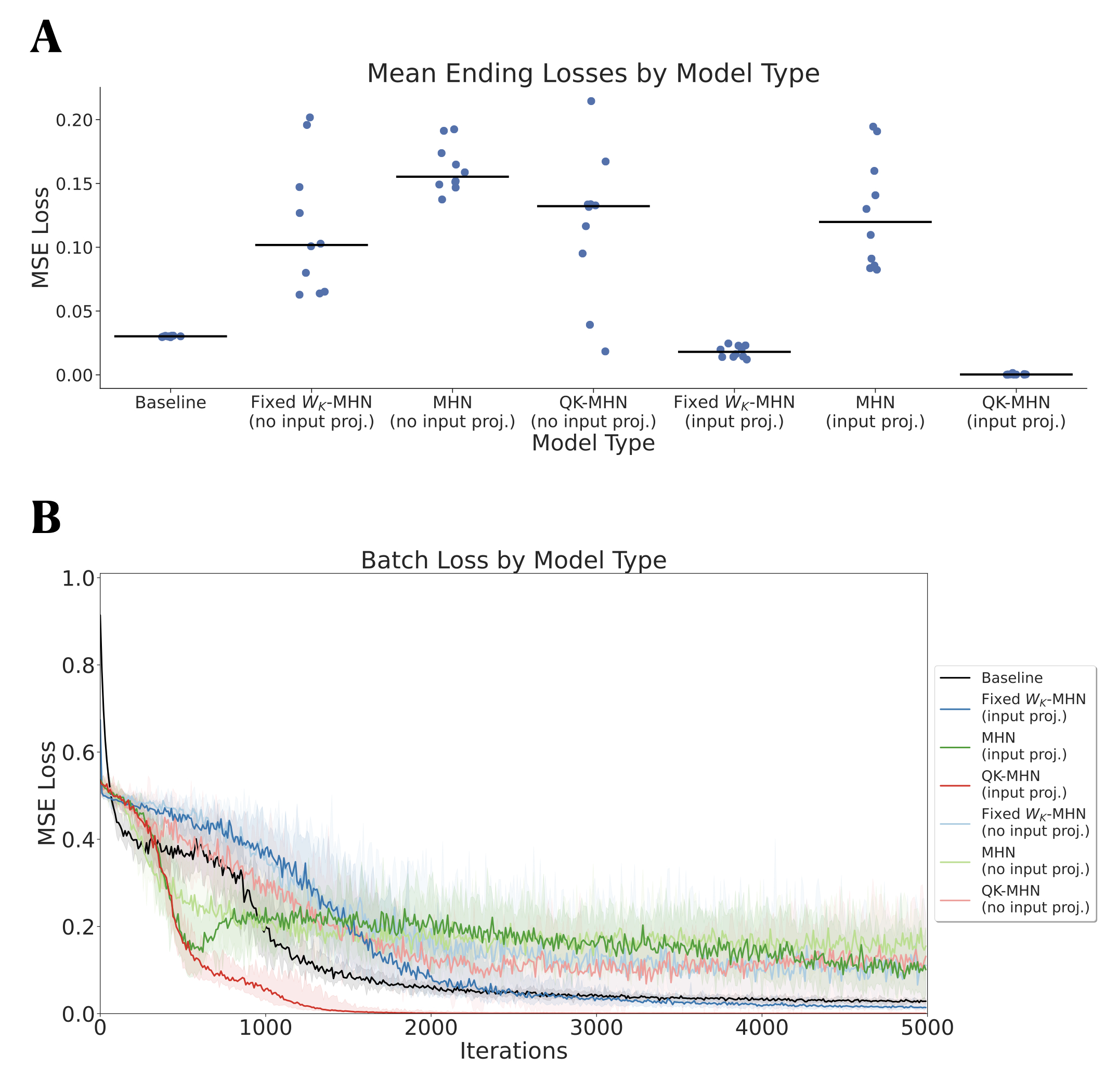}
        \caption{Comparisons of mean-squared error (MSE) loss on the Case Sequence Task across model types. \textbf{A.} Average final loss for $10$ independent runs of each model, where each measurement of final loss (for one run of a given model type) is averaged over the last 1,000 batches of training. Median losses for each model are shown via horizontal black bars.  \textbf{B.} Median loss curves shown across $10$ independent runs of each model, with shading to indicate the minimum and maximum losses attained by a given model at any particular iteration (across any of its $10$ independent runs).}
        \label{fig:model_loss_comparison}
    \end{figure}

First, we present a comparison of MSE losses across each model type (Fig. \ref{fig:model_loss_comparison}). Across each of the $10$ runs of every model, we computed the mean of the MSE losses over the last 1,000 batches of training to obtain a "mean ending loss"; we plot the mean ending losses obtained for each run of each model as a categorical plot in Fig. \ref{fig:model_loss_comparison}A. Consistent with our findings from Fig. \ref{fig:model_performances} of the main text, the baseline transformer, Fixed $W_K$ model (with input projections), and Q-K transformer (with input projections) show the best training performance, with the Q-K model consistently attaining the lowest loss. 

These trends are further highlighted in Fig. \ref{fig:model_loss_comparison}B, which presents the median MSE loss for each model throughout training: chiefly, the Q-K transformer demonstrates an ability to more quickly learn than the other models. Interestingly, the loss curves for both the Q-K model as well as the baseline transformer exhibit multiple separate periods of descent; we further explore this phenomenon in Section \ref{subsubsec:empirical_learning_dynamics}. Furthermore, while model in which $W_K$ is trained via the MHN (with input projections) initially exhibits a decrease in loss on par with that of the Q-K transformer, it subsequently shows a "U-shaped" profile. Given that the Fixed $W_K$ model (with input projections) demonstrates a smoothly decreasing loss curve, this suggests that the procedures for training $W_K$ and for training $W_Q$ and $W_V$ may be hampering one another. This phenomenon remains to be explored in future work.

Next, we analyze summary statistics for the uppercase key-lowercase key similarity matrix, the lowercase key-query similarity matrix, and the uppercase key-query similarity matrix; these are presented in Tables \ref{table:upper_lower_key_covar_table}, \ref{table:query_lower_key_covar_table}, and \ref{table:query_upper_key_covar_table}, respectively. For each model and any given one of these three similarity matrices, suppose that $M_1, \dots, M_{10} \in \RR^{L \times L}$ are the similarity matrices produced across the $10$ separate runs of that model. We first compute the mean $\mu_i$ of the diagonal elements for each $M_i$. Accordingly, the first two columns of each table presented below illustrates the mean and the range of the $\mu_i$'s, respectively. Similarly, we also compute the mean $\nu_i$ of the off-diagonal elements for each $M_i$; the third and fourth columns of each tables illustrate the mean and the range of the $\nu_i$'s, respectively. 

Given each of these three matrices is expected to show a clear diagonal structure for a model that successfully represents queries, keys, and values (which would reflect keys encoding letter identity and queries aligning with keys, as outlined in Section \ref{subsec:tf_and_goals} of the main text), the tables presented below provide a trial-averaged summary of the on-diagonal versus off-diagonal activity for each similarity matrix and each model type. Crucially, across Tables \ref{table:upper_lower_key_covar_table}, \ref{table:query_lower_key_covar_table}, and \ref{table:query_upper_key_covar_table}, we observe that each of the similarity matrices for both the baseline transformer and the Q-K transformer (both with and without input projections) show strongly positive on-diagonal activity relative to off-diagonal activity, illustrating that these models have successfully learned to represent keys and queries. While the Fixed $W_K$ models (with and without input projections) show alignment of queries with both lowercase and uppercase keys, we (predictably) observe that the uppercase key-lowercase key covariance shows no strong differences beween on- and off-diagonal activity, consistent with the fact that $W_K$ is randomly initialized in this case.

\begin{table}[!h]   
\begin{center}
\begin{tabular}{ |p{3cm}||p{2cm}|p{2cm}|p{2cm}|p{2cm}|  }
 \hline
 \multicolumn{5}{|c|}{Summary Statistics for Uppercase-Lowercase Key Covariance} \\
 \hline
 Model Type & On-Diagonal Mean & On-Diagonal Range & Off-Diagonal Mean & Off-Diagonal Range \vspace{2mm}\\
 \hline

 transformer & 3.34 & 0.323 & -1.11 & 0.693 \vspace{3mm}\\ 

 Fixed $W_K$ \newline(no input proj.) & 0.0563 & 0.280 & -0.00152 & 0.470 \vspace{6mm}\\ 
 
 Training $W_K$ \newline(no input proj.) & -0.0541 & 2.22 & -1.27 & 2.32 \vspace{6mm}\\ 
 
 Q-K Alignment\newline(no input proj.) & 7.19 & 2.07 & -1.42 & 2.48 \vspace{6mm}\\ 

 Fixed $W_K$ \newline(input proj.) & 0.00806 & 0.324 & 0.00531 & 0.482 \vspace{6mm}\\ 
 
 Training $W_K$ \newline(input proj.) & 0.532 & 1.57 & -0.933 & 2.02 \vspace{6mm}\\ 
 
 Q-K Alignment \newline (input proj.) & 7.96 & 1.88 & -0.916 & 2.16 \vspace{6mm}\\
 
 \hline
\end{tabular}
\caption{Mean and range---taken across $10$ separate runs of each model type---of averaged on-diagonal and  averaged off-diagonal entries in the lowercase-uppercase key similarity matrix (a submatrix of $W_K^T W_K$) for each model.}
\label{table:upper_lower_key_covar_table}
\end{center}
\end{table}

\begin{table}[!h]   
\begin{center}
\begin{tabular}{ |p{3cm}||p{2cm}|p{2cm}|p{2cm}|p{2cm}|  }
 \hline
 \multicolumn{5}{|c|}{Summary Statistics for Covariance between Queries and Lowercase Keys} \\
 \hline
 Model Type & On-Diagonal Mean & On-Diagonal Range & Off-Diagonal Mean & Off-Diagonal Range \vspace{2mm}\\
 \hline

 transformer & 4.73 & 0.246 & -1.57 & 1.11 \vspace{3mm}\\ 

 Fixed $W_K$ \newline(no input proj.) & 1.83 & 1.06 & -0.497 & 1.52 \vspace{6mm}\\ 
 
 Training $W_K$ \newline(no input proj.) & 4.06 & 1.39 & -0.126 & 2.62 \vspace{6mm}\\ 
 
 Q-K Alignment\newline(no input proj.) & 7.40 & 1.94 & -1.32 & 1.91 \vspace{6mm}\\ 

 Fixed $W_K$ \newline(input proj.) & 2.58 & 0.727 & -0.313 & 1.64 \vspace{6mm}\\ 
 
 Training $W_K$ \newline(input proj.) & 5.52 & 2.83 & 0.334 & 4.17 \vspace{6mm}\\ 
 
 Q-K Alignment \newline (input proj.) & 7.95 & 1.89 & -0.878 & 2.21 \vspace{6mm}\\
 
 \hline
\end{tabular}
\caption{Mean and range---taken across $10$ separate runs of each model type---of averaged on-diagonal and  averaged off-diagonal entries in the lowercase key-query similarity matrix (a submatrix of $W_Q^T W_K$) for each model.}
\label{table:query_lower_key_covar_table}
\end{center}
\end{table}

\begin{table}[!h]   
\begin{center}
\begin{tabular}{ |p{3cm}||p{2cm}|p{2cm}|p{2cm}|p{2cm}|  }
 \hline
 \multicolumn{5}{|c|}{Summary Statistics for Covariance between Queries and Uppercase Keys} \\
 \hline
 Model Type & On-Diagonal Mean & On-Diagonal Range & Off-Diagonal Mean & Off-Diagonal Range \vspace{2mm}\\
 \hline

 transformer & 4.78 & 0.307 & -1.59 & 1.01 \vspace{3mm}\\ 

 Fixed $W_K$ \newline(no input proj.) & 1.89 & 1.00 & -0.582 & 1.35 \vspace{6mm}\\ 
 
 Training $W_K$ \newline(no input proj.) & 4.00 & 1.91 & -0.0940 & 2.32 \vspace{6mm}\\ 
 
 Q-K Alignment\newline(no input proj.) & 7.41 & 1.92 & -1.33 & 1.98 \vspace{6mm}\\ 

 Fixed $W_K$ \newline(input proj.) & 2.90 & 0.547 & -0.351 & 1.76 \vspace{6mm}\\ 
 
 Training $W_K$ \newline(input proj.) & 6.12 & 1.10 & -0.944 & 4.14 \vspace{6mm}\\ 
 
 Q-K Alignment \newline (input proj.) & 7.95 & 1.89 & -0.879 & 2.21 \vspace{6mm}\\
 
 \hline
\end{tabular}
\caption{Mean and range---taken across $10$ separate runs of each model type---of averaged on-diagonal and  averaged off-diagonal entries in the uppercase key-query similarity matrix (a submatrix of $W_Q^T W_K$) for each model.}
\label{table:query_upper_key_covar_table}
\end{center}
\end{table}

\subsection{Gradient Equations} \label{subsec:gradient_eqns}

In this section, we present the (manually computed) gradients for $W_Q$, $W_K$, and $W_V$ that are used to update these weights in each model presented in the main text. Using these equations (listed below), we carry out further mathematical analyses in Section \ref{subsubsec:W_V_learns_first}. For clarity and ease of notation, we present gradient update equations for each model for a \emph{single} training sequence; we simply note that when training our models, gradients are summed for the training sequences across an entire mini-batch of size $B$. Presently, we consider a single input sequence $(x_1, x_2, \dots, x_C, x_q, y)$ consisting of the context item sequence, the query probe, and ground truth target. For this training sequence, we also define $s$ as the softmaxed representation obtained during the query time step. Concretely, for the baseline transformer, $s := \text{softmax}\left( \beta {x_q}^T W_Q^T W_K X \right)$ (where $\beta$ is usually taken to be $\frac{1}{\sqrt{N}}$), and for our MHN-based models, $s := \text{softmax}\left( W_{HK} W_Q x_q \right)$. Here, $X := \begin{bmatrix}
    \vert & \vert &  & \vert \\
    x_1  & x_2 & \dots & x_C   \\
    \vert & \vert &  & \vert 
\end{bmatrix}$.

\paragraph{Baseline Transformer.} Using the squared error loss $\mathcal{L} = ||\hat{y} - y ||_2^2$ (where $\hat{y}$ is given by Eq. \ref{tf_equation}), we obtain the gradient equations 
\begin{equation} \label{eq:tf_WV_gradient}
\begin{aligned}
    \frac{\partial \mathcal{L}}{\partial W_V} = 2 \sum_{i=1}^C s_i (\hat{y} - y) x_i^T = 2 \sum_{b=1}^{B} \left( \hat{y} - y \right) \tilde{x}^{T}  
\end{aligned}
\end{equation}
where $\tilde{x} = \sum_{i=1}^C s_i x_i$ (and where we may set $\delta_{V, i} := 2s_i (\hat{y} - y)$ to obtain the gradient outer products shown in Fig. \ref{fig:slotbased_backprop}A),

\begin{equation}
\begin{aligned}
    \frac{\partial \mathcal{L}}{\partial W_K} =  \sum_{i=1}^{C} {\delta_{K,i}} {x_i}^T 
\end{aligned}
\end{equation}
where ${\delta_{K,i}} = 2\beta s_i(\hat{y} - y)^T (-\hat{y} + W_V x_i) W_Q x_q$ (so that the $\delta_{K, i}$ give rise to the gradient outer products shown in Fig. \ref{fig:slotbased_backprop}A), and 

\begin{equation}
\begin{aligned}
    \frac{\partial \mathcal{L}}{\partial W_Q} = {\delta_{Q}} {x_q}^T = \sum_{i=1}^C {\theta_{Q, i}} W_K x_i{x_q}^T 
\end{aligned}
\end{equation}
where ${\delta_{Q}} = 2 \beta (\hat{y} - y)^T \sum_{i=1}^{C} s_i (-\hat{y} + W_V x_i) W_K x_i$ (as illustrated in Fig. \ref{fig:slotbased_backprop}) and where $\theta_{Q, i} = \frac{2}{\sqrt{N}}s_i (\hat{y} - y)^T (-\hat{y} + W_V x_i)$.

\paragraph{MHN-Transformer with Fixed $W_K$ (with and without input projections).} 

Using the squared error loss $\mathcal{L} = ||\hat{y} - y ||_2^2$ (where $\hat{y}$ is given by Eq. \ref{tf_equation}), we obtain the gradient equations 
\begin{equation} \label{eq:slotfree_tf_WV_gradient}
\begin{aligned}
    \frac{\partial \mathcal{L}}{\partial W_V} = 2 \left( \hat{y} - y \right) \tilde{x}^T 
\end{aligned}
\end{equation}
where $\tilde{x} = W_{IH} s$ and where we may define $\tilde{\delta}_V := 2(\hat{y} - y)$ to obtain the gradient outer product shown in Fig. \ref{fig:mhn_tf_fixed_WK}B, and 

\begin{equation}
\begin{aligned}
    \frac{\partial \mathcal{L}}{\partial W_Q} = {\tilde\delta_{Q}} {x_q}^T 
\end{aligned}
\end{equation}
where ${\tilde\delta_{Q}} = 2(\hat{y} - y)^T \sum_{m=1}^{n_h} s_m \left(-\hat{y} + W_V \left( W_{IH} \right)_{\cdot, m}\right) \left(W_{HK}\right)_m$ (as illustrated in Fig. \ref{fig:mhn_tf_fixed_WK}B).

Here, if $A \in \RR^{P \times Q}$ is an arbitrary matrix, we use the notation $A_l$ to denote the $l$th row of $A$ as a vector in $R^Q$ and $A_{\cdot, m}$ to denote the $m$th column of $A$ as a vector in $\RR^P$.

\paragraph{MHN-Transformer (with and without input projections).} 

For this model, the gradients for $W_V$ and $W_Q$ are precisely the same as those for the MHN-Transformer with fixed $W_K$ (above). Moreover, in this model, $W_K$ is trained using the loss $\mathcal{L}_K := ||\hat{y}_K - y ||_2^2$, where $\hat{y}_K = W_{VH}\text{softmax}\left( W_{HK} W_Q \tilde{x} \right)$, giving the gradient equation

\begin{equation}
\begin{aligned}
    \frac{\partial \mathcal{L}_K}{\partial W_K} = {\tilde\delta_{K}} \tilde{x}^T 
\end{aligned}
\end{equation}
where ${\tilde\delta_{K}} = 2\left(\hat{y}_K - y\right)^T \sum_{m=1}^{n_h} s_{K,m} \left(-\hat{y}_K + \left( W_{VH} \right)_{\cdot, m}\right) \left(W_{HK}\right)_m$ (as illustrated in Fig. \ref{fig:training_W_K}A). 

Here, $\tilde{x} = W_{IH} s$, and we have additionally defined $s_K := \text{softmax}\left( W_{HK} W_K \tilde{x} \right)$.

\paragraph{QK-MHN-Transformer (with and without input projections).} 

For this model, the gradients for $W_V$ and $W_Q$ are precisely the same as those for the MHN-Transformer with fixed $W_K$ (above). Moreover, in this model, $W_K$ is trained using the loss $\mathcal{L}_K := ||W_K \tilde{x} - W_Q x_q ||_2^2$, giving rise to the gradient equation

\begin{equation}
\begin{aligned}
    \frac{\partial \mathcal{L}_K}{\partial W_K} = 2  \left( W_K \tilde{x} - W_Q x_q \right)  \tilde{x}^T 
\end{aligned}
\end{equation}
where $\tilde{x} = W_{IH} s$ (and where we may take $\tilde\delta_{K} := 2  \left( W_K \tilde{x} - W_Q x_q \right)$ in Fig. \ref{fig:training_W_K}B).

\subsection{Learning Dynamics for $W_Q$, $W_K$, and $W_V$}

In this section, we elaborate on the learning dynamics (as illustrated in Fig. \ref{fig:model_performances}B and Fig. \ref{fig:model_loss_comparison}B) for some of the models discussed in this paper. 

\subsubsection{Empirical Findings} \label{subsubsec:empirical_learning_dynamics}

\begin{figure}[!t]
        \centering
        \includegraphics[width=\textwidth]{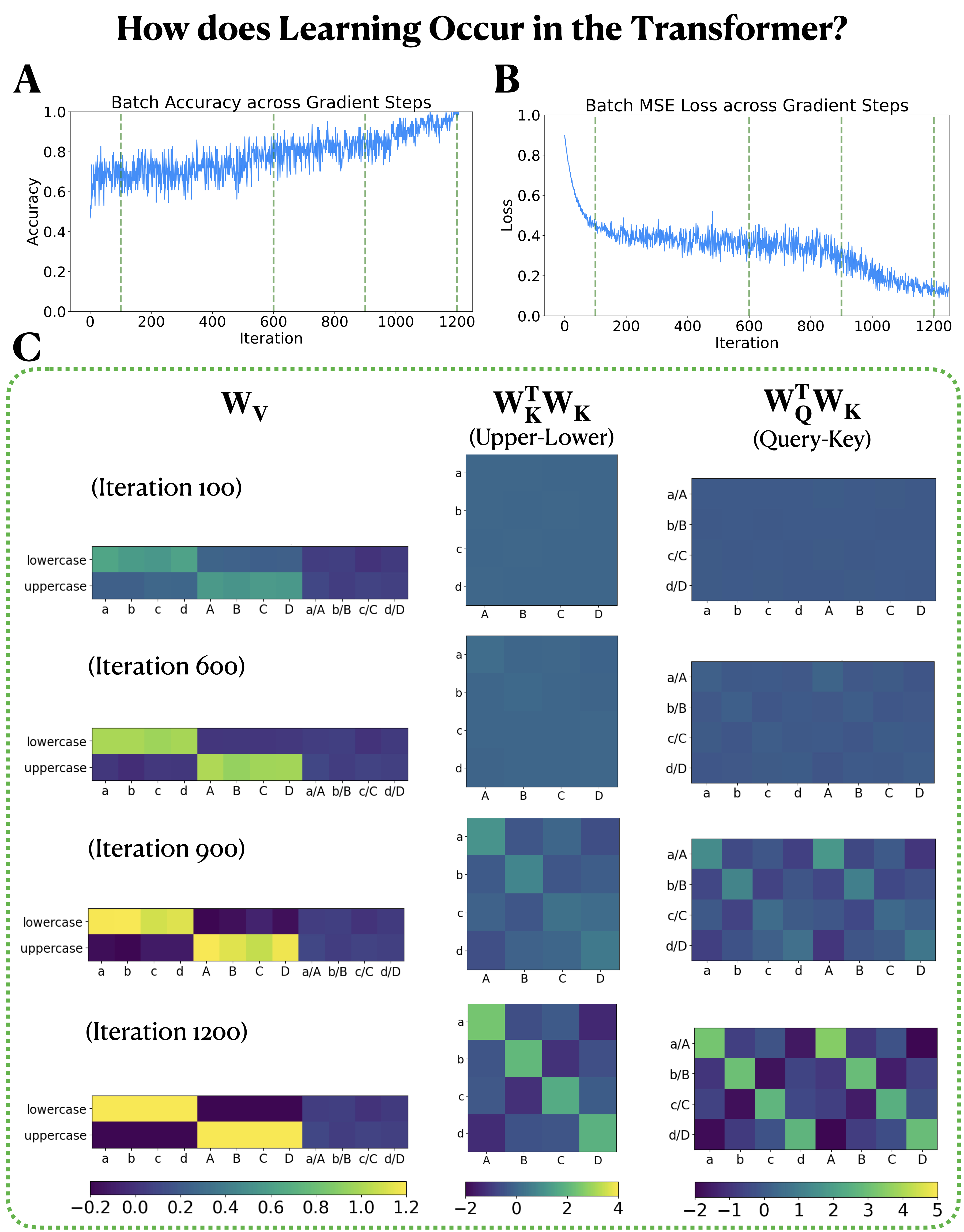}
        \caption{Learning dynamics for the baseline transformer shown across a single trial of training (on the Case Sequence task). \textbf{A.} Accuracy of the model throughout training. \textbf{B.} Batch mean-squared error loss of the model throughout training \textbf{C.} Heatmaps of $W_V$, the uppercase-lowercase submatrix of $W_K^T W_K$, and the query-key submatrix of $W_Q^T W_K$ shown across $4$ key points during learning (which are also marked in panels \textbf{A} and \textbf{B} via dashed green lines).}
        \label{fig:tf_qkv_over_learning}
    \end{figure}

\begin{figure}[!t]
        \centering
        \includegraphics[width=\textwidth]{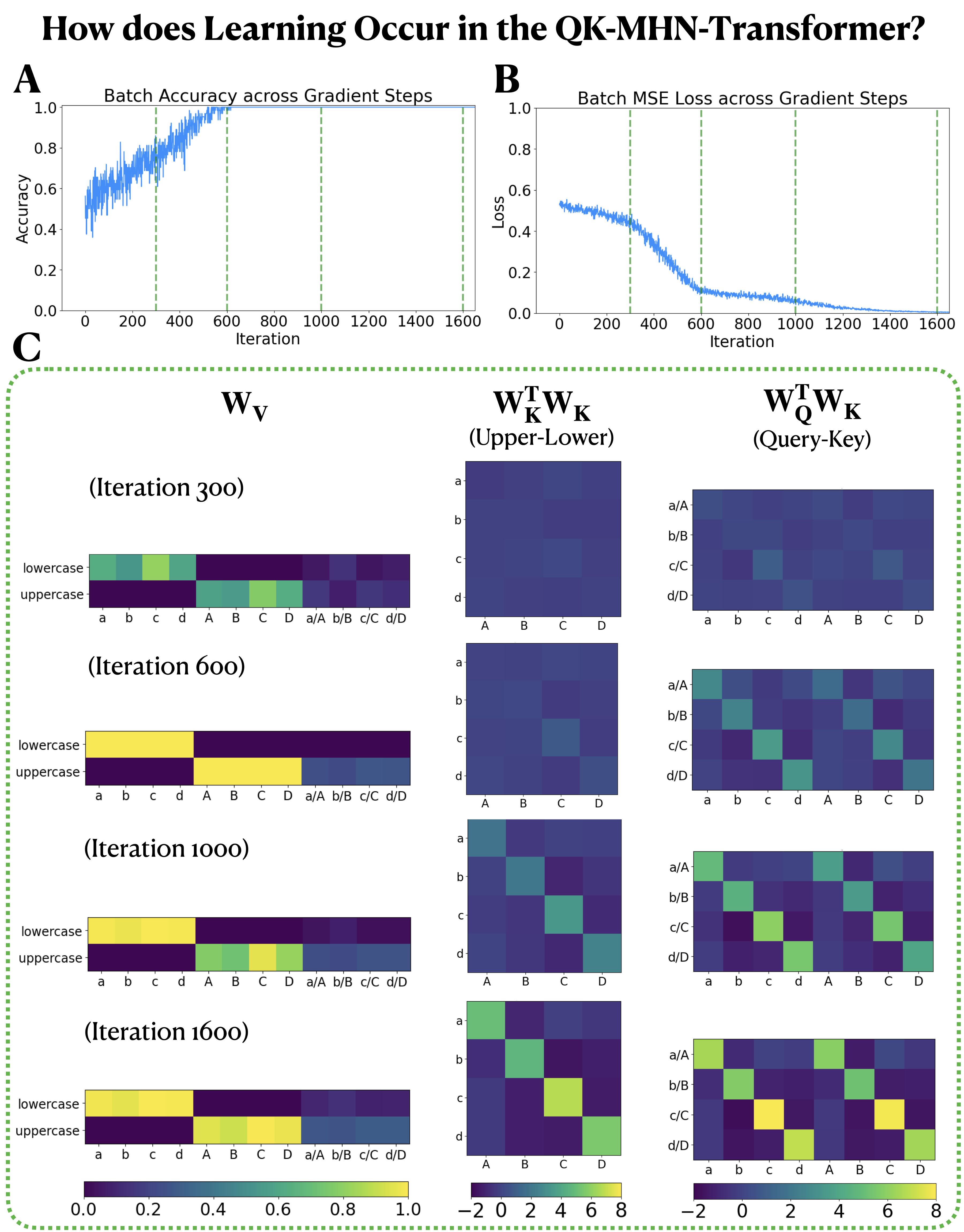}
        \caption{Learning dynamics for the QK-MHN-Transformer shown across a single trial of training (on the Case Sequence task). \textbf{A.} Accuracy of the model throughout training. \textbf{B.} Batch mean-squared error loss of the model throughout training \textbf{C.} Heatmaps of $W_V$, the uppercase-lowercase submatrix of $W_K^T W_K$, and the query-key submatrix of $W_Q^T W_K$ shown across $4$ key points during learning (which are also marked in panels \textbf{A} and \textbf{B} via dashed green lines).}
        \label{fig:qk_mhn_tf_qkv_over_learning}
    \end{figure}

To further unpack how the baseline transformer as well as its slot-free counterpart learns to perform the Case Sequence Task, we analyzed multiple measures of learning performance for the transformer and the QK-MHN-Transformer across a single run of training (Figs. \ref{fig:tf_qkv_over_learning} and \ref{fig:qk_mhn_tf_qkv_over_learning}). We first find that both models' accuracy steadily climb to 100\%, although the QK-MHN-Transformer achieves perfect accuracy sooner in training (Figs. \ref{fig:tf_qkv_over_learning}A and \ref{fig:qk_mhn_tf_qkv_over_learning}A). Further differences are illustrated by their respective training loss profiles; the batch MSE loss for baseline transformer exhibits a two periods of descent with a period of stagnation in between, whereas the QK-MHN-Transformer appears to show multiple stages of descent in succession (Figs. \ref{fig:tf_qkv_over_learning}B and \ref{fig:qk_mhn_tf_qkv_over_learning}B). 

To understand how these differences in learning correspond to the learning of semantically relevant structure in the Case Sequence task, we track how $W_V$, the uppercase key-lowercase key submatrix of $W_K^T W_K$, and the query-key submatrix of $W_Q^T W_K$ evolve across training (Fig. \ref{fig:tf_qkv_over_learning}C and \ref{fig:qk_mhn_tf_qkv_over_learning}C). Across both models, we find that representation of case robustly emerges---as evidenced by the matrix $W_V$--before representations of letter identity and aligment of queries and keys are learned. The learning of the entries in $W_V$ appear to loosely correspond to the first descent periods of either model's loss curve. 

The baseline transformer subsequently undergoes a period of in which the loss appears to stagnate, during which case representations in $W_V$ grow stronger (and more differentiated), while entries in $W_K^T W_K$ and $W_Q^T W_K$ exhibit no discernible structure. Eventually, however, the requisite structure in these matrices begins to appear in parallel, which appears to correlate with the loss decreasing for a second time (see Fig. \ref{fig:tf_qkv_over_learning}B and Iteration 900 of Fig. \ref{fig:tf_qkv_over_learning}C). By Iteration 1200 of this run, representations of letter identity and alignment between queries and keys have grown strong. Thus, we find that while case representations emerge early in training, it takes much longer (relatively) for representations of letter and query-key alignment to occur.

In the QK-MHN-Transformer, after $W_V$ learns representations of case, the model successfully learns to align queries with their respective keys, even before uppercase-lowercase key alignment occurs (see Iteration 600 of Fig. \ref{fig:qk_mhn_tf_qkv_over_learning}C). During this period in which only structure in $W_Q^T W_K$ emerges, the model loss appears to exhibit a second period of descent (see Iterations ~300-600 of Fig. \ref{fig:qk_mhn_tf_qkv_over_learning}B). Following the learning of query-key alignment, the model then begins to learn representations of letter identity, which is correspondingly matched by a third period of descent in the model loss (see Iterations ~600-1000 of Fig. \ref{fig:qk_mhn_tf_qkv_over_learning}B and Iteration 1000 of Fig. \ref{fig:qk_mhn_tf_qkv_over_learning}C). Finally, by Iteration 1600 of this run, the requisite structure across all three matrices has grown strong. Thus, we find that the QK-MHN-Transformer undergoes a stage-like learning process in which representations of case are learned first, followed by alignment between queries and their respective keys, which is finally followed by the learning of case-agnostic letter identity.

A consequence of this analysis is that, while both models ultimately achieve the same computational goals---in the sense that they both solve the Case Sequence task and learn to exhibit the same structure in the matrices $W_V$, $W_K^T W_K$, and $W_Q^T W_K$--they differ in how this requisite structure is learned. It is certainly true that the entries of $W_V$ are learned in a similar manner across both models. However, learning of the information in the other two matrices occurs in a different manner for either model. This warrants further theoretical analysis into how either model learns the given task structure.

\subsubsection{Mathematical Analysis} \label{subsubsec:W_V_learns_first}

Section \ref{subsubsec:empirical_learning_dynamics} illustrates important differences in how the baseline transformer and an MHN-based variant (the QK-MHN-Transformer) learn to solve the Case Sequence Task. To that end, we carry out a deeper theoretical analysis of how either model learns. Presently, we illustrate how both models first robustly learn representations of case (as exhibited by $W_V$). In future work, we seek to illustrate how structure in the matrices $W_K^T W_K$ and $W_Q^T W_K$ emerges for the baseline transformer and our MHN-based variants.

Our central result is that $W_V$ can successfully learn to exhibit case structure, even in the absence of any learning for $W_K$ and $W_Q$:

\begin{thm} \label{thm:W_V_learning}
    Fix the input dimension $d = 3L$ (where $L$ is the number of letters used) and a context window length $C < L$. Furthermore, define the MSE loss across a given batch $\left( (x_t^{(b)})_{t=1}^{C}, x_q^{(b)}, y^{(b)}  \right)_{b=1}^{B}$ of size $B$ to be $$\mathcal{L}\left( \left( \hat{y}^{(b)}, y^{(b)}  \right)_{b=1}^{B} \right) = \frac{1}{2B} \sum_{b=1}^{B} \norm{\hat{y}^{(b)} - y^{(b)}}_2^2$$ where $\hat{y}$ denotes a model output. Consider the following two models:
    \begin{enumerate}
        \item A baseline transformer where $W_K, W_Q \in \RR^{N \times d}$ are held fixed at initialization, with each entry in these two matrices sampled iid from $\mathcal{N}\left( 0, \frac1N \right)$ and only $W_V$ is learned through stochastic minibatch gradient descent (given by Eq. \ref{eq:tf_WV_gradient}). For simplicity, we also assume that the softmax inverse temperature is given by $\beta = 1$ (rather than $\beta = \frac{1}{\sqrt{N}}$), which will become clearer in Lemma \ref{lemma:tf_equivalence} below.

        \item An MHN-based transformer (with input projection weights $W_{HI}$) where $W_K, W_Q \in \RR^{N \times d}$ are held fixed at initialization, with each entry in these two matrices sampled iid from $\mathcal{N}\left( 0, \frac1N \right)$ and only $W_V$ is learned through stochastic minibatch gradient descent (given by Eq. \ref{eq:slotfree_tf_WV_gradient}). Moreover, suppose that $n_h = C$ and that for any given input sequence, each entry in $W_{VH}^{(0)}$ and $W_{IH}^{(0))}$ is sampled iid from $\text{Unif}\left( [0, 1] \right)$ and that each entry in $W_{HK}^{(0)}$ is sampled iid from $\text{Unif}\left( [0, a] \right)$, where $a$ is chosen to satisfy $a << 1$.
    \end{enumerate}

    Finally, suppose that the initial weights of $W_V$ are bounded in magnitude (by a constant independent of $N$). Then, in the limit of large $N$ and batch size $B$, training $W_V$ with stochastic minibatch gradient descent with step size $\eta = \Theta(1)$ (and with $\eta < \frac{C^2}{2\gamma d}$, where $\gamma$ is the fraction of all input sequences in which any one given letter appears) in either model will cause each column of $W_V$ corresponding to a lowercase letter to converge to $\begin{bmatrix}
        1  \\
        0\\
    \end{bmatrix}$ and each column of $W_V$ corresponding to an uppercase letter to converge to $\begin{bmatrix}
        0 \\
        1 \\
    \end{bmatrix}$ with high probability. In particular, $W_V$ will learn case-sensitive representations, in which columns corresponding to like (resp. opposite) cases will be aligned (resp. anti-aligned).
\end{thm}

To prove this theorem, we first show that both of the above models are essentially the same. That is, for any given input sequence $(x_t)_{t=1}^{C}$, with high probability the 1-winner MHN in the MHN-based transformer is able to perfectly store all the key value pairs $(k_t, v_t)_{t=1}^{C}$ in memory (without any overlap), so that the output of the slot free model becomes

$$\hat{y} = W_V W_{IH} \sm\left( W_{HK} W_Q x_q \right) = W_V \begin{bmatrix}
    \vert & \vert &  & \vert \\
    x_1   & v_2 & \dots & x_{C}   \\
    \vert & \vert &  & \vert 
\end{bmatrix} \sm\left( \begin{bmatrix}
    \vert & \vert &  & \vert \\
    k_1   & k_2 & \dots & k_C   \\
    \vert & \vert &  & \vert 
\end{bmatrix}^T \right) q$$ $$ = \sum_{i=1}^{C} \sm\left( k_t^T q \right) v_t$$ which is precisely the output of the baseline transformer. Consequently, it suffices to prove the stated result for the baseline transformer. In this latter case, fixing the matrices $W_K$ and $W_Q$ causes the attention to become uniform over all items in the context window, yielding noisy gradient updates for $W_V$. Nonetheless, when the batch size $B$ is sufficiently large, $W_V$ is effectively being optimized via gradient descent on a fixed convex objective that attains its minimum at the matrix $W_V^{\star}$ whose columns are $\begin{bmatrix}
        1 \\
        0
    \end{bmatrix}$ and $\begin{bmatrix}
        0 \\
        1
    \end{bmatrix}$ depending on whether they correspond to lowercase or uppercase letters, respectively. We now go through the full proof, step by step.

\begin{lemma}   \label{lemma:tf_equivalence}
    In the limit of large $N$, the MHN-based transformer (with input projection weights $W_{HI}$) defined in Theorem \ref{thm:W_V_learning} is computationally equivalent to the baseline transformer with high probability, in the sense that they compute the same function given by $\hat{y} = \sum_{i=1}^{C} \sm\left( x_q^T W_Q^T W_K x_i \right) W_V x_i$ for any input sequence $\left( (x_t)_{t=1}^{C}, x_q \right)$.
\end{lemma}

\begin{proof}
    Consider any sequence of inputs $\left( (x_t)_{t=1}^{C}, x_q \right)$, where each $x_t$ (and the query probe $x_q$) is a one-hot vector in $\RR^d$. This implies that each $k_t = W_K x_t \sim \mathcal{N}\left( 0, \frac1N I_{N \times N} \right)$. Now, consider any time step $t \in \{1, \dots, C\}$ during the context window. The row of $W_{HK}^{(t-1)}$ that will be used to store $k_t$ precisely corresponds to the row of $W_{HK}^{(t-1)}$ that yields the largest dot product with $k_t$. In particular, for any row of $W_{HK}^{(t-1)}$ that stores some key $k_{t'} = W_K x_{t'}$ from a previous time step $t' < t$, we have that $$\mathbb{E}\left[ k_t^T k_{t'} \right] = \sum_{i=1}^{N} \mathbb{E}\left[\left( k_t \right)_i\right] \mathbb{E}\left[ \left( k_{t'} \right)_i\right] = 0$$ $$\text{and } \text{Var}\left[ k_t^T k_{t'} \right] = \sum_{i=1}^{N} \mathbb{E}\left[\left( k_t \right)_i^2\right] \mathbb{E}\left[ \left( k_{t'} \right)_i^2\right] = \sum_{i=1}^{N} \frac1N \cdot \frac1N = \frac1N$$ where we have used the fact that, since $x_t$ and $x_{t'}$ are distinct one-hot vectors, $k_t$ and $k_{t'}$ are independent; moreover, the components of $k_t$ and $k_{t'}$ are independent. Next, consider $l \in \RR^{N}$ is any row of $W_{HK}^{(t-1)}$ that has not yet been allocated, i.e. each $l_i \sim \Unif([0, a])$ is iid. Then, $$\mathbb{E}\left[ k_t^T l \right] = \sum_{i=1}^{N} \sum_{i=1}^{N} \mathbb{E}\left[\left( k_t \right)_i\right] \mathbb{E}\left[ l_i\right] = 0$$ and $$\text{Var}\left[ k_t^T l \right] = \sum_{i=1}^{N} \mathbb{E}\left[\left( k_t \right)_i^2\right] \mathbb{E}\left[ l_i^2\right] = \sum_{i=1}^{N} \frac1N \cdot \frac{a^2}{12} = \frac{a^2}{12}$$ where we have used the fact that $\mathbb{E}\left[ l_i^2\right] = \int_{0}^{a} \frac{t^2}{a} dt = \frac{a^2}{12}$. Furthermore, observe that based on the initialization of $W_{HI}$ defined in Section \ref{subsubsec:training_details}, $W_{HI} x_t \in \RR^{C}$ is a unique one-hot vector that is distinct from any of the one-hot vectors $W_{HI} x_{t'}$ for any $t' < t$. Consequently, letting $r$ denote the index of the '1' entry in $W_{HI} x_t$ and $l_r = \left( W_{HK} \right)_r$, we have $$\mathbb{E}\left[ \left(W_{HK} k_t + W_{HI} x_t \right)_r \right] = \mathbb{E}\left[ l_r^T k_t + 1 \right] = 1$$ and $$\text{Var}\left[ \left(W_{HK} k_t + W_{HI} x_t \right)_r \right] = \text{Var}\left[ l_r^T k_t + 1 \right] = \text{Var}\left[ l_r^T k_t \right] = \frac{a^2}{12}.$$ Thus, since $a = o_N(1)$, $\left(W_{HK} k_t + W_{HI} x_t \right)_r$ concentrates to $1$ while for any $r' \in \{1, \dots, n_h\} \\ \{r\}$, $\left(W_{HK} k_t + W_{HI} x_t \right)_{r'}$ concentrates to $0$ in the large $N$ limit. It follows that, with high probability, the $r$th hidden neuron in the 1-winner MHN will be allocated to represent $k_t$ and $v_t$. Because $r$ ranges from $1$ to $n_h = C$ over the time steps of the context window, the 1-winner MHN will be loaded with all the pairs $(k_t, v_t)_{t=1}^{C}$ at the end of the context window with high probability. Moreover, because a distinct hidden neuron in the MHN is allocated at each time step, the matrix $W_{IH}$ will also be loaded with each of $x_1, \dots, x_C$ as its columns (with high probability). Thus, the MHN-based model's functionality perfectly coincides with that of the transformer, in the sense that $\hat{y} = \sum_{i=1}^{C} \sm\left( k_t^T q \right) v_t$. Moreover, the gradients obtained for $W_V$ will coincide with either model (with high probability). 
\end{proof}

Having shown an equivalence between the two models described in the statement of Theorem \ref{thm:W_V_learning}, it now suffices to prove the theorem for the baseline transformer.

\begin{proof}[Proof of Theorem \ref{thm:W_V_learning}]
    Recall from Equation \ref{eq:tf_WV_gradient} (but with additional scaling by a factor of $\frac{1}{2B}$) that the gradient of the stochastic minibatch loss with respect to $W_V$ is $$\frac{\partial \mathcal{L}}{\partial W_V} = \frac{1}{B} \sum_{b=1}^{B} \left( \hat{y}^{(b)} - y^{(b)} \right) \left(\tilde{x}^{(b)}\right)^{T}$$ where $\tilde{x}^{(b)} = \sum_{i=1}^{C} \sm\left( x_q^{(b)} W_Q^T W_K x_i^{(b)} \right) x_i^{(b)}$. Because $x_q^{(b)}$ and $x_i^{(b)}$ are one-hot vectors, the term $x_q^{(b)} W_Q^T W_K x_i^{(b)}$ simply recovers the entry of the matrix $W_Q^T W_K$ that corresponds to the indices of the '1' entry in $x_q^{(b)}$ and $x_i^{(b)}$. Because each entry in $W_Q$ and $W_K$ is sampled iid from $\mathcal{N}\left( 0, \frac1N \right)$, for any $1 \leq \alpha, \beta \leq d$, $$\mathbb{E}\left[ \left( W_Q^T W_K \right)_{\alpha \beta} \right] = \sum_{\gamma=1}^{N} \mathbb{E}\left[ \left( W_Q \right)_{\alpha \gamma} \right] \mathbb{E}\left[ \left( W_K \right)_{\beta \gamma} \right] = 0$$ $$\text{and } \text{Var}\left[ \left( W_Q^T W_K \right)_{\alpha \beta} \right] = \sum_{\gamma=1}^{N} \text{Var}\left[ \left( W_Q \right)_{\alpha \gamma} \left( W_K \right)_{\beta \gamma} \right]$$ $$ = \sum_{\gamma=1}^{N} \mathbb{E}\left[ \left( W_Q \right)_{\alpha \gamma}^2 \left( W_K \right)_{\beta \gamma}^2 \right] = \sum_{\gamma=1}^{N} \mathbb{E}\left[ \left( W_Q \right)_{\alpha \gamma}^2 \right] \mathbb{E}\left[ \left( W_K \right)_{\beta \gamma}^2 \right] = N \cdot \frac1N \cdot \frac1N = \frac1N.$$ Thus, in the large-$N$ limit, each entry in $W_Q^T W_K$ concentrates about $0$. Consequently, with high probability the attention mechanism becomes uniform, i.e. $\sm\left( x_q^{(b)} W_Q^T W_K x_i^{(b)} \right) = \frac{1}{C}$ for each $i = 1, \dots, C$. In turn, $\tilde{x} = \frac{1}{C} \sum_{i=1}^{C} x_i^{(b)}$ with high probability. Thus, the calculated gradient simplifies to $$\frac{\partial \mathcal{L}}{\partial W_V} = \frac{1}{B} \sum_{b=1}^{B} \left( \hat{y}^{(b)} - y^{(b)} \right) \left(\tilde{x}^{(b)}\right)^{T} = \frac{1}{B} \sum_{b=1}^{B} \left( W_V \tilde{x}^{(b)} - y^{(b)} \right) \left(\tilde{x}^{(b)}\right)^{T}$$ $$ = \frac{1}{B} \sum_{b=1}^{B} \left( W_V \frac1C \sum_{i=1}^{C} x_i^{(b)} - y^{(b)} \right) \left(\frac1C \sum_{i=1}^{C} x_i^{(b)}\right)^{T}.$$

    Now, consider column $j$ of $W_V$; this column corresponds to some one-hot vector $x_J$ that represents some letter in a particular case. That is, $(W_V)_{\cdot, j} = W_V x_J$; additionally, let $y_J \in \RR^2$ denote the output (case) label for the letter given by $x_J \in \RR^d$. By virtue of the $\left(\frac1C \sum_{i=1}^{C} x_i^{(b)}\right)^{T}$ term in the above gradient, $(W_V)_{\cdot, j}$ is provided a nonzero update if and only if $x_J$ appears in the context window as one of $x_1^{(b)}, \dots, x_C^{(b)}$. Thus, we may re-index the outer sum (and re-order the input sequences within the given batch) over the $R \leq B$ input sequences in which $x_J$ appears in the context window. Then, we have $$\frac{\partial \mathcal{L}}{\partial (W_V)_{\cdot, j}} = \frac{R}{B} \cdot \frac{1}{R} \sum_{b=1}^{R} \left( W_V \frac1C \sum_{i=1}^{C} x_i^{(b)} - y^{(b)} \right) \left(\frac1C \sum_{i=1}^{C} \left(x_i^{(b)}\right)_j\right)$$ where each of the sequences $(x_1^{(b)}, \dots x_C^{(b)})$ for $b = 1, \dots, R$ contains $x_J$ as one element. Note moreover that, for each of the given $R$ sequences, $\frac1C \sum_{i=1}^{C} \left(x_i^{(r)}\right)_j = \frac1C(x_J)_j = \frac1C$, so we may further write  
    $$\frac{\partial \mathcal{L}}{\partial (W_V)_{\cdot, j}} = \frac{R}{BC} \cdot \frac{1}{R} \sum_{b=1}^{R} \left( W_V \frac1C \sum_{i=1}^{C} x_i^{(b)} - y^{(b)} \right)$$ $$ = \frac{R}{BC} \cdot \frac{1}{R} \sum_{b=1}^{R} \left( \frac1C W_V x_J + \frac1C \sum_{i \hspace{0.5mm} : \hspace{0.5mm} x_i^{(b)} \neq x_J, x_{J'}} W_V x_i^{(b)} - y^{(b)} \right)$$ $$ = \frac{R}{BC} \left( \frac1C W_V x_J - \frac{1}{R} \sum_{b=1}^{R} y^{(b)} + \frac{1}{R} \sum_{b=1}^{R} \left( \frac1C \sum_{i \hspace{0.5mm} : \hspace{0.5mm} x_i^{(b)} \neq x_J, x_{J'}} W_V x_i^{(b)} \right) \right).$$ Here, we have defined $x_{J'}$ to be the unique letter with the same letter type and opposite case as $x_J$; since we have indexed over input sequences that already contain $x_J$, none of these input sequences can contain $x_{J'}$. Now, observe that in the large-$B$ limit, $R$ also grows large because $\frac{R}{B}$ converges to the fraction $\gamma \in (0, 1)$ of all case sequences that contain $x_J$ in the context window.
    Thus, because batches are sampled randomly, we may model the $y^{(b)}$ as iid random variables, so that $\frac1R \sum_{b=1}^{R} y^{(b)} \to \mathbb{E}\left[y^{(b)}\right]$ by the law of large numbers. Note that $y^{(b)} = y_J$ with probability $\frac1C$ (i.e. if the letter corresponding to $x_J$ \emph{is} queried); moreover, $y_1^{(b)} \sim \text{Bernoulli}(1/2)$ (and $y_2^{(b)} = 1 - y_1^{(b)}$) with probability $1 - \frac1C$ (i.e. if the queried letter type one of the $L-1$ letter types that is \emph{not} the letter type for $x_J$). Altogether, we have that $\mathbb{E}\left[y^{(b)}\right] = \frac1C y_J + \left( 1 - \frac1C \right)\left( \frac12 \begin{bmatrix}
        1 \\ 
        0
    \end{bmatrix} + \frac12 \begin{bmatrix}
        0 \\ 
        1
    \end{bmatrix}\right) = \frac1C y_J + \left(  1 - \frac1C \right)\begin{bmatrix}
        1/2 \\
        1/2
    \end{bmatrix}$.
    Similarly, because the $R$ quantities $ \frac1C \sum_{i \hspace{0.5mm} : \hspace{0.5mm} x_i^{(b)} \neq x_J, x_{J'}} W_V x_i^{(b)}$, for $b \in \{1, \dots, R\}$, are independent, the law of large numbers again implies that $$\frac{1}{R} \sum_{b=1}^{R} \left( \frac1C \sum_{i \hspace{0.5mm} : \hspace{0.5mm} x_i^{(b)} \neq x_J, x_{J'}} W_V x_i^{(b)} \right) \to \mathbb{E}\left[ \frac1C \sum_{i \hspace{0.5mm} : \hspace{0.5mm} x_i^{(b)} \neq x_J, x_{J'}} W_V x_i^{(b)} \right]$$ $$= \frac{C-1}{C} \mathbb{E}_{x \neq x_J, x_{J'}}\left[ W_V x \right] = \left( 1 - \frac1C \right) W_V \mathbb{E}_{x \neq x_J, x_{J'}}[x]$$ where $x \in \RR^d$ is sampled from the distribution of one-hot encodings of letters excluding $x_J$ and $x_{J'}$. This distribution is uniform over the remaining $2L-2$ letters.

    As a result, for each $o \in \{1, 2\}$, $$\left(W_V \mathbb{E}_{x \neq x_J, x_{J'}}[x]\right)_o = (W_V)_o^T \mathbb{E}_{x \neq x_J, x_{J'}}[x] = \frac{1}{2L-2} \sum_{l \neq j, j'} (W_V)_{ol}.$$ Here, we have defined $j' \in \{1, \dots, 2L\}$ as the one-hot index for $x_{J'}$ (and $j$ as the one-hot index of $x_J$). Moreover, we have used the fact that $\mathbb{E}_{x \neq x_J, x_{J'}}[x]$ is simply an average of the $2L-2$ one-hot vectors that are not $x_J$ or $x_{J'}$. 
    Thus, in the large-$B$ limit, we find that $\frac{\partial \mathcal{L}}{\partial (W_V)_{oj}}$ concentrates to

    \begin{equation}    \label{eq:effective_WV_grad}
        \frac{\gamma}{C} \left( \frac1C \left((W_V)_{oj} - (y_J)_o\right) + \left(1 - \frac1C\right)\left( \frac{1}{2L-2} \sum_{l \neq j, j'} (W_V)_{ol} - \frac12 \right) \right).
    \end{equation}
    
Here, we observe that the gradient is a sum of two terms: one that drives $(W_V)_{oj}$ towards $(y_J)_o$ and another term that drives the sum of the elements corresponding to other letters in the same row of $W_V$ towards $\frac12$. Intuitively, gradient descent should result in a matrix $W_V^\star$ whose columns are near $\begin{bmatrix}
        1 \\
        0
    \end{bmatrix}$ or $\begin{bmatrix}
        0 \\
        1
    \end{bmatrix}$ depending on whether the columns correspond to lowercase or uppercase letters, respectively; this follows from the fact that when the above gradient is zero, we expect $(W_V)_{oj} \approx (y_J)_o$. Formally, the above gradient precisely arises from the loss function 
    \begin{align*}
        \tilde{\mathcal{L}} := \frac{\gamma}{C} \sum_{o \in \{1, 2\}} \sum_{j=1}^{2L} \left[ \frac{1}{2C}\left|(W_V)_{oj} - (y_J)_o\right|^2 + \frac12 \left( 1 - \frac1C \right) \frac{1}{2L-2} \left(\sum_{l \neq j, j'} (W_V)_{ol} (W_V)_{oj} \right)  \right. \\ - \left. \frac12 \left(  1 - \frac1C \right) (W_V)_{oj} \right] \\ 
    \end{align*}

    \begin{align*}
        = \frac{\gamma}{C} \sum_{o \in \{1, 2\}} \left[ \frac12 \left( 1 - \frac1C \right) \frac{1}{2L-2} \left(\sum_{l = 1}^{2L} (W_V)_{ol} \right)^2 + \sum_{j=1}^{2L} \left[ -\frac1C (W_V)_{oj}(y_J)_o + (y_J)_o^2  \right. \right. \\+ \left( \frac{1}{2C} - \left( 1 - \frac1C \right) \frac{1}{2L - 2} \right) (W_V)_{oj}^2 - \left. \frac12 \left(  1 - \frac1C \right) (W_V)_{oj} \right] \\ + \sum_{j=1}^{2L} \left. \frac14 \left( 1 - \frac1C \right) \frac{1}{2L-2} \left( (W_V)_{oj} - (W_V)_{oj'}  \right)^2 \right] 
    \end{align*}

    which is convex in $W_V$, where we have noted that $\frac{1}{2C} - \left( 1 - \frac1C \right) \frac{1}{2L - 2} > 0$ since $C < L$. Thus, gradient descent with a suitably small step size will result in convergence to a global minimizer of the loss. From convex analysis, we have the following lemma \hyperlink{tibshirani}{[S5]}:

    \begin{lemma}   \label{lemma:convex_GD_analysis}
        For $f: \RR^n \to \RR$ convex and differentiable, and suppose that its derivative $\nabla f$ is $L$-Lipschitz continuous. Letting $x^\star$ denote a global minimizer of $f$, then performing $T$ steps of gradient descent from the initialization $x^{(0)} \in \RR^n$ with step size $\eta \leq \frac{1}{L}$ will result in a point $x^{(T)}$ such that 
        $$f(x^{(T)}) - f(x^{\star}) \leq \frac{\norm{x^{(0)} - x^\star}_2^2}{2\eta T}.$$ In particular, for $f(x^{(T)})$ to be within some fixed $\epsilon > 0$ of the optimal value $f(x^\star)$, it suffices to perform at most $\left\lfloor\frac{\norm{x^{(0)} - x^\star}_2^2}{2\eta \epsilon}\right\rfloor$ gradient descent steps.
    \end{lemma}

    In particular, from Equation \ref{eq:effective_WV_grad}, it can be verified through a straightforward algebraic calculation that $\nabla_{W_V} \tilde{\mathcal{L}}$ is Lipschitz-continuous (in the entries of $W_V$; in fact, each entry of $\nabla_{W_V} \tilde{\mathcal{L}}$ is linear in the entries of $W_V$) with Lipschitz constant at most $\frac{2\gamma d}{C^2}$ (where $d$ is the dimension of the one-hot inputs, noting that $W_V \in \RR^{2 \times d}$). In particular, for $W_V^\star$ a minimizer of $\tilde{\mathcal{L}}$, $\eta = \Theta(1)$ (satisfying $\eta \leq \frac{C^2}{2\gamma d}$), and any fixed constant $\epsilon > 0$, performing $T_\epsilon := \left\lfloor\frac{\norm{W_V^{(0)} - W_V^\star}_F^2}{2\eta \epsilon}\right\rfloor$ (which is $O(1)$ and independent of $N$) gradient descent steps will cause $\tilde{\mathcal{L}}(W_V^{(T_\epsilon)})$ to be within $\epsilon$ of the global minimum of $\tilde{\mathcal{L}}$.

    We now show that there is a unique minimizer $W_V^\star$ of $\tilde{\mathcal{L}}$ and that $\lim_{t \to \infty} W_V^{(t)} \to W_V^\star$ under gradient descent. Since $\tilde{\mathcal{L}}$ is convex, any global minimizer is precisely characterized by the condition that $\frac{\partial \mathcal{L}}{\partial W_V} = 0$, i.e. for all $o \in \{1, 2\}, j \in \{1, \dots, 2L\}$, $$\frac{\gamma}{C} \left( \frac1C \left((W_V)_{oj} - (y_J)_o\right) + \left(1 - \frac1C\right)\left( \frac{1}{2L-2} \sum_{l \neq j, j'} (W_V)_{ol} - \frac12 \right) \right) = 0.$$ We then define $\alpha_{oj} := (W_V)_{oj} - (y_J)_o$, so that 
    $$0 = \frac{\gamma}{C} \left( \frac1C \alpha_{oj} + \left(1 - \frac1C\right)\left( \frac{1}{2L-2} \sum_{l \neq j, j'} (\alpha_{ol} + (y_L)_o) - \frac12 \right) \right)$$

    $$ \implies \frac1C \alpha_{oj} + \left(1 - \frac1C\right)\left( \frac{1}{2L-1} \sum_{l \neq j} \alpha_{ol} \right) = \left( 1 - \frac1C \right) \left( \frac12 - \frac{1}{2L-2}\sum_{l\neq j, j'} (y_L)_o \right) $$ $$ = \frac12 - \frac{L-1}{2L-2} = 0.$$ Here, we have used $y_L$ to denote the ground truth case label assigned to the one-hot input vector whose '1' is in the $l$th position. We have moreover used the fact that exactly half of the $2L-2$ letters aside from $x_J$ and $x_{J'}$ are uppercase and half are lowercase; thus, $\sum_{l \neq j, j'} (y_L)_o = L-1$. 
    
We may succinctly write the above equation as $$A \alpha_o = 0$$ where $\alpha_o := [\alpha_{o,1}, \dots, \alpha_{o,2L}]^T \in \RR^{L}$ and $A \in \RR^{2L \times 2L}$ is given in block-matrix form as $$A = \left[\begin{array}{c|c}
        B & D \\
        \hline
        D & B  \\
        \end{array}
        \right]$$ where $$B = \begin{bmatrix}
        \frac{1}{C} & \left( 1 - \frac1C \right)\frac{1}{2L-2} & \cdots & \left( 1 - \frac1C \right)\frac{1}{2L-2} \\
        \left( 1 - \frac1C \right)\frac{1}{2L-2} & \frac1C & \cdots & \left( 1 - \frac1C \right)\frac{1}{2L-2} \\
        \vdots & \vdots  &  & \vdots \\
        \left( 1 - \frac1C \right)\frac{1}{2L-2} & \left( 1 - \frac1C \right)\frac{1}{2L-2} & \cdots & \frac1C \\
    \end{bmatrix} \in \RR^{L \times L} \text{ and }$$

    $$D = \begin{bmatrix}
        0 & \left( 1 - \frac1C \right)\frac{1}{2L-2} & \cdots & \left( 1 - \frac1C \right)\frac{1}{2L-2} \\
        \left( 1 - \frac1C \right)\frac{1}{2L-2} & 0 & \cdots & \left( 1 - \frac1C \right)\frac{1}{2L-2} \\
        \vdots & \vdots  &  & \vdots \\
        \left( 1 - \frac1C \right)\frac{1}{2L-2} & \left( 1 - \frac1C \right)\frac{1}{2L-2} & \cdots & 0 \\
    \end{bmatrix} \in \RR^{L \times L} \text{ and }$$

    Noting that $B = \left(\frac1C - \left( 1 - \frac1C \right)\frac{1}{2L-2}\right) I_{L \times L} + \left( 1 - \frac1C \right)\frac{1}{2L-2} \mathbf{1}_L \mathbf{1}_L^T$, we see that $B$ is positive definite; in particular, it has one eigenvalue of $\frac12\left(1 + \frac{1}{C}\right)$ (with eigenvectors in $\text{span}(\{\mathbf{1}_L\})$ and $L-1$ repeated eigenvalues of $\frac{1}{C} - \left( 1 - \frac1C \right)\frac{1}{2L - 2}$ (with all eigenvectors in $\text{span}(\{\mathbf{1}_L\})^\perp$). Additionally, since $D = - \left( 1 - \frac1C \right)\frac{1}{2L-2} I_{L \times L} + \left( 1 - \frac1C \right)\frac{1}{2L-2} \mathbf{1}_L \mathbf{1}_L^T$, we see that it has one eigenvalue of $\frac12\left(1 - \frac{1}{C}\right)$ ((with eigenvectors in $\text{span}(\{\mathbf{1}_L\})$) and $L-1$ repeated eigenvalues of $- \left( 1 - \frac1C \right)\frac{1}{2L - 2}$ (with all eigenvectors in $\text{span}(\{\mathbf{1}_L\})^\perp$).

We intend to show that $\alpha_o$ must necessarily be $0$, for which it suffices to show that $A$ has full rank. In fact, we prove the stronger result that $A$ is positive definite. Because $A$ is expressed in block matrix form, we may invoke that matrix lemma that $A$ is positive definite if and only if $B$ is positive definite (which it is) and the Schur complement $B - DB^{-1}D$ is positive definite. Earlier, we showed that the eigenvectors of $B$ and $D$ perfectly coincide, which means that $B - DB^{-1}D$ has the same eigenvectors. In particular, for the eigenvector $\mathbf{1}_L$, we have $$(B - DB^{-1}D) \mathbf{1}_L = \left(\frac12 \left(1 + \frac1C\right) - \left( \frac12 \left(1 - \frac1C\right) \right) \left( \frac12 \left(1 + \frac1C\right) \right)^{-1} \left( \frac12 \left(1 - \frac1C\right) \right) \right) \mathbf{1}_L$$ $$ = \left(\frac12 \frac{C+1}{C} - \frac12 \frac{(C-1)^2}{C(C+1)}\right) \mathbf{1}_L$$ and $\frac12 \frac{C+1}{C} - \frac12 \frac{(C-1)^2}{C(C+1)} > 0$. Next, consider any eigenvector $v \in \text{span}({\mathbf{1}_T})^\perp$. Here, we have 
        $$(B - DB^{-1}D) v = \left( \frac{1}{C} - \left( 1 - \frac1C \right)\frac{1}{2L - 2} - \left( - \left( 1 - \frac1C \right)\frac{1}{2L - 2} \right)^2  \left( \frac{1}{C} - \left( 1 - \frac1C \right)\frac{1}{2L - 2} \right)^{-1} \right)v$$ $$ = \frac{\left( 2L - C - 1 \right)^2 - (C-1)^2}{C(2L-2)(2L-C-1)} v$$ 
and $\frac{\left( 2L - C - 1 \right)^2 - (C-1)^2}{C(2L-2)(2L-C-1)} > 0$ holds because $\left( 2L - C - 1 \right)^2 > (C-1)^2$ (since $C < L$). Thus, we conclude that the eigenvectors of $B - DB^{-1}D$ (which are the same as those of $B$, $B^{-1}$, and $D$) all have positive eigenvalues, establishing that $B - DB^{-1}D$ is positive definite. In turn, we conclude that $A$ is positive definite.

In particular, we find that $A$ has trivial null space, and hence $A \alpha_o = 0$ implies that $\alpha_o = 0$. This precisely means that the unique minimizer $W_V^\star$ of $\tilde{\mathcal{L}}$ is given by $(W_V)_{oj}^\star = (y_J)_o$, i.e. the column $(W_V)_{\cdot, j}^\star$ is either $\begin{bmatrix}
            1 \\ 
            0
        \end{bmatrix}$ (lowercase) or $\begin{bmatrix}
            0 \\ 
            1
        \end{bmatrix}$ (uppercase) depending on the case identity of the item $x_J$. 
    

    Now, to show that the sequence of weight matrices $W_V^{(T)}$ obtained through gradient descent steps converges to the given $W_V^\star$, we utilize a fact about \emph{strong convexity}---namely that if the Hessian $\nabla_{W_V}^2 \tilde{\mathcal{L}} \in \RR^{4L \times 4L}$ (where $W_V$ is flattened as the vector $[(W_V)_{11}, \dots, (W_V)_{1d}, (W_V)_{21}, \dots, (W_V)_{2d}] \in \RR^{4L}$) has minimum singular value $m  > 0$, then for all $W_1, W_2 \in \RR^{2 \times d}$ we have $$\tilde{\mathcal{L}}(W_2) \geq \tilde{\mathcal{L}}(W_1) + \text{Tr}\left(\nabla_{W_V} \tilde{\mathcal{L}}(W_1) \left( W_2 - W_1 \right)\right) + \frac{m}{2}\norm{W_2 - W_1}_F^2.$$ (See Boyd \& Vandenberge \hyperlink{boyd}{[S6]}, 
    Section 9.1.2, for more details.) Thus, applying this to $W_V^{(T)}$ and $W_V^\star$, and additionally noting that $\nabla_{W_V}\tilde{\mathcal{L}}(W_V^\star) = 0$, we obtain 
    $$\frac{m}{2}\norm{W_V^{(T)} - W_V^\star}_F^2 \leq \tilde{\mathcal{L}}(W_V^{(T)}) - \tilde{\mathcal{L}}(W_V^{\star}) \leq \frac{\norm{W_V^{(0)} - W_V^\star}_F^2}{2 \eta T}$$ where we have invoked Lemma \ref{lemma:convex_GD_analysis}. It would then follow that 
    $$\norm{W_V^{(T)} - W_V^\star}_F \leq \frac{\norm{W_V^{(0)} - W_V^\star}_F}{\sqrt{\eta m T}}$$ which clearly goes to $0$ as $T \to \infty$. 
    
    Thus, to finish our proof, we show that $\nabla_{W_V}^2 \tilde{\mathcal{L}}$ is positive definite, implying that its minimum singular value $m$ is indeed strictly positive. Since $\tilde{\mathcal{L}}$ is convex, $\nabla_{W_V}^2 \tilde{\mathcal{L}}$ is certainly positive semi-definite. In fact, straightforward computation shows that $$\nabla_{W_V}^2\tilde{\mathcal{L}} = \frac{\gamma}{C}\left[\begin{array}{c|c}
        A & 0 \\
        \hline
        0 & A  \\
        \end{array}
        \right] \in \RR^{4L \times 4L}$$ is positive definite. To verify this fact, we again invoke the fact that a block matrix $$\left[\begin{array}{c|c}
        W & X \\
        \hline
        X^T & Z  \\
        \end{array}
        \right]$$ is positive definite if $Z$ is positive definite and the Schur complement $W - XZX^T$ is positive definite. Indeed, $A$ is positive definite and $A - 0 \cdot A^{-1} \cdot 0 = A$, so we conclude that the Hessian is indeed positive definite, i.e. its eigenvalues are strictly positive.
    

\end{proof}






\section*{Supplementary References}

{\small

\hypertarget{Ramsaueretal2021}{[S1]}
Ramsauer, H., Sch\"afl, B., Lehner, J., Seidl, P., Widrich, M., Gruber, L., Holzleitner, M., Pavlovic, M., Sandk\"uhler, M., Gerrits, T., et al. (2021). \textit{Hopfield Networks is All You Need}. \textit{arXiv preprint} arXiv:2008.02217.

\hypertarget{deHaan}{[S2]} 
de Haan, L. and Ferreira, A. (2006). \textit{Extreme Value Theory: An Introduction} (Springer Series in Operations Research and Financial Engineering). Springer.

\hypertarget{Hall1979}{[S3]}
Hall, P. (1979). On the rate of convergence of normal extremes. \textit{Journal of Applied Probability}, \textit{16}(2), 433--439. \url{http://www.jstor.org/stable/3212912}

\hypertarget{DasGuptaEtal2014}{[S4]}
DasGupta, A., Lahiri, S. N., \& Stoyanov, J. (2014). Sharp fixed $n$ bounds and asymptotic expansions for the mean and the median of a Gaussian sample maximum, and applications to the Donoho--Jin model. \textit{Statistical Methodology}, \textit{20}, 40--62. Re-sampling and Contemporary Inference: A tribute to Kesar Singh. DOI: \url{https://doi.org/10.1016/j.stamet.2014.01.002}. Available at: \url{https://www.sciencedirect.com/science/article/pii/S1572312714000124}

\hypertarget{tibshirani}{[S5]} 
Tibshirani, R. (2013). \textit{Gradient Descent -- Convergence Analysis}.

\hypertarget{boyd}{[S6]} 
Boyd, S. and Vandenberghe, L. (2004). \textit{Convex Optimization}. Cambridge University Press.

}








\end{document}